\documentclass[10pt,twocolumn,letterpaper]{article}

\usepackage{cvpr}              
\definecolor{cvprblue}{rgb}{0.21,0.49,0.74}
\usepackage[pagebackref,breaklinks,colorlinks,allcolors=cvprblue]{hyperref}
\newtheorem{lemma}{Lemma}
\newtheorem{proposition}{Proposition}
\newtheorem{corollary}{Corollary}
\newtheorem{definition}{Definition}
\newtheorem{theorem}{Theorem}
\usepackage{booktabs}
\usepackage{bm}
\usepackage{tabularx}
\usepackage{multirow}
\usepackage{graphicx}
\usepackage{subcaption}
\usepackage{algorithm}
\usepackage{algorithmic}
\usepackage{lipsum}
\usepackage[hyphens]{url}
\usepackage{hyperref}
\hypersetup{
    colorlinks=true,
    linkcolor=blue,
    urlcolor=blue,
    citecolor=blue
}
\newcounter{suppsection}
\renewcommand{\thesuppsection}{A\arabic{suppsection}}

\newcommand{\suppsection}[1]{%
    \refstepcounter{suppsection}%
    \section*{\thesuppsection: #1}%
    \addcontentsline{toc}{section}{Supplementary Section \thesuppsection: #1}%
    \label{suppsec:\thesuppsection}
}

\newcounter{suppsubsection}[suppsection]

\renewcommand{\thesuppsubsection}{\thesuppsection.\arabic{suppsubsection}}

\newcommand{\suppsubsection}[1]{%
    \refstepcounter{suppsubsection}
    \subsection*{\thesuppsubsection: #1}
    \addcontentsline{toc}{subsection}{Supplementary Section \thesuppsubsection: #1}%
    \label{suppsec:\thesuppsubsection}%
}
\def\paperID{} 
\def\confName{CVPR}
\def\confYear{2026}

\title{DMGD: Train-Free Dataset Distillation with Semantic-Distribution Matching in Diffusion Models}

\author{
Qichao Wang$^{1}$ \quad
Yunhong Lu$^{1}$ \quad
Hengyuan Cao$^{1}$ \quad
Junyi Zhang$^{1}$\quad
Min Zhang$^{1,2,3}$\thanks{Corresponding author} \\
$^{1}$ Zhejiang University  \quad
$^{2}$ Shanghai Institute for Advanced Study-Zhejiang University \\
$^{3}$Shanghai Institute for Mathematics and Interdisciplinary Sciences\\
\small
\{qichaowang, yunhonglu, caohy, gavin.jy,  min\_zhang\}@zju.edu.cn
}
\begin{document}
\maketitle
\begin{abstract}
Dataset distillation enables efficient training by distilling the information of large-scale datasets into significantly smaller synthetic datasets. Diffusion based paradigms have emerged in recent years, offering novel perspectives for dataset distillation. However, they typically necessitate additional fine-tuning stages, and effective guidance mechanisms remain underexplored. To address these limitations, we rethink diffusion based dataset distillation and propose a Dual Matching Guided Diffusion (DMGD) framework, centered on efficient training-free guidance. We first establish \textbf{ Semantic Matching} via conditional likelihood optimization, eliminating the need for auxiliary classifiers. Furthermore, we propose a dynamic guidance mechanism that enhances the diversity of synthetic data while maintaining semantic alignment. Simultaneously, we introduce an optimal transport (OT) based \textbf{Distribution Matching} approach to further align with the target distribution structure. To ensure efficiency, we develop two enhanced strategies for diffusion based framework: Distribution Approximate Matching and Greedy Progressive Matching. These strategies enable effective distribution matching guidance with minimal computational overhead. Experimental results on ImageNet-Woof, ImageNet-Nette, and ImageNet-1K demonstrate that our training-free approach achieves significant improvements, outperforming state-of-the-art (SOTA) methods requiring additional fine-tuning by average accuracy gains of $2.1\%$, $5.4\%$, and $2.4\%$, respectively. The code is available on \href{https://github.com/solomonWQC/DMGD}{https://github.com/solomonWQC/DMGD}
\end{abstract}    
\section{Introduction}
\label{sec:intro}

\begin{figure}[h] 
    
    \centering
    \includegraphics[width=\linewidth]{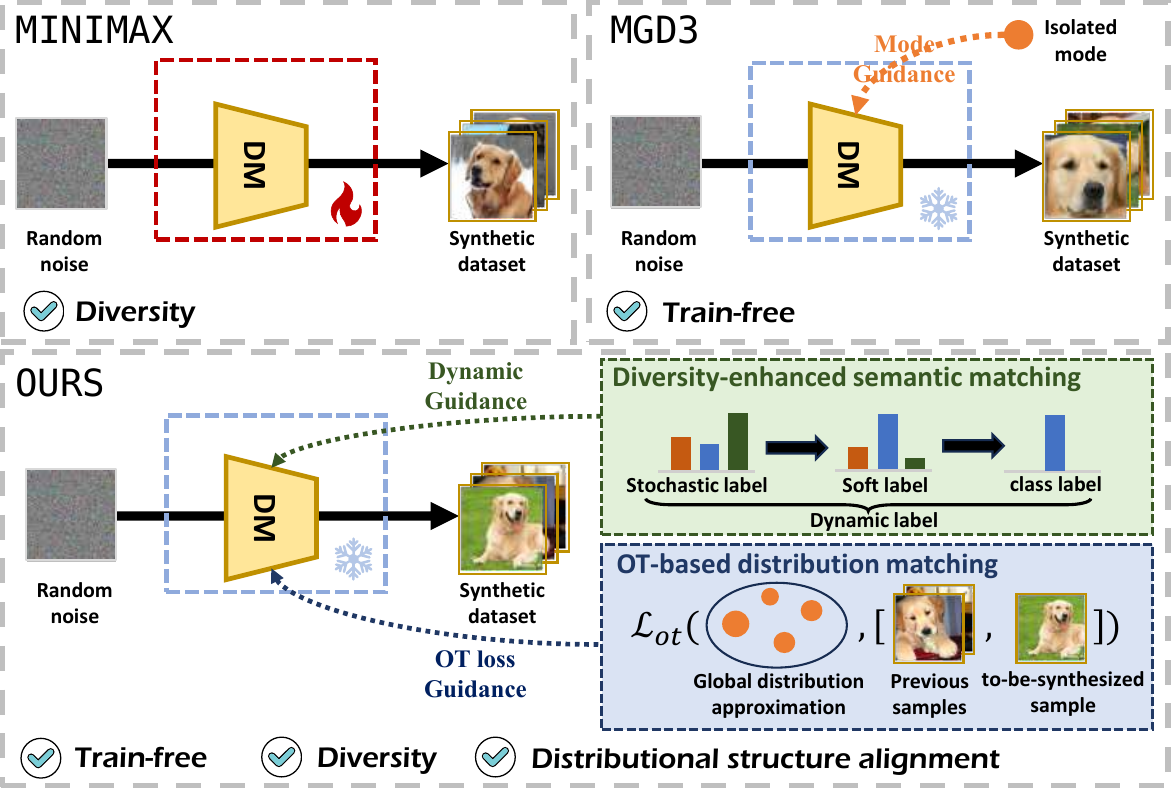}
    \caption{A comparison of different diffusion-based paradigms for dataset distillation. The Minimax \cite{gu2024efficient} relies on additional fine-tuning of diffusion models on the target dataset. In contrast, MGD$^3$ \cite{chan2025mgd} employs isolated guidance via predicted mode points, neglecting the underlying distribution structure and inter-sample diversity. We decouple dataset distillation into semantic matching and distribution matching.  
    Our method achieves enhanced diversity and distribution alignment without any training, resulting in superior dataset distillation performance.}
    \label{fig:teaser}

\end{figure}
The exponential growth of datasets has significantly advanced artificial intelligence, yet concurrently introduces formidable challenges regarding storage overheads and computational demands \cite{achiam2023gpt, schuhmann2022laion, lin2025towards}. In this context, dataset distillation has emerged as a prominent research paradigm \cite{lei2023comprehensive,yang2024datasetdistillationlearning,zhao2020dataset,kungurtsev2024datasetdistillationprinciplesintegrating,wang2020datasetdistillation}. The core objective of dataset distillation is to synthesize surrogate datasets that preserve the critical information of the original training data. These surrogate datasets enable model training with substantially lower storage and computational costs, thereby democratizing access to advanced artificial intelligence.

Ongoing research efforts have spurred novel methodologies in dataset distillation, including gradient matching \cite{zhao2020dataset}, distribution matching \cite{zhao2023dataset}, and trajectory matching \cite{cazenavette2022dataset}. These approaches optimize matching objectives through iterative gradient backpropagation into synthetic samples, achieving remarkable performance on small-scale datasets like CIFAR-10 \cite{canas2012learning}. SRe$^2$L \cite{yin2023squeeze} proposes a framework that decouples model training from data synthesis, first extending dataset distillation to large-scale datasets.
In parallel, researchers are exploring generative models for dataset distillation \cite{cazenavette2023generalizing}. Recent breakthroughs demonstrate that diffusion models \cite{ho2020denoising}, achieve better dataset distillation performance\cite{Su_2024_CVPR,gu2024efficient,chan2025mgd,chen2025influence}. However, they typically necessitate additional fine-tuning stages, and effective guidance mechanisms remain underexplored.\par
This work explores strategies to boost diffusion based dataset distillation performance at sampling process while avoiding additional training. Our insight is that designing effective guidance objectives unlocks the potential of diffusion models for dataset distillation. We prove that under semantic alignment, the optimal transport (OT) distance between surrogate and target datasets serves as an upper bound for their risk discrepancy. Thus, we propose the Dual-Matching Guided Diffusion Model (DMGD), an innovative framework incorporating two decoupled objectives: \textbf{Semantic Matching} and \textbf{Distribution Matching}. First, we formalize the semantic matching objective via conditional likelihood optimization. We incorporate classifier-free guidance \cite{ho2022classifier}, which establishes the connection between diffusion models and conditional likelihood optimization, thereby achieving semantic alignment without requiring additional discriminative models. However, conditional likelihood optimization with hard labels inevitably impairs diversity, confining the diffusion model's outputs to high-density regions \cite{um2025minority}. To address these limitations, we propose a dynamic semantic matching guidance that modulates label guidance across sampling stages to enable exploration of the diffusion model's generative distribution.\par
Beyond extracting semantic information, capturing the distributional characteristics of the target dataset is equally critical for effective dataset distillation. Optimal transport distance serves as the theoretical foundation for quantifying distributional discrepancies \cite{arjovsky2017wasserstein}. Consequently, we propose an optimal transport based Distribution Matching objective and devise two improvement strategies to enhance its efficiency. \textit{Distribution Approximate Matching} utilizes optimal quantization theory, extracting an approximate distribution that preserves the distribution structure of the target dataset to enable efficient optimal transport computation. \textit{Greedy Progressive Matching} adopts a greedy optimization paradigm that progressively optimizes each synthetic sample to align distributions, addressing diffusion model limitation in multi-sample optimization.
Our main contributions are summarized as follows:
\begin{itemize}
    \item We rethink the dataset distillation framework based on diffusion models and propose a training-free guided framework DMGD, which consists of two  guidance components: semantic matching and distribution matching. We conduct extensive experiments demonstrating that our approach achieves state-of-the-art (SOTA) performance without requiring additional training time. 
    \item In semantic matching, we propose a novel soft label based dynamic guidance mechanism, enhancing diversity while ensuring the semantic alignment.
    \item In distribution matching, we propose a guidance loss based on optimal transport and theoretically prove that it can optimize the upper bound of the risk for the distillation dataset. To further enhance computational efficiency, we also introduce two strategies: distribution approximate matching and greedy progressive matching.
\end{itemize}

\section{Related work}
\paragraph{Diffusion based dataset distillation }Recently, diffusion \cite{ho2020denoising} models have provided a powerful foundation for dataset distillation. Minimax \cite{gu2024efficient} introduced an efficient fine-tuning-based approach to further enhance the alignment between diffusion models and target datasets. IGD \cite{chen2025influence} incorporates additional classifier training trajectories to guide the diffusion process. However, these methods require extra training, which limits the efficiency. Both D$^4$M \cite{Su_2024_CVPR} and MGD$^3$ \cite{chan2025mgd} leverage mode centers discovered by clustering algorithms to control synthetic sample generation. Nevertheless, these approaches may overemphasize invalid modes from clustering, such as proximate cluster centers or outliers, thereby disrupting distribution structure alignment. They also neglect interrelationships among synthetic samples, leading to diversity deficiencies. Concurrent to our work, \cite{cui2025optimizing} also explores the application of optimal transport-based diffusion models in dataset distillation. As illustrated in  \Cref{fig:teaser}, our method focuses on training-free efficient guidance mechanisms and theoretically decouples the design space into semantic matching and distribution matching. Meanwhile, our approach optimizes the complete synthetic data distribution rather than individual sample, further enhancing diversity and distribution structure alignment. Additional comparative discussions with other dataset distillation paradigms are provided in Appendix \ref{supsec:BG}.
\begin{figure*}[t]
    \centering
    \includegraphics[width=1\linewidth]{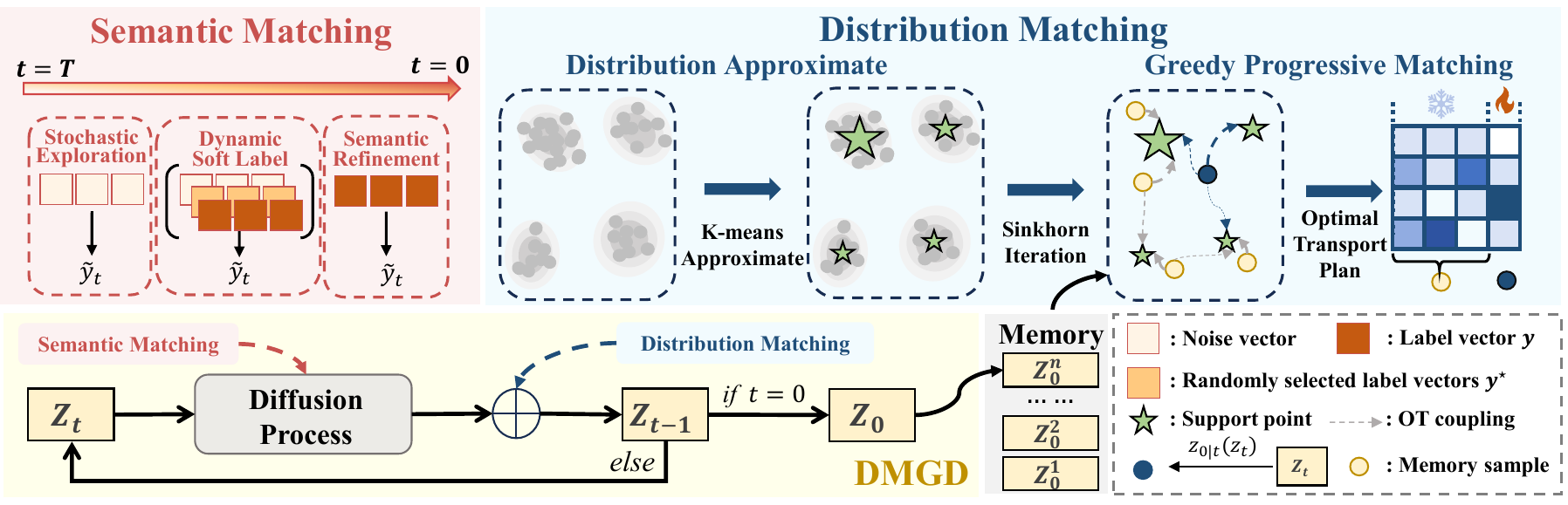}
    \caption{\textbf{Framework of our DMGD method.} Our method establishes two guidance modules during the sampling process: semantic matching and distribution matching. In semantic matching, we propose a dynamic soft label mechanism to unlock the potential of diffusion models for diversified generation while ensuring semantic alignment. In distribution matching, we optimize optimal transport computation through distribution approximation and greedy progressive matching to enable optimal transport-based distribution alignment guidance. We present the corresponding pseudocode in the Appendix \ref{suppsec:ID} \Cref{alg:overall}.}
    \label{fig:DMGD}
\end{figure*}
\section{Preliminaries}

\subsection{Dataset distillation}
Given a large-scale labeled dataset $\mathcal{T} =    \{(x_i,y_i)\}^{N_\mathcal{T}}_{i=1} $, where $x \in \mathbb{R}^D$, $y \in \mathcal{Y}=\{1,2,...,C\}$, we aim to obtain a surrogate dataset $\mathcal{S}=\{\bar{x_i} ,\bar{y_i} \}^{N_\mathcal{S}}_{i=1}$, where $\bar{x} \in \mathbb{R}^D$, $\bar{y} \in \mathcal{Y}=\{1,2,...,C\}$ and $N_\mathcal{T}\gg N_\mathcal{S}$. The surrogate dataset $\mathcal{S}$ should retain the critical information from $\mathcal{T}$ such that the model $\theta$ trained on $S$ achieves effective and comparable performance on the target dataset.
\begin{equation}
        \mathbb{E}_{(x,y) \sim \mathcal{T}} \left[ \ell(x, y; \theta_{\mathcal{S}}^\star) \right] \simeq \mathbb{E}_{(x,y) \sim \mathcal{T}} \left[ \ell(x, y; \theta_{\mathcal{T}}^\star) \right]
\end{equation}
Here, $\theta_\mathcal{S}^\star$ and $\theta_\mathcal{T}^\star$ are the optimal parameters obtained from training on $\mathcal{S}$ and $\mathcal{T}$, respectively. $\ell(x, y; \theta)$ denotes the evaluation function designed to validate the performance of model $\theta$ on data pairs $(x, y)$. 
\subsection{Diffusion Models}
Diffusion models \cite{ho2020denoising} comprise a forward process $\{q(\boldsymbol{x}_{t})\}_{t\in[0,T]}$ that gradually adds noise to data $\boldsymbol{x}_{0} \sim q(\boldsymbol{x}_{0})$, alongside a learned reverse process $\{p(\boldsymbol{x}_{t})\}_{t\in[0,T]}$ targeting to denoise the data. The forward process is formulated as $q(\boldsymbol{x}_{t}|\boldsymbol{x}_{0}):=\mathcal{N}(\sqrt{\alpha_{t}}\boldsymbol{x}_{0},(1-\alpha_{t})\mathbf{I})$ and $q(\boldsymbol{x}_{t}):= \int q(\boldsymbol{x}_{t}|\boldsymbol{x}_{0})q(\boldsymbol{x}_{0}) \mathrm{d}\boldsymbol{x}_{0}$, with $\alpha_{t}$ representing a noise schedule. The reverse process, initialized from $p(\boldsymbol{x}_{T}):=\mathcal{N}(\mathbf{0},\mathbf{I})$, is characterized by a parameterized denoiser $\boldsymbol{\epsilon}_{\theta}^{t}(\boldsymbol{x}_{t})$, which aims to predict the noise added to $\boldsymbol{x}_{0}$. The denoiser $\epsilon_\theta$ is optimized by minimizing:
\begin{equation}
  \mathcal{L}_{\mathrm{DM}}:=\mathbb{E}_{x_{0},t,\boldsymbol{\epsilon}}\left\|\boldsymbol{\epsilon}_{\theta}^{t}(\sqrt{\alpha_{t}}\boldsymbol{x}_{0}+\sqrt{1-\alpha_{t}}\boldsymbol{\epsilon})-\boldsymbol{\epsilon}\right\|^{2}_{2}
  \label{eq:diffusion_loss}
\end{equation}
where $\boldsymbol x_{0}\sim q(\boldsymbol x_{0}),t\sim \mathcal{U}(0,T)$, and $\boldsymbol{\epsilon} \sim \mathcal{N}(\mathbf{0},\mathbf{I})$. A more widely adopted approach is the Latent Diffusion Model \cite{rombach2022high}, which leverages a Variational Autoencoder \cite{kingma2013auto} to encode input $x$ into latent space samples $z$. Our method focuses on the sampling process of LDM. For compact representation, we define whole sample process as $z_{t-1}=D_\theta(z_t,t,y)$, where $t$ is the sampling step, and $y$ is the label condition. Furthermore, we can incorporate other conditional guidance during sample process to guide diffusion \cite{yu2023freedom}. Given a differentiable conditioning function $E(z_t,c)$, where $c$ represents another conditional input of arbitrary form, we can define a single-step guided diffusion process as:
\begin{equation}
\label{TFG_paper}
    z_{t-1}=D_\theta(z_t,t,y)-\rho_t\nabla_{z_t} E(z_t,c)
\end{equation}
\subsection{Optimal Transport}
Optimal Transport \cite{montesuma2024recent} provides a principled framework for measuring the dissimilarity between two probability distributions. Given two discrete probability distributions $\mathbf{a} \in \Delta^n$ and $\mathbf{b} \in \Delta^m$, where $\Delta$ denotes the probability simplex, and a cost matrix $\mathbf{C} \in \mathbb{R}^{n \times m}$, where $\mathbf{C}_{ij}$ represents the cost of moving mass from $\mathbf{a}_i$ to $\mathbf{b}_j$, the OT problem seeks a transport plan $\gamma \in \mathbb{R}^{n \times m}_+$ that minimizes the total transportation cost:
\begin{equation}
W(\mathbf{a}, \mathbf{b}) = \min_{\gamma \in \Gamma(\mathbf{a}, \mathbf{b})} \langle \gamma, \mathbf{C} \rangle,  
\end{equation}  
where $\Gamma(\mathbf{a}, \mathbf{b}) = \left\{ \gamma \in \mathbb{R}^{n \times m}_+ \mid \gamma \mathbf{1} = \mathbf{a}, \gamma^T \mathbf{1} = \mathbf{b} \right\}$ is the set of admissible coupling matrices. \(\langle * \rangle\) is inner product. The exact OT problem is computationally expensive for large-scale applications. To improve efficiency, a common approach is to introduce an entropic regularization term:
\begin{equation} 
W_{\varepsilon}(\mathbf{a}, \mathbf{b}) = \min_{\gamma \in \Gamma(\mathbf{a}, \mathbf{b})} \langle \gamma, \mathbf{C} \rangle - \varepsilon H(\gamma),  
\end{equation}
where $\varepsilon > 0$ controls the strength of the regularization, and $H(\gamma) = -\sum_{i,j} \gamma_{ij} \log \gamma_{ij}$ is the entropy of the transport plan. This modification ensures numerical stability, smoothness and differentiability, enabling integration into gradient-based optimization and
faster computation via the Sinkhorn algorithm \cite{cuturi2013sinkhorn}. We provide a detailed introduction to the Sinkhorn algorithm in the Appendix \ref{supsec:BG}.

\section{Method}

\subsection{Motivation}
We rethink diffusion based dataset distillation methods, with a focus on establishing efficient training-free guidance during the sampling process. Motivated by applications of optimal transport theory in machine learning \cite{arjovsky2017wasserstein,kocc2025domain}, we formally propose \textbf{\Cref{thm:risk_bound_paper}}.
\begin{theorem}
\label{thm:risk_bound_paper}
Let $\mathcal{T}$ and $\mathcal{S}$ denote the target and surrogate datasets, respectively, with $\theta_{\mathcal{T}}^*$ and $\theta_{\mathcal{S}}^*$ being their optimal parameters. Define the target risk as: $R_{\mathcal{T}}(\theta) = \mathbb{E}_{(x,y) \sim \mathcal{T}} \left[ \ell(x, y, \theta) \right],$ where $\ell(\cdot)$ is an $L$-Lipschitz continuous evaluation function. Under semantic class alignment (i.e., no label mismatch), consider the marginal sample distributions $P_{\mathcal{T}}$ and $P_{\mathcal{S}}$ with optimal transport distance: $W(P_{\mathcal{T}}, P_{\mathcal{S}}) = \inf_{\gamma \in \Gamma(P_{\mathcal{T}}, P_{\mathcal{S}})} \mathbb{E}_{(x_{\mathcal{T}}, x_{\mathcal{S}}) \sim \gamma} \left[ d(x_{\mathcal{T}}, x_{\mathcal{S}}) \right],$
where $\Gamma(P_{\mathcal{T}}, P_{\mathcal{S}})$ is the set of all couplings between the distributions, and $d(\cdot,\cdot)$ is a distance metric on the sample space. Then the risk discrepancy satisfies:
\begin{equation}
\left| R_{\mathcal{T}}(\theta_{\mathcal{T}}^*) - R_{\mathcal{T}}(\theta_{\mathcal{S}}^*) \right| \leq 2L \cdot W(P_{\mathcal{T}}, P_{\mathcal{S}}).
\label{eq:risk_bound}
\end{equation}
\end{theorem}
We provide detailed proofs and analyses in the Appendix \ref{supsec:proof}. This insight motivates us to decouple \textbf{Semantic Matching} and \textbf{Distribution Matching} as the core objectives of dataset distillation.
We propose Dual Matching Guided Diffusion (DMGD), a training-free framework for dataset distillation that synergistically coordinates these objectives. Our proposed framework is illustrated in \Cref{fig:DMGD}. \par

\subsection{Semantic Matching}
Due to the informational redundancy in sample dimensions, direct sample guidance fails to distill representative semantic information. Previous work has demonstrated that conditional likelihood optimization constitutes an effective approach for extracting semantic information \cite{yin2023squeeze}. Our insight is that diffusion models can serve as zero-shot classifiers, eliminating the need to train additional classifiers on the target dataset. From this perspective, we introduce Classifier-Free Guidance (CFG) theory \cite{ho2022classifier} as \textbf{\Cref{lemma1_paper}} to establish that diffusion models efficiently approximate conditional log-likelihood.
\begin{lemma}[Classifier-Free Guidance \cite{ho2022classifier}]  
\label{lemma1_paper}
Consider a noise prediction network $\boldsymbol{\epsilon}_\theta(\boldsymbol{z}_t, t, y)$, where $\boldsymbol{z}_t$ denotes the representation of an original sample $\boldsymbol{x}$ at timestep $t$, and $y$ is a label. Assuming the $\boldsymbol{\epsilon}$ models both the conditional generative distribution $p(\boldsymbol{z}_t | y)$ and the unconditional distribution $p(\boldsymbol{z}_t)$, the gradient of the conditional log-likelihood $\log p(y|\boldsymbol{z}_t)$ with respect to $\boldsymbol{z}_t$ can be implicitly approximated by the difference between the network's conditional and unconditional outputs:  
\begin{equation}
    \nabla_{\boldsymbol{z}_t} \log p(y | \boldsymbol{z}_t) \approx \omega \, \Big( \boldsymbol{\epsilon}_\theta(\boldsymbol{z}_t, t, \emptyset) - \boldsymbol{\epsilon}_\theta(\boldsymbol{z}_t, t, y) \Big) 
\end{equation}
Here, $\omega$ denotes a scalar guidance scale, and $\boldsymbol{\epsilon}_\theta(\boldsymbol{z}_t, t, \emptyset)$ represents the network's unconditional output (i.e., without a specified class label).  
\end{lemma}

Based on \textbf{\Cref{lemma1_paper}}, we can achieve semantic matching via classifier-free guided conditional generation. Notably, while diffusion models provide conditional generative capacity for semantic alignment, they tend to over-sample high-density regions of the conditional distribution \cite{um2025minority}. This compromises the diversity of surrogate datasets while amplifying the effects of distribution shift.
\paragraph{Dynamic Semantic Matching for Diversity-Enhanced} Inspired by the diversity enhancement of diffusion models \cite{sadat2023cads}, we reveal that introducing slight perturbations during the sampling process does not disrupt semantic alignment. Consequently, we reframe semantic matching from a static into a dynamic guidance process. Based on a key observation of the properties of the diffusion sampling process\cite{yu2023freedom}, we partition the semantic guidance into three distinct stages: stochastic exploration ($t\geq45$), dynamic soft label guidance ($t\in[25,45]$), and semantic refinement ($t\leq25$). The dynamic guidance process is illustrated in \Cref{fig:DMGD}. Further implementation details are available in the Appendix \ref{suppsec:ID}. We provide a theoretical analysis and a deliberate design for the dynamic soft-label guidance \cite{sadat2023cads}.
\begin{proposition}
\label{prop:ddim_deterministic_modulation_paper}
Given a single step sampling process (such as DDIM) based on $\boldsymbol{\epsilon}_\theta$ to update $z_{t-1}^{(0)}$ using condition $y$, consider a dynamic label $\hat{y}_t = y + \delta_t$ where $\delta_t$ is a time-dependent vector. The modified sampling step admits the first-order approximation:
\begin{equation}
    z_{t-1} \approx z_{t-1}^{(0)} + \Lambda_t(\delta_t)
\end{equation}
where the condition shift operator $\Lambda_t$ is defined as:
$\Lambda_t(\delta_t) = c_t \cdot \bigl(\nabla_y \boldsymbol{\epsilon}_\theta(z_t, t, y)\bigr)^\top \delta_t$ with $c_t = \sqrt{1-\alpha_{t-1}} - \sqrt{\alpha_{t-1}} \cdot \sqrt{1-\alpha_t}/\sqrt{\alpha_t}$ as the intrinsic time-scaling factor.
\end{proposition}

 According to \textbf{\Cref{prop:ddim_deterministic_modulation_paper}}, the dynamic label is equivalent to introducing an additional shift term into the sampling dynamics. This enables us to enhance the coverage of data modes and improve the diversity by designing dynamic labels. Motivated by the concept of soft labels \cite{sadat2023cads,yin2023squeeze}, we propose a label diffusion process to construct dynamic soft label vectors at timestep $t$. Given a label encoder $f_{Y}$ and target label $y$, the dynamic soft label vector is defined as:
\begin{equation}
    \widetilde{f_Y}(y)=\sqrt{\sigma_t}f_Y(y)+(1-\sqrt{\sigma_t})(\beta_sf_Y(y^\star)+\beta_n n)
\label{equ:DSL}
\end{equation}

Where, $\beta$ is the modulation coefficient and $\sigma_t$ represents a time-dependent scheduling. $n$ is an anisotropic Gaussian noise term, aiding the sampling process to escape local modes and more fully explore the data distribution. $y^\star$ is a randomly chosen label, which induces a deterministic shift towards class boundaries to generate more informative samples. To ensure representative semantic matching, we rescale the dynamic soft label vector to align with the mean and standard deviation of the original label vectors. The final soft label-guided formula that we adopted is as follows:
\begin{equation}
\hat{\boldsymbol{\epsilon}}_\theta (z_t,t,\widetilde{y}_t)= (1+\omega)\boldsymbol{\epsilon}_\theta({z}_t, t, \widetilde{y}_t)-\omega\boldsymbol{\epsilon}_\theta({z}_t, t,\emptyset)
\end{equation}
 For compactness, we replace $\widetilde{f_Y}(y)$ with $\widetilde{y}$. We can define the dynamic soft label denoising process as $ z_{t-1} = D_\theta(z_t,t,\widetilde{y}).$ Taking advantage of dynamic guidance, we achieve a thorough exploration of the distribution space while ensuring semantic consistency, thus establishing a robust foundation for distributional alignment.\par
\subsection{Distribution Matching}
Furthermore, we aim to explore how to construct an effective guidance loss for the distribution alignment. 
Traditional distribution matching methods, such as mean matching \cite{zhao2023dataset}, often overlook inter-sample relationships and the distribution structure. Consequently, we propose an optimal transport guided objective, which achieves distribution alignment by optimizing the optimal transport distance between the target dataset and the surrogate dataset.
\begin{equation}
\mathop{\arg\min}\limits_{\mathcal{S}} W(P_\mathcal{T}, P_\mathcal{S}) =  \mathop{\arg\min}\limits_{\mathcal{S}}\min_{\gamma\in \Gamma(P_\mathcal{T}, P_\mathcal{S})} \sum_{i,j} \gamma_{ij} \cdot \mathbf{C}_{ij} 
\end{equation}

Here, $\mathbf{C}_{ij}$ denotes a cost metric. As noted in previous work \cite{cazenavette2023generalizing}, defining the metric in a high-information representation space is a more efficient and effective choice. Thus, we utilize the latent space of the diffusion model combined with hyperspherical projection as the distribution space, and the Euclidean distance as the distance metric. We use the Sinkhorn algorithm \cite{cuturi2013sinkhorn} to compute entropy-regularized optimal transport ($W_{\varepsilon}$), and the final guidance term is expressed as:
\begin{equation}
\mathcal{L}_{\text{OT}}(P_\mathcal{S}^t,P_\mathcal{T})=W_{\varepsilon}(P^t_\mathcal{S}, P_\mathcal{T})=\langle \gamma^{*}, \mathbf{C} \rangle
\end{equation}

$\gamma^{*}$ is the optimal transport plan. We employ the training-free guidance technique \cite{yu2023freedom} to embed the OT guided loss into the diffusion model framework. By \Cref{TFG_paper}, we have the following guided sampling process:
\begin{equation}
\begin{aligned}
    z^i_{t-1} & =D_\theta(z^i_t)-\rho_t\nabla_{z^i_t} \mathcal{L}_{\text{OT}}(P_\mathcal{S}^t,P_\mathcal{T}) \\&=D_\theta(z^i_t)-\rho_t\nabla_{z^i_t}\sum_{j} \gamma^{*}_{ij}\cdot \mathbf{C}_{ij}
\end{aligned}
\end{equation}
Intuitively, this loss encourages samples in the surrogate dataset to shift toward the nearest misaligned regions of the target distribution, thereby achieving alignment with the complete distribution structure. However, applying the optimal transport loss to the diffusion based dataset distillation framework still poses challenges: 1) For large-scale target datasets, the computational complexity of optimal transport becomes prohibitively expensive. 2) In high IPC (Instances Per Class) setting, memory constraints preclude end-to-end optimization due to excessive resource demands. To address these issues, we propose two enhanced strategies. 

\paragraph{Distribution Approximate Matching. }
 When processing large-scale target datasets, optimal transport iterations become computationally infeasible due to memory and time complexity constraints. To address this challenge, we can first employ a smaller discrete approximation distribution $\widetilde{P}_{\mathcal{T}}$ that preserves the essential geometric  properties of the original target distribution $P_{\mathcal{T}}$. Building upon \Cref{thm:risk_bound_paper}, we derive the following corollary:
\begin{corollary}
\label{cor:approx_bound_paper}
Under the conditions of \Cref{thm:risk_bound_paper}, consider an approximate distribution $\widetilde{P}_{\mathcal{T}}$ satisfying $W(\widetilde{P}_{\mathcal{T}}, P_{\mathcal{T}}) \leq \epsilon$ for small $\epsilon > 0$. Assuming the distance metric satisfies the triangle inequality and distributions lie in a Polish space. the risk discrepancy is bounded by:
\begin{equation}
\left| R_{\mathcal{T}}(\theta_{\mathcal{T}}^*) - R_{\mathcal{T}}(\theta_{\mathcal{S}}^*) \right| \leq 2L \cdot \left( W(P_{\mathcal{S}}, \widetilde{P}_{\mathcal{T}}) + W(P_{\mathcal{T}}, \widetilde{P}_{\mathcal{T}}) \right)
\label{eq:approx_bound}
\end{equation}
\end{corollary}
\textbf{\Cref{cor:approx_bound_paper}} indicates that we can find an approximate distribution $\widetilde{P}_{\mathcal{T}}$ satisfying the property $W(\widetilde{P}_{\mathcal{T}}, P_{\mathcal{T}}) \leq \epsilon$ to simplify the computation of $W(P_{\mathcal{S}}, \widetilde{P}_{\mathcal{T}})$. To obtain $\widetilde{P}_{\mathcal{T}}$, we define the distribution approximation problem, also known as the optimal quantization problem \cite{gruber2004optimum}.
\begin{definition}
\label{def:discrete_approx_paper}
Given a target distribution $P_{\mathcal{T}}$, the discrete distribution approximation problem seeks to find a set of support points $\{x_i\}_{i=1}^N $
and corresponding mass coefficients $\{m_i\}_{i=1}^N$ with $m_i \geq 0$ and $\sum_{i=1}^N m_i = 1$ that minimize the optimal transport distance to $P_{\mathcal{T}}$. Formally, we solve:
\begin{equation}
\min_{\substack{\{x_i\} \subset \mathcal{X}, m_i}} W\left(P_{\mathcal{T}}, \widetilde{P}_{\mathcal{T}}\right)
\label{eq:approx_problem_paper}
\end{equation}
where $\widetilde{P}_{\mathcal{T}} = \sum_{i=1}^N m_i \delta_{x_i}$ is the discrete approximation.
\end{definition}

For the discrete distribution approximation problem, clustering algorithms have been proven to be methods with favorable convergence bounds \cite{canas2012learning}. Thus, We propose a class-wise K-means based approximation method. Let $\{C_i\}_{i=1}^K$ denote the clusters obtained by partitioning the target subset $\mathcal{T}_y$, where $\mathcal{T}_y$ is a specific class within the target dataset \(\mathcal{T}\). And $k_i \in \mathbb{R}^D$ is the centroid of the $i$-th cluster and $c_i = |C_i|$ denotes the cardinality of the $i$-th cluster. The discrete approximation $\widetilde{P}_{\mathcal{T}}$ defined as follows:
\begin{equation}
\widetilde{P}_{\mathcal{T}} = \sum_{i=1}^K m_i \delta_{k_i} \ \text{with} \  \mathcal{K} = \{k_i\}_{i=1}^K, \; m_i = \frac{c_i}{\sum_{j=1}^K c_j}
\end{equation}

 Mean matching method can be regarded as a special case of distribution approximation. \textbf{\Cref{prop2}} provides a theoretical analysis of the error between mean matching method and our proposed method.
\begin{proposition}
\label{prop2}
Let $\tilde{P}_{\mathcal{T}}^{(1)}$ denote the mean matching approximation of $P_{\mathcal{T}}$ defined by a Dirac measure $\delta_{\mu}$ concentrated at the mean $\mu$ of $P_{\mathcal{T}}$, and $\tilde{P}_{\mathcal{T}}^{(2)}$ denote the proposed approximation constructed with cluster count $K $. The Wasserstein distance satisfies:
\begin{equation}
W(P_{\mathcal{T}}, \tilde{P}_{\mathcal{T}}^{(2)}) \leq W(P_{\mathcal{T}}, \tilde{P}_{\mathcal{T}}^{(1)})
\label{eq:prop5_wasserstein}
\end{equation}

\end{proposition}
Intuitively, clustering provides a way to discover fine-grained patterns within clusters and uses them as support points. Compared to mean matching, this acquires more comprehensive distribution information. Some similar ideas has been discussed in prior work on dataset distillation \cite{liu2023dream,chan2025mgd}, but those approaches focused mainly on representative points without deeper exploration of the distributional structure. By incorporating optimal transport, we utilize mass coefficients to better align distribution structures and reduce distribution shift effects.
\paragraph{Greedy Progressive Matching. }From the perspective of greedy optimization, we propose a progressive alignment framework. When optimizing $z^i$, we freeze all $z^j$, where $ j < i $, thereby defining the partially optimized surrogate dataset distribution as $ P_{\mathcal{S}^t_{\left[ i \right]}} $. Our optimization objective can then be rewritten as:
\begin{equation}
\begin{aligned}
    z^i_{t-1}&=D_\theta(z^i_t)-\rho_t\nabla_{z^i_t} \mathcal{L}_{\text{OT}}(P_{\mathcal{S}^t_{\left[ i \right]}},P_\mathcal{T}) 
\end{aligned}
\end{equation}

Under progressive matching, the guidance term further aligns the currently generated surrogate sample with previously unaligned regions of the target distribution space. Meanwhile, freezing the earlier samples prevents the surrogate samples from over-concentrating toward the mean of target distribution, thereby enhancing diversity. 
\section{Experiment}
\begin{table*}[t]  
\centering
\setlength{\tabcolsep}{9pt}
\begin{tabular}{l|ccc|ccc}  
\toprule
          & \multicolumn{3}{c|}{ImageNet-Woof} & \multicolumn{3}{c}{ImageNet-Nette}\\
\midrule
IPC       & 10    & 20    & 50   & 10    & 20    & 50   \\  
\midrule
Random    &29.4$_{\pm 0.8}$&32.7$_{\pm 0.4}$ &  47.2$_{\pm 1.3}$  &  54.2$_{\pm 1.6}$     & 63.5$_{\pm 0.5}$      &   76.1$_{\pm 1.1}$\\


DM \cite{zhao2023dataset}        &  30.3$_{\pm 1.2}$  &35.2$_{\pm 0.6}$       &47.1$_{\pm 1.1}$       & 60.8$_{\pm 0.6}$   & 66.5$_{\pm 1.1}$      & 76.2$_{\pm 0.4}$      \\
\midrule
GLaD \cite{cazenavette2023generalizing}      &    32.9$_{\pm 0.9}$   &  -     &  -     & -      & -      &  -     \\

DiT \cite{peebles2023scalable}       & 34.7$_{\pm 0.5}$      & 41.1$_{\pm 0.8}$      & 49.3$_{\pm 0.2}$      & 59.1$_{\pm 0.7}$      & 64.8$_{\pm 1.2}$      & 73.3$_{\pm 0.9}$      \\
MiniMax \cite{gu2024efficient}   &   39.2$_{\pm 1.3}$    &  45.8$_{\pm 0.5}$     &    56.3$_{\pm 1.0}$   & 62.0$_{\pm 0.2}$      &  66.8$_{\pm 0.4}$     &  76.6$_{\pm 0.2}$     \\

$\text{MGD}^3$ \cite{chan2025mgd}   &  40.4$_{\pm 1.9}$     &  43.6$_{\pm 1.6}$     &  56.5$_{\pm 0.8}$     & 66.4$_{\pm 2.4}$      &  \underline{71.2}$_{\pm 0.5}$     &79.5$_{\pm 1.3}$       \\
\midrule
DiT \cite{peebles2023scalable} + \textbf{Ours}  &  \underline{40.8}$_{\pm 1.1}$ &   \underline{46.7}$_{\pm 1.4}$    &   \underline{60.1}$_{\pm 0.8}$    &   \underline{68.4}$_{\pm 0.2}$    &   \textbf{72.6}$_{\pm 0.6}$    &  \underline{80.6}$_{\pm 0.5}$     \\
\multicolumn{1}{c|}{$\Delta$}  &  \textcolor{red}{$+6.1$} &   \textcolor{red}{$+5.6$}    &   \textcolor{red}{$+10.8$}    &  \textcolor{red}{$+9.3$}     &  \textcolor{red}{$+7.8$}     & \textcolor{red}{$+7.3$}      \\
MiniMax \cite{gu2024efficient} + \textbf{Ours}  &  \textbf{42.4}$_{\pm 0.5}$ & \textbf{47.7}$_{\pm0.4}$       &  \textbf{60.8}$_{\pm 0.2}$     &   \textbf{68.7}$_{\pm 0.8}$    &  71.1$_{\pm 0.5}$     &\textbf{80.7}$_{\pm 0.8}$       \\
\multicolumn{1}{c|}{$\Delta$}  & \textcolor{red}{$+3.2$}  &   \textcolor{red}{$+1.9$}    &   \textcolor{red}{$+4.5$}  &  \textcolor{red}{$+6.7$}  &   \textcolor{red}{$+4.3$}  &  \textcolor{red}{$+4.1$}     \\
\midrule
Full dataset & \multicolumn{3}{c|}{\textcolor{gray}{$87.5_{\pm 0.5}$}} & \multicolumn{3}{c}{\textcolor{gray}{$94.6_{\pm 0.5}$}}       \\
\bottomrule
\end{tabular}
\caption{Performance comparison between our method and state-of-the-art methods across different ImageNet subsets, evaluated under the hard-label protocol. Results are reported as Top-1 accuracy on ResNet-10 with average pooling (Resnet10-AP). The best performance is highlighted in \textbf{bold}, while the second-best is \underline{underlined}. } 
\label{tab:imagenet_setup}  
\end{table*}
\begin{figure*}[htbp] 
    \centering 
    \begin{subfigure}[b]{0.24\textwidth} 
        \centering
        \includegraphics[width=\textwidth]{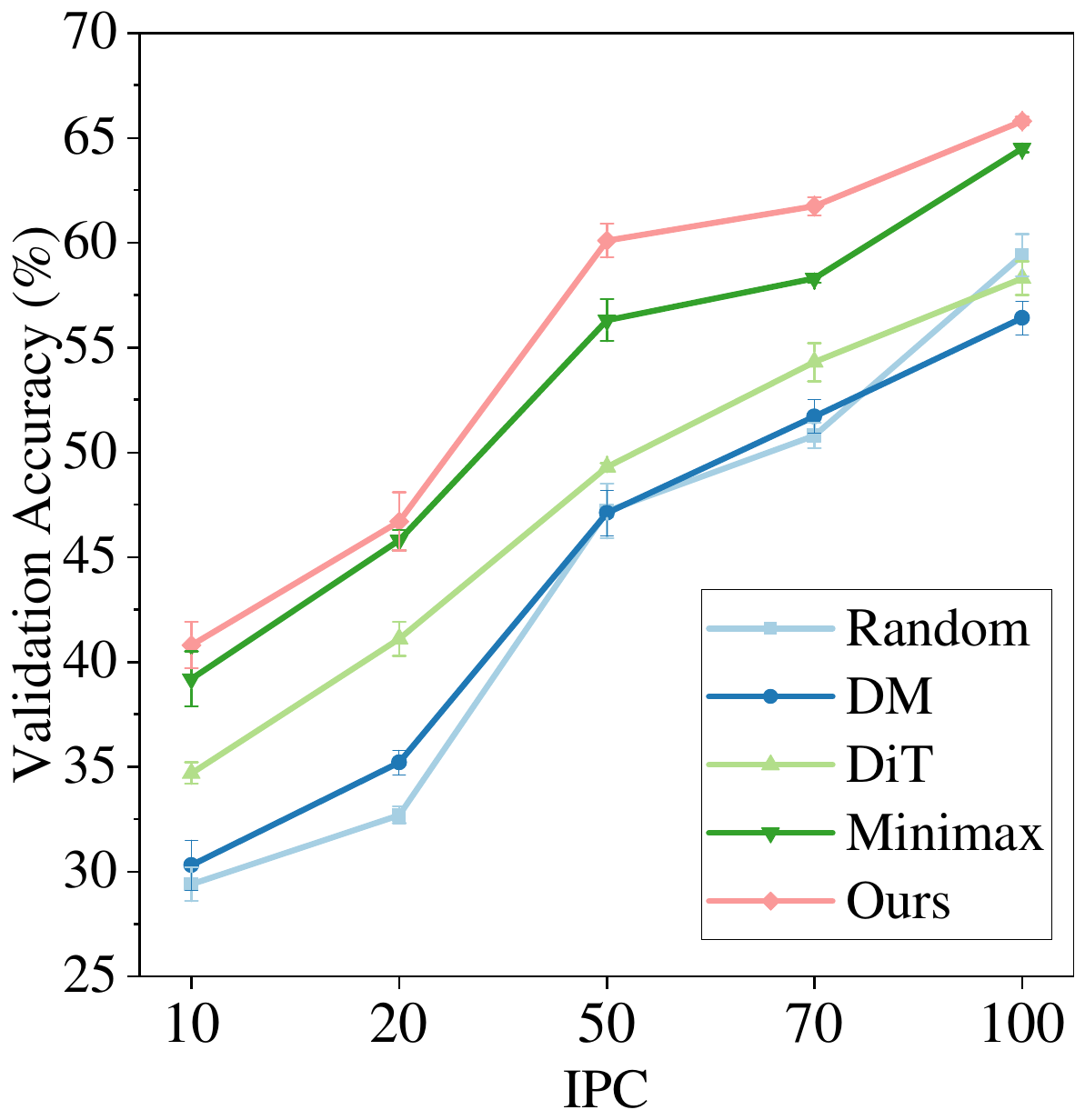} 
        \caption{} 
        \label{fig:sub1}
    \end{subfigure}%
    \begin{subfigure}[b]{0.24\textwidth}
        \centering
        \includegraphics[width=\textwidth]{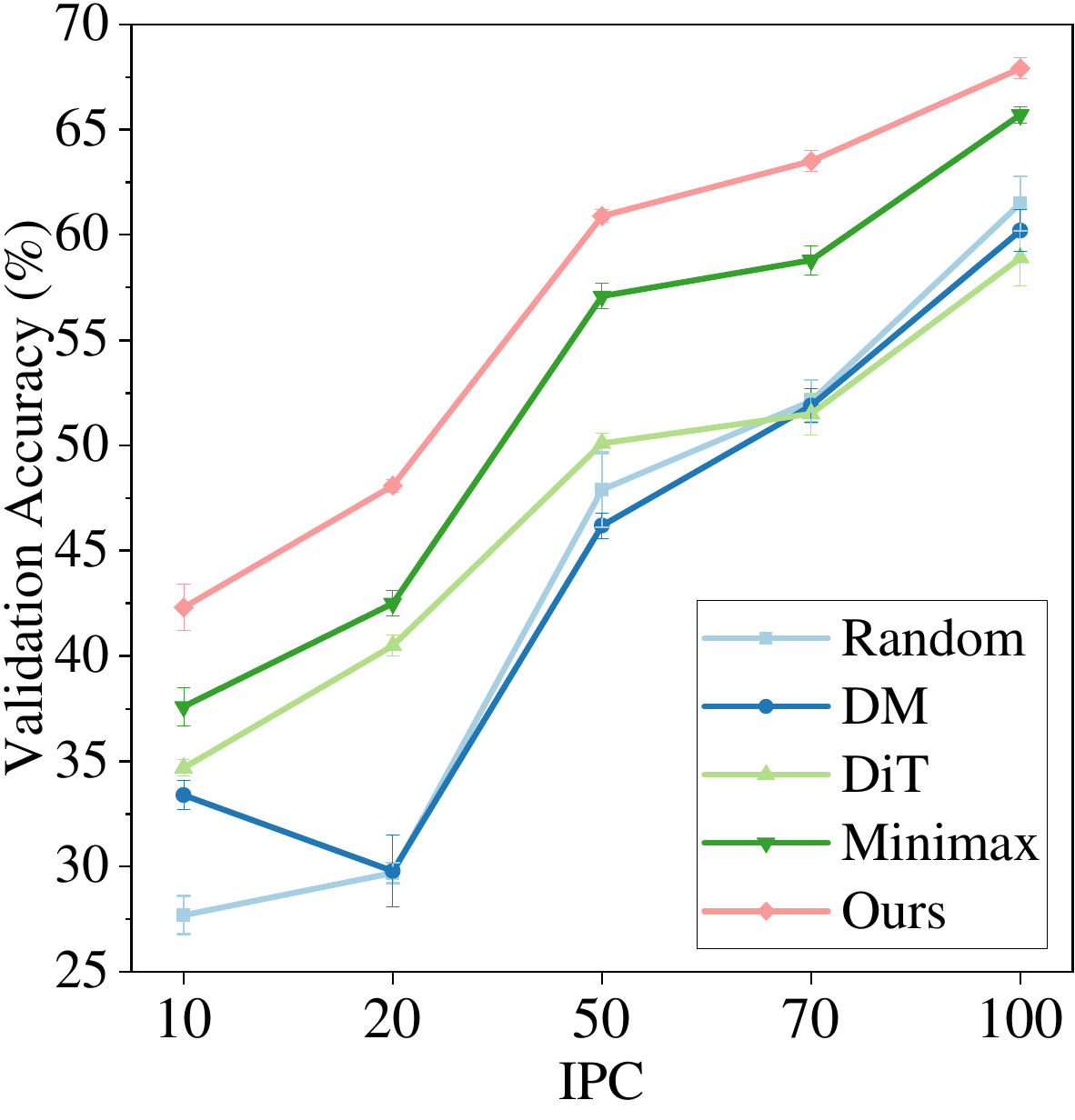}
        \caption{}
        \label{fig:sub2}
    \end{subfigure}%
    \begin{subfigure}[b]{0.24\textwidth}
        \centering
        \includegraphics[width=\textwidth]{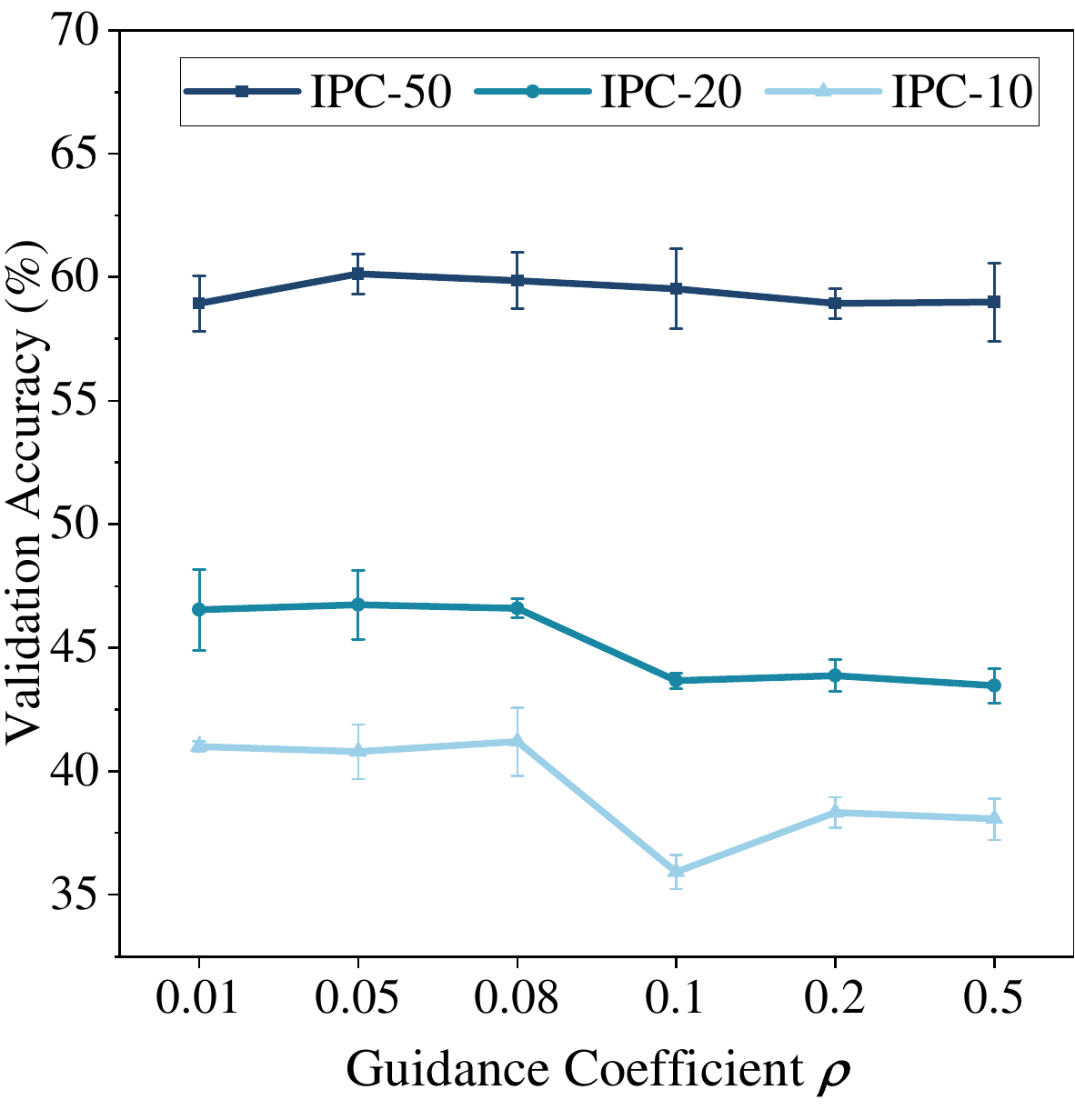}
        \caption{}
        \label{fig:sub3}
    \end{subfigure}%
    \begin{subfigure}[b]{0.24\textwidth}
        \centering
        \includegraphics[width=\textwidth]{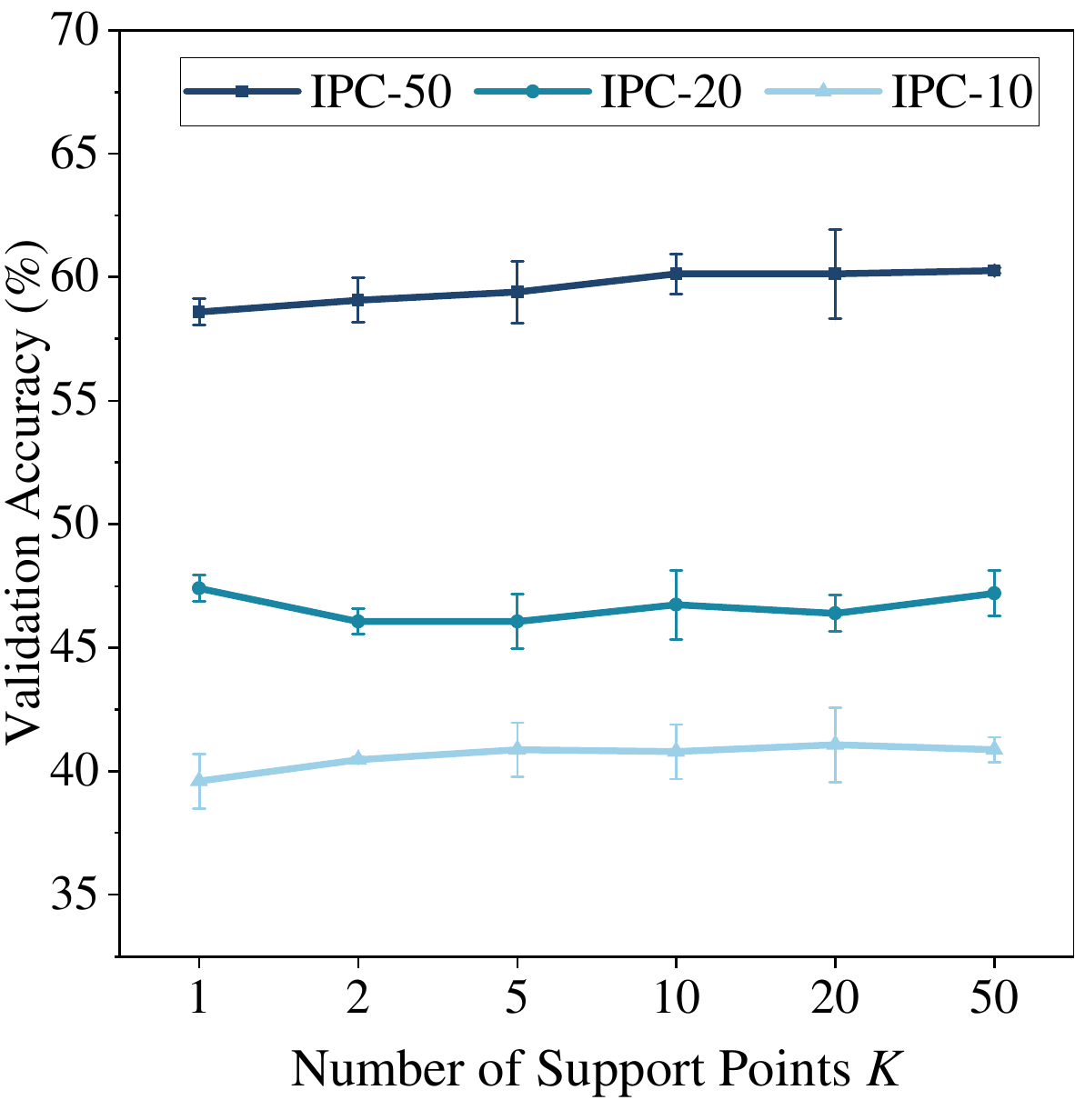}
        \caption{}
        \label{fig:sub4}
    \end{subfigure}%
    \caption{Evaluation results: (a-b) Evaluation of our method's performance across different architectures and higher IPC settings: Results are reported as Top-1 accuracy on (a) ResNet10-AP and (b) ResNet-18. (c-d) Evaluation of our method's performance under different hyperparameters: (c) distribution matching guidance coefficient $\rho$ and (d) number of support points $K$ for distribution approximation.} 
    \label{fig:four_figs}
\end{figure*}
\subsection{Experiment Setting}
\paragraph{Datasets and Evaluation Metric} We evaluate our proposed method using general high-resolution ($256\times 256$) dataset distillation benchmarks. Our evaluation datasets include ImageNet-1K \cite{deng2009imagenet} and its subsets, ImageNet-Woof and ImageNet-Nette. For evaluation, we follow the setup of \citet{gu2024efficient} by training classifier on the surrogate dataset and reporting the performance on the testset. Consistent with previous work, we adopt the hard-label evaluation protocol for ImageNet subsets and the soft-label evaluation protocol for the ImageNet-1K. We provide the results of other datasets in Appendix \ref{supsec:AR}.
\paragraph{Baseline} We examine state-of-the-art (SOTA) dataset distillation algorithms based on generative models, especially diffusion models, including GlaD \cite{cazenavette2023generalizing}, Minimax \cite{gu2024efficient}, D$^4$M \cite{Su_2024_CVPR}, and MGD$^3$ \cite{chan2025mgd}. We also incorporate the pre-trained DiT XL \cite{peebles2023scalable}, which represents the performance of directly using diffusion models for dataset distillation. Additionally, we include the distribution match (DM) method \cite{zhao2023dataset}. In the comparison on ImageNet-1k, we also examined other methods including SRe$^2$L \cite{yin2023squeeze} , G-VBSM \cite{shao2024generalized}, and RDED \cite{sun2024diversity}.
\paragraph{Implementation Details} We employ the pre-trained DiT \cite{peebles2023scalable} as our baseline model. For semantic matching, we configure the CFG scale $1+\omega=4$, with the modulation coefficient $ \beta_n=0.06$ and $ \beta_s=0.01$. For distribution matching, we set $K$ to the minimal IPC configuration of $10$. The regularization coefficient $\rho$ is set to $0.05$ for ImageNet-Woof and 0.5 for ImageNet-Nette, respectively. Crucially, we strategically confine distribution matching application exclusively to the temporal window $t \in[30,45]$. We adopt Sinkhorn algorithm configurations \cite{shi2024ot}: $\varepsilon=0.1$ with $5$ iterations. All experiments are conducted on a NVIDIA RTX 4090 GPU.
\subsection{Comparison with Other Methods}
\paragraph{Comparison on ImageNet subset}

Our DiT-based implementation demonstrates significant performance improvements across ImageNet subsets, as summarized in \Cref{tab:imagenet_setup}. On the challenging ImageNet-Woof dataset, our method achieves performance gains of $0.4\%$, $3.1\%$, and $3.6\%$ over the state-of-the-art MGD$^3$ under varying IPC configurations. The proposed diversity-enhanced distribution alignment mechanism exhibits increasing efficacy at higher IPC settings in this challenging subset. On ImageNet-Nette, we observe improvements of $2.0\%$, $1.4\%$, and $1.1\%$ over MGD$^3$. Our approach surpasses all competing models.\

We quantitatively demonstrate the plug-and-play capability of our model in \Cref{tab:imagenet_setup}. Deployment of our method yields significant performance gains over baseline models. On the Imagenet-Nette subset, it achieves average improvements of $8.1\%$ against the baseline DiT and $5.0\%$ against Minimax. When deployed on Minimax, our method achieves optimal performance across all setting. These validate the plug-and-play efficacy of our DMGD framework and suggest significant potential for broader applicability.

Furthermore, we conduct extensive evaluations of our approach across diverse evaluation architectures and higher IPC configurations (IPC-70 and IPC-100). As demonstrated in \Cref{fig:four_figs} (a) and (b), our method consistently surpasses baseline models across all architectures and IPC settings.
\paragraph{Comparison on ImageNet-1k}

To validate the scalability of our method in larger-scale datasets, we conducted comparative experiments under the soft-label evaluation protocol \cite{sun2024diversity} using DiT-based deployment on ImageNet-1K, with results detailed in \Cref{tab:imagenet_results}. Under IPC-10 settings, our method achieves  $4.3\%$ improvement over  RDED and  $2.0\%$ improvement over Minimax. Under IPC-50 settings, our method consistently attains state-of-the-art performance comparable to MGD$^3$. We also maintained the best performance on the larger model architecture ResNet-101.

\subsection{Hyperparameter Analysis}
We conduct experiments and sensitivity analyses on the hyperparameters of our proposed method. We conduct the evaluation on the ImageNet-Woof dataset and report the top-1 accuracy of ResNet10-AP. \Cref{fig:four_figs} (c) and (d) quantitatively illustrate our method  performance sensitivity to two critical hyperparameters of distribution matching. Additional experimental results examining others hyperparameters of semantic
matching are detailed in Appendix \ref{supsec:AR}.
\paragraph{Guidance Coefficient $\rho$.} \Cref{fig:four_figs} (c) delineates the impact of the guidance constant $\rho$ for distribution matching. At low IPC settings, larger $\rho$ values may induce performance degradation, whereas under high IPC settings, performance remains stable within an appropriate range of $\rho$. Based on evaluations, we use $\rho=0.05$ for ImageNet-Woof.
\paragraph{Number of Support Points $K$.}
\Cref{fig:four_figs} (d) reveals the influence of the number of support points $K$ for distribution approximation. We observe that under high IPC settings, excessively small $K$ values produce overly coarse approximations, leading to performance degradation. While larger 
$K$ values enhance accuracy, they incur computational overhead of optimal transport. Through rigorous performance-efficiency tradeoff analysis, We select $K=10$ to achieve an optimal balance between performance and efficiency.
\begin{table}[t]
\centering
\setlength{\tabcolsep}{4pt}
\begin{tabular}{l|cccc}
\toprule
\multirow{2}{*}{Method} & \multicolumn{2}{c}{Resnet-18}& \multicolumn{2}{c}{Resnet-101} \\ 
& IPC-10 & IPC-50& IPC-10 & IPC-50 \\
\midrule
SRe$^2$L \cite{yin2023squeeze}     &21.3$_{\pm0.6}$ &   46.8$_{\pm 0.2}$&30.9$_{\pm 0.1}$ & 60.8$_{\pm 0.5}$\\
G-VBSM\cite{shao2024generalized} &31.4$_{\pm0.5}$&51.8$_{\pm0.4}$&38.2$_{\pm0.4}$&63.7$_{\pm0.2}$\\
RDED \cite{sun2024diversity}      &42.0$_{\pm0.2}$ &   56.5$_{\pm 0.1}$ & \underline{48.3}$_{\pm 1.0}$ & 61.2$_{\pm 0.4}$   \\
D$^4$M \cite{Su_2024_CVPR}     &27.9$_{\pm{\leq1}}$ &   55.2$_{\pm \leq1}$ &34.2$_{\pm \leq1 }$& 63.4$_{\pm \leq1}$  \\
Minimax \cite{gu2024efficient}      &\underline{44.3}$_{\pm0.5}$ &   58.6$_{\pm 0.3}$&-&-   \\
MGD$^3$ \cite{chan2025mgd}      &- &   \underline{60.2}$_{\pm 0.1}$ &-& {\underline{67.7}$_{\pm{0.4}}$}  \\
\textbf{Ours}      &\textbf{46.3}$_{\pm0.8}$ &   \textbf{61.4}$_{\pm 0.6}$ & \textbf{50.6}$_{\pm 1.2}$ &\textbf{68.4}$_{\pm 0.4}$ \\
\midrule
Full dataset&\multicolumn{2}{c}{\textcolor{gray}{69.8}}&\multicolumn{2}{c}{\textcolor{gray}{81.9}}\\
\bottomrule
\end{tabular}
\caption{Performance comparison between our method and state-of-the-art methods on ImageNet-1k, evaluated under the soft-label protocol. Results are reported as Top-1 accuracy on ResNet-18 and ResNet-101. The best performance is highlighted in \textbf{bold}, while the second-best is \underline{underlined}. Missing values are due to the original paper not reporting them.}
\label{tab:imagenet_results}
\end{table}
\subsection{Ablation Study}
To evaluate the individual contributions of the proposed components, we conduct component-wise ablation studies assessing the dynamic guidance semantic matching (SM) and optimal transport guidance distribution matching (DM), with results presented in \Cref{tab:aS}. Our dynamic guidance semantic matching boost surrogate dataset diversity, achieving significant performance gains under high IPC settings. At low IPC settings, optimal transport guidance prioritizes distribution alignment, achieving exceptional performance and demonstrating efficacy in generating critical samples. Consequently, our DMGD framework attains overall best performance for dataset distillation. We present more detailed ablation study in the Appendix \ref{supsec:AR}.
\begin{table}[t]
\centering
\setlength{\tabcolsep}{12pt}
\begin{tabular}{l|cc|cc}
\toprule
IPC   & SM & DM & Woof&Nette \\
\midrule
10   &   \multirow{2}{*}{-}    & \multirow{2}{*}{-}    &34.7$_{\pm 0.5}$ &  59.1$_{\pm 0.7}$    \\
50  &             &                        &49.3$_{\pm0.2}$ &   73.3$_{\pm 0.9}$   \\
\midrule
10   &   \multirow{2}{*}{\checkmark}   & \multirow{2}{*}{-}    & 38.9$_{\pm1.2}$&\underline{67.1}$_{\pm0.5}$ \\
50  &            &      & 59.3$_{\pm0.4}$   & 79.7$_{\pm0.1}$  \\
\midrule
10   &  \multirow{2}{*}{\textbf{-}}  & \multirow{2}{*}{\checkmark}    & \textbf{41.6}$_{\pm1.1}$&66.8$_{\pm1.8}$ \\
50    &    &          & 56.8$_{\pm0.2}$& 76.7$_{\pm0.5}$ \\
\midrule
10    &   \multirow{2}{*}{\checkmark}   & \multirow{2}{*}{\checkmark}    &\underline{40.8}$_{\pm 1.1}$& \textbf{68.4}$_{\pm 0.2}$ \\
50  &    &  & \textbf{60.1}$_{\pm 0.8}$&\textbf{80.6}$_{\pm 0.5}$ \\
\bottomrule
\end{tabular}
\caption{Ablation study on the components of our method. Results are reported as Top-1 accuracy on ResNet10-AP. The best performance is highlighted in \textbf{bold}, while the second-best is \underline{underlined}.}
\label{tab:aS}
\end{table}

\subsection{Representativeness and Diversity Analysis}
    
\begin{figure}[t]
    \centering
    \includegraphics[width=1\linewidth]{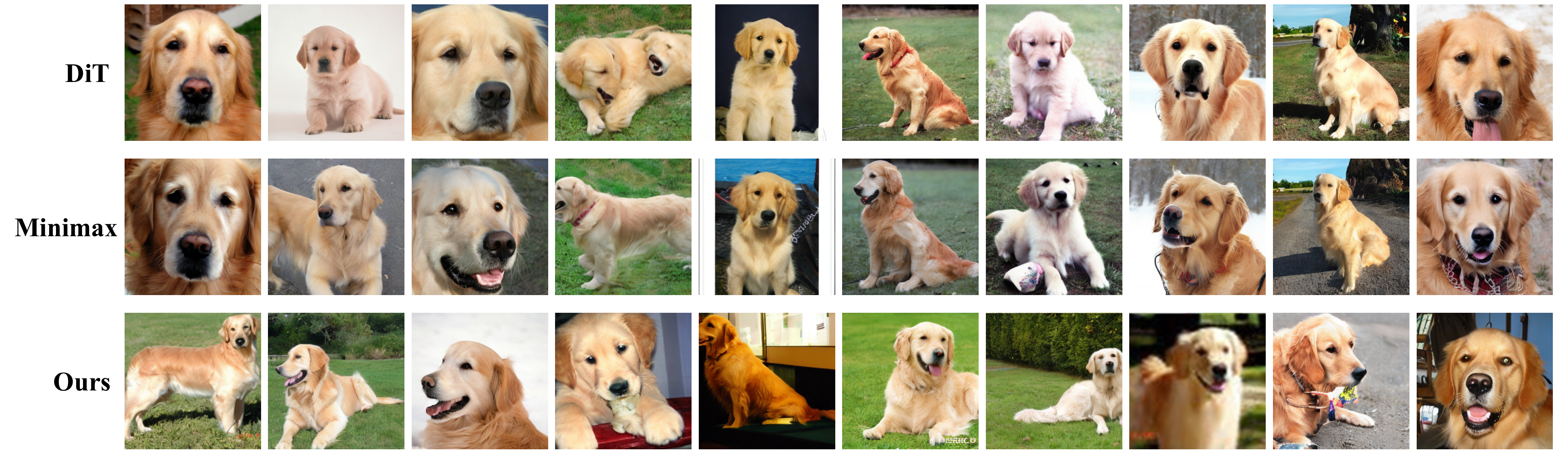}
    \caption{\textbf{Generated Samples Visualization:} the visual comparison of Golden Retriever in ImageNet-WOOF, we present the generated samples from different methods under the IPC-10 setting. The method names are
marked at the left of each row.}
    \label{fig:gsv1_paper}
\end{figure}
\begin{table}[h]
\centering
\setlength{\tabcolsep}{5pt}
\begin{tabular}{l|cccc}
\toprule
Method   & Cov.$\uparrow $ & OTDD$\downarrow$ & Diversity$\uparrow$ & FID$\downarrow$ \\
\midrule
DiT \cite{peebles2023scalable}      &25.4 &   142.2 &70.1 & \textbf{48.6}  \\
Minimax \cite{gu2024efficient}      &28.5 &   88.5&72.9 & 49.2   \\
\textbf{Ours}      &\textbf{30.7} &   \textbf{66.4}& \textbf{74.4}&48.8  \\
\bottomrule
\end{tabular}
\caption{Evaluation of representativeness and diversity of 10 classes each with 100 images in ImageNet-Woof. The evaluation metrics include coverage (cov.), optimal transport dataset distance (OTDD), diversity metric (Diversity), and FID. $\downarrow$ means lower is better and $\uparrow$ means higher is better.}
\label{tab:rd_results_paper}
\end{table}
We quantitatively evaluate representativeness and diversity in the feature space. Representativeness is measured via Coverage \cite{ferjad2020icml} and Dataset Distance \cite{alvarez2020geometric}. Diversity is computed as the mean minimum pairwise distance among all intra-class samples. We additionally report FID results to assess the visual quality. As summarized in \Cref{tab:rd_results_paper}, our method achieves significant improvements in representativeness metrics and diversity metrics. \Cref{fig:gsv1_paper} provides a  visualization to intuitively demonstrate the representativeness and diversity of our results. We also provide evaluation of other relevant metrics in the Appendix  \ref{supsec:AR}.
 \subsection{Computational Cost Analysis} Our method achieves SOTA performance across all datasets while introducing only marginal computational overhead during the sampling process. For instance, on a 10-class ImageNet subset, the MiniMax requires nearly 0.7 hours for fine-tuning. In contrast, our distribution approximation requires only 0.03 seconds per class. Under the IPC-50 setting, our method processes each image in 1.65 seconds, while baseline DiT requires 1.49 seconds. Crucially, the complete generation of a surrogate dataset for ImageNet-Woof with IPC-50 takes only 0.26 hours, demonstrating the efficiency superiority of our train-free framework.

\section{Conclusion}
\label{sec:conclusion}
In this work, we propose a dual matching guided diffusion (DMGD) framework, achieving efficient dataset distillation by introducing training free guidance during the sampling. Our insights encompass two improved matching objectives: a diversified semantic matching objective based on dynamic guidance, and a distribution matching objective based on optimal transport guidance. Theoretically grounded and experimentally validated across multiple datasets, our method achieves SOTA on all metrics. We conducted an analysis of each component and hyperparameter, validating their effectiveness through ablation study.\par
\section*{Acknowledgments}
This work was supported by the National Major Science and Technology Projects (the grant number 2022ZD0117000), the National Natural Science Foundation of China (the grant number 62202426), and the Shanghai Institute for Mathematics and Interdisciplinary Sciences (SIMIS) Fund (the grant number SIMIS-ID-2025-AD).

{
    \small
    \bibliographystyle{ieeenat_fullname}
    \bibliography{main}

\begin{thebibliography}{84}
\providecommand{\natexlab}[1]{#1}
\providecommand{\url}[1]{\texttt{#1}}
\expandafter\ifx\csname urlstyle\endcsname\relax
  \providecommand{\doi}[1]{doi: #1}\else
  \providecommand{\doi}{doi: \begingroup \urlstyle{rm}\Url}\fi

\bibitem[Achiam et~al.(2023)Achiam, Adler, Agarwal, Ahmad, Akkaya, Aleman, Almeida, Altenschmidt, Altman, Anadkat, et~al.]{achiam2023gpt}
Josh Achiam, Steven Adler, Sandhini Agarwal, Lama Ahmad, Ilge Akkaya, Florencia~Leoni Aleman, Diogo Almeida, Janko Altenschmidt, Sam Altman, Shyamal Anadkat, et~al.
\newblock Gpt-4 technical report.
\newblock \emph{arXiv preprint arXiv:2303.08774}, 2023.

\bibitem[Alvarez-Melis and Fusi(2020)]{alvarez2020geometric}
David Alvarez-Melis and Nicolo Fusi.
\newblock Geometric dataset distances via optimal transport.
\newblock \emph{Advances in Neural Information Processing Systems}, 33:\penalty0 21428--21439, 2020.

\bibitem[Arjovsky et~al.(2017)]{arjovsky2017wasserstein}
Martin Arjovsky et~al.
\newblock Wasserstein generative adversarial networks.
\newblock In \emph{International conference on machine learning}, pages 214--223. PMLR, 2017.

\bibitem[Canas et~al.(2012)]{canas2012learning}
Guillermo Canas et~al.
\newblock Learning probability measures with respect to optimal transport metrics.
\newblock \emph{Advances in neural information processing systems}, 25, 2012.

\bibitem[Cao et~al.(2025)Cao, Feng, Gong, Tian, Lu, Liu, and Wang]{cao2025dimension}
Hengyuan Cao, Yutong Feng, Biao Gong, Yijing Tian, Yunhong Lu, Chuang Liu, and Bin Wang.
\newblock Dimension-reduction attack! video generative models are experts on controllable image synthesis.
\newblock \emph{arXiv preprint arXiv:2505.23325}, 2025.

\bibitem[Cazenavette et~al.(2022)Cazenavette, Wang, Torralba, Efros, and Zhu]{cazenavette2022dataset}
George Cazenavette, Tongzhou Wang, Antonio Torralba, Alexei~A Efros, and Jun-Yan Zhu.
\newblock Dataset distillation by matching training trajectories.
\newblock In \emph{Proceedings of the IEEE/CVF Conference on Computer Vision and Pattern Recognition}, pages 4750--4759, 2022.

\bibitem[Cazenavette et~al.(2023)Cazenavette, Wang, Torralba, Efros, and Zhu]{cazenavette2023generalizing}
George Cazenavette, Tongzhou Wang, Antonio Torralba, Alexei~A Efros, and Jun-Yan Zhu.
\newblock Generalizing dataset distillation via deep generative prior.
\newblock In \emph{Proceedings of the IEEE/CVF Conference on Computer Vision and Pattern Recognition}, pages 3739--3748, 2023.

\bibitem[Chan-Santiago et~al.(2025)Chan-Santiago, Tirupattur, Nayak, Liu, and Shah]{chan2025mgd}
Jeffrey~A Chan-Santiago, Praveen Tirupattur, Gaurav~Kumar Nayak, Gaowen Liu, and Mubarak Shah.
\newblock Mgd$^3$: Mode-guided dataset distillation using diffusion models.
\newblock \emph{arXiv preprint arXiv:2505.18963}, 2025.

\bibitem[Chen et~al.(2025{\natexlab{a}})Chen, Yan, Chen, Cen, Wang, Ma, Zhen, Qian, Lu, and Gan]{chen2025rapverse}
Jiaben Chen, Xin Yan, Yihang Chen, Siyuan Cen, Zixin Wang, Qinwei Ma, Haoyu Zhen, Kaizhi Qian, Lie Lu, and Chuang Gan.
\newblock Rapverse: Coherent vocals and whole-body motion generation from text.
\newblock In \emph{Proceedings of the IEEE/CVF International Conference on Computer Vision}, pages 10097--10107, 2025{\natexlab{a}}.

\bibitem[Chen et~al.(2025{\natexlab{b}})Chen, Du, Huang, Wang, Zhang, and Wang]{chen2025influence}
Mingyang Chen, Jiawei Du, Bo Huang, Yi Wang, Xiaobo Zhang, and Wei Wang.
\newblock Influence-guided diffusion for dataset distillation.
\newblock In \emph{The Thirteenth International Conference on Learning Representations}, 2025{\natexlab{b}}.

\bibitem[Courty et~al.(2016)Courty, Flamary, Tuia, and Rakotomamonjy]{courty2016optimal}
Nicolas Courty, R{\'e}mi Flamary, Devis Tuia, and Alain Rakotomamonjy.
\newblock Optimal transport for domain adaptation.
\newblock \emph{IEEE transactions on pattern analysis and machine intelligence}, 39\penalty0 (9):\penalty0 1853--1865, 2016.

\bibitem[Croitoru et~al.(2023)Croitoru, Hondru, Ionescu, and Shah]{croitoru2023diffusion}
Florinel-Alin Croitoru, Vlad Hondru, Radu~Tudor Ionescu, and Mubarak Shah.
\newblock Diffusion models in vision: A survey.
\newblock \emph{IEEE transactions on pattern analysis and machine intelligence}, 45\penalty0 (9):\penalty0 10850--10869, 2023.

\bibitem[Cui et~al.(2023)Cui, Wang, Si, and Hsieh]{cui2023scaling}
Justin Cui, Ruochen Wang, Si Si, and Cho-Jui Hsieh.
\newblock Scaling up dataset distillation to imagenet-1k with constant memory.
\newblock In \emph{International Conference on Machine Learning}, pages 6565--6590. PMLR, 2023.

\bibitem[Cui et~al.(2025{\natexlab{a}})Cui, Qin, Zhou, Li, and Li]{cui2025optical}
Xiao Cui, Yulei Qin, Wengang Zhou, Hongsheng Li, and Houqiang Li.
\newblock Optical: Leveraging optimal transport for contribution allocation in dataset distillation.
\newblock In \emph{Proceedings of the Computer Vision and Pattern Recognition Conference}, pages 15245--15254, 2025{\natexlab{a}}.

\bibitem[Cui et~al.(2025{\natexlab{b}})Cui, Qin, Zhou, Li, and Li]{cui2025optimizing}
Xiao Cui, Yulei Qin, Wengang Zhou, Hongsheng Li, and Houqiang Li.
\newblock Optimizing distributional geometry alignment with optimal transport for generative dataset distillation.
\newblock \emph{arXiv preprint arXiv:2512.00308}, 2025{\natexlab{b}}.

\bibitem[Cuturi(2013)]{cuturi2013sinkhorn}
Marco Cuturi.
\newblock Sinkhorn distances: Lightspeed computation of optimal transport.
\newblock \emph{Advances in neural information processing systems}, 26, 2013.

\bibitem[Deng et~al.(2009)Deng, Dong, Socher, Li, Li, and Fei-Fei]{deng2009imagenet}
Jia Deng, Wei Dong, Richard Socher, Li-Jia Li, Kai Li, and Li Fei-Fei.
\newblock Imagenet: A large-scale hierarchical image database.
\newblock In \emph{2009 IEEE conference on computer vision and pattern recognition}, pages 248--255. Ieee, 2009.

\bibitem[Du et~al.(2023)Du, Shi, and Zhou]{du2023sequential}
Jiawei Du, Qin Shi, and Joey~Tianyi Zhou.
\newblock Sequential subset matching for dataset distillation.
\newblock \emph{Advances in Neural Information Processing Systems}, 36:\penalty0 67487--67504, 2023.

\bibitem[Feydy et~al.(2019)Feydy, S{\'e}journ{\'e}, Vialard, Amari, Trouv{\'e}, and Peyr{\'e}]{feydy2019interpolating}
Jean Feydy, Thibault S{\'e}journ{\'e}, Fran{\c{c}}ois-Xavier Vialard, Shun-ichi Amari, Alain Trouv{\'e}, and Gabriel Peyr{\'e}.
\newblock Interpolating between optimal transport and mmd using sinkhorn divergences.
\newblock In \emph{The 22nd international conference on artificial intelligence and statistics}, pages 2681--2690. PMLR, 2019.

\bibitem[Goodfellow et~al.(2014)Goodfellow, Pouget-Abadie, Mirza, Xu, Warde-Farley, Ozair, Courville, and Bengio]{goodfellow2014generative}
Ian~J Goodfellow, Jean Pouget-Abadie, Mehdi Mirza, Bing Xu, David Warde-Farley, Sherjil Ozair, Aaron Courville, and Yoshua Bengio.
\newblock Generative adversarial nets.
\newblock \emph{Advances in neural information processing systems}, 27, 2014.

\bibitem[Gruber(2004)]{gruber2004optimum}
Peter~M Gruber.
\newblock Optimum quantization and its applications.
\newblock \emph{Advances in Mathematics}, 186\penalty0 (2):\penalty0 456--497, 2004.

\bibitem[Gu et~al.(2024)Gu, Vahidian, Kungurtsev, Wang, Jiang, You, and Chen]{gu2024efficient}
Jianyang Gu, Saeed Vahidian, Vyacheslav Kungurtsev, Haonan Wang, Wei Jiang, Yang You, and Yiran Chen.
\newblock Efficient dataset distillation via minimax diffusion.
\newblock In \emph{Proceedings of the IEEE/CVF Conference on Computer Vision and Pattern Recognition}, pages 15793--15803, 2024.

\bibitem[Guo et~al.(2023)Guo, Wang, Cazenavette, Li, Zhang, and You]{guo2023towards}
Ziyao Guo, Kai Wang, George Cazenavette, Hui Li, Kaipeng Zhang, and Yang You.
\newblock Towards lossless dataset distillation via difficulty-aligned trajectory matching.
\newblock \emph{arXiv preprint arXiv:2310.05773}, 2023.

\bibitem[He et~al.(2016)He, Zhang, Ren, and Sun]{he2016deep}
Kaiming He, Xiangyu Zhang, Shaoqing Ren, and Jian Sun.
\newblock Deep residual learning for image recognition.
\newblock In \emph{Proceedings of the IEEE conference on computer vision and pattern recognition}, pages 770--778, 2016.

\bibitem[Ho and Salimans(2022)]{ho2022classifier}
Jonathan Ho and Tim Salimans.
\newblock Classifier-free diffusion guidance.
\newblock \emph{arXiv preprint arXiv:2207.12598}, 2022.

\bibitem[Ho et~al.(2020)]{ho2020denoising}
Jonathan Ho et~al.
\newblock Denoising diffusion probabilistic models.
\newblock \emph{Advances in neural information processing systems}, 33:\penalty0 6840--6851, 2020.

\bibitem[Howard(2019)]{Howard_Imagenette_2019}
Jeremy Howard.
\newblock Imagenette: A smaller subset of 10 easily classified classes from imagenet, 2019.

\bibitem[Kim et~al.(2022)Kim, Kim, Oh, Yun, Song, Jeong, Ha, and Song]{kim2022dataset}
Jang-Hyun Kim, Jinuk Kim, Seong~Joon Oh, Sangdoo Yun, Hwanjun Song, Joonhyun Jeong, Jung-Woo Ha, and Hyun~Oh Song.
\newblock Dataset condensation via efficient synthetic-data parameterization.
\newblock In \emph{International Conference on Machine Learning}, pages 11102--11118. PMLR, 2022.

\bibitem[Kingma and Welling(2013)]{kingma2013auto}
Diederik~P Kingma and Max Welling.
\newblock Auto-encoding variational bayes.
\newblock \emph{arXiv preprint arXiv:1312.6114}, 2013.

\bibitem[Ko{\c{c}} et~al.(2025)Ko{\c{c}}, Soen, Chiang, and Sugiyama]{kocc2025domain}
Okan Ko{\c{c}}, Alexander Soen, Chao-Kai Chiang, and Masashi Sugiyama.
\newblock Domain adaptation and entanglement: an optimal transport perspective.
\newblock \emph{arXiv preprint arXiv:2503.08155}, 2025.

\bibitem[Kungurtsev et~al.(2024)Kungurtsev, Peng, Gu, Vahidian, Quinn, Idlahcen, and Chen]{kungurtsev2024datasetdistillationprinciplesintegrating}
Vyacheslav Kungurtsev, Yuanfang Peng, Jianyang Gu, Saeed Vahidian, Anthony Quinn, Fadwa Idlahcen, and Yiran Chen.
\newblock Dataset distillation from first principles: Integrating core information extraction and purposeful learning, 2024.

\bibitem[Lei and Tao(2023)]{lei2023comprehensive}
Shiye Lei and Dacheng Tao.
\newblock A comprehensive survey of dataset distillation.
\newblock \emph{IEEE Transactions on Pattern Analysis and Machine Intelligence}, 46\penalty0 (1):\penalty0 17--32, 2023.

\bibitem[Lin et~al.(2025)Lin, Cen, Jiang, Karhade, Wang, Mitra, Ling, Huang, Liu, Chen, et~al.]{lin2025towards}
Zhiqiu Lin, Siyuan Cen, Daniel Jiang, Jay Karhade, Hewei Wang, Chancharik Mitra, Tiffany Ling, Yuhan Huang, Sifan Liu, Mingyu Chen, et~al.
\newblock Towards understanding camera motions in any video.
\newblock \emph{arXiv preprint arXiv:2504.15376}, 2025.

\bibitem[Liu et~al.(2024)Liu, Gu, Cao, Trinitis, and Schulz]{liu2024dataset}
Dai Liu, Jindong Gu, Hu Cao, Carsten Trinitis, and Martin Schulz.
\newblock Dataset distillation by automatic training trajectories.
\newblock In \emph{European Conference on Computer Vision}, pages 334--351. Springer, 2024.

\bibitem[Liu et~al.(2023{\natexlab{a}})Liu, Li, Xing, Dalal, Li, He, and Wang]{liu2023dataset}
Haoyang Liu, Yijiang Li, Tiancheng Xing, Vibhu Dalal, Luwei Li, Jingrui He, and Haohan Wang.
\newblock Dataset distillation via the wasserstein metric.
\newblock \emph{arXiv preprint arXiv:2311.18531}, 2023{\natexlab{a}}.

\bibitem[Liu et~al.(2022)Liu, Wang, Yang, Ye, and Wang]{liu2022dataset}
Songhua Liu, Kai Wang, Xingyi Yang, Jingwen Ye, and Xinchao Wang.
\newblock Dataset distillation via factorization.
\newblock \emph{Advances in neural information processing systems}, 35:\penalty0 1100--1113, 2022.

\bibitem[Liu et~al.(2023{\natexlab{b}})Liu, Gu, Wang, Zhu, Jiang, and You]{liu2023dream}
Yanqing Liu, Jianyang Gu, Kai Wang, Zheng Zhu, Wei Jiang, and Yang You.
\newblock Dream: Efficient dataset distillation by representative matching.
\newblock In \emph{Proceedings of the IEEE/CVF International Conference on Computer Vision}, pages 17314--17324, 2023{\natexlab{b}}.

\bibitem[Lu et~al.(2025{\natexlab{a}})Lu, Shao, Ding, Chen, Wu, Su, Yang, Zhang, Wang, Shi, et~al.]{lu2025discovery}
Renzhi Lu, Zonghe Shao, Yuemin Ding, Ruijuan Chen, Dongrui Wu, Housheng Su, Tao Yang, Fumin Zhang, Jun Wang, Yang Shi, et~al.
\newblock Discovery of the reward function for embodied reinforcement learning agents.
\newblock \emph{Nature Communications}, 16\penalty0 (1):\penalty0 11064, 2025{\natexlab{a}}.

\bibitem[Lu et~al.(2025{\natexlab{b}})Lu, Wang, Cao, Wang, Xu, and Zhang]{lu2025inpo}
Yunhong Lu, Qichao Wang, Hengyuan Cao, Xierui Wang, Xiaoyin Xu, and Min Zhang.
\newblock Inpo: Inversion preference optimization with reparametrized ddim for efficient diffusion model alignment.
\newblock In \emph{Proceedings of the Computer Vision and Pattern Recognition Conference}, pages 28629--28639, 2025{\natexlab{b}}.

\bibitem[Lu et~al.(2025{\natexlab{c}})Lu, Wang, Cao, Xu, and Zhang]{lu2025smoothed}
Yunhong Lu, Qichao Wang, Hengyuan Cao, Xiaoyin Xu, and Min Zhang.
\newblock Smoothed preference optimization via renoise inversion for aligning diffusion models with varied human preferences.
\newblock \emph{arXiv preprint arXiv:2506.02698}, 2025{\natexlab{c}}.

\bibitem[Lu et~al.(2025{\natexlab{d}})Lu, Zeng, Li, Ouyang, Wang, Cheng, Zhu, Cao, Zhang, Zhu, et~al.]{lu2025reward}
Yunhong Lu, Yanhong Zeng, Haobo Li, Hao Ouyang, Qiuyu Wang, Ka~Leong Cheng, Jiapeng Zhu, Hengyuan Cao, Zhipeng Zhang, Xing Zhu, et~al.
\newblock Reward forcing: Efficient streaming video generation with rewarded distribution matching distillation.
\newblock \emph{arXiv preprint arXiv:2512.04678}, 2025{\natexlab{d}}.

\bibitem[Maaten and Hinton(2008)]{maaten2008visualizing}
Laurens van~der Maaten and Geoffrey Hinton.
\newblock Visualizing data using t-sne.
\newblock \emph{Journal of machine learning research}, 9\penalty0 (Nov):\penalty0 2579--2605, 2008.

\bibitem[Montesuma et~al.(2024)]{montesuma2024recent}
Eduardo~Fernandes Montesuma et~al.
\newblock Recent advances in optimal transport for machine learning.
\newblock \emph{IEEE Transactions on Pattern Analysis and Machine Intelligence}, 2024.

\bibitem[Moser et~al.(2024)Moser, Raue, Palacio, Frolov, and Dengel]{moser2025unlockingdatasetdistillationdiffusion}
Brian~B Moser, Federico Raue, Sebastian Palacio, Stanislav Frolov, and Andreas Dengel.
\newblock Unlocking dataset distillation with diffusion models.
\newblock \emph{arXiv preprint arXiv:2403.03881}, 2024.

\bibitem[Naeem et~al.(2020)Naeem, Oh, Uh, Choi, and Yoo]{ferjad2020icml}
Muhammad~Ferjad Naeem, Seong~Joon Oh, Youngjung Uh, Yunjey Choi, and Jaejun Yoo.
\newblock Reliable fidelity and diversity metrics for generative models.
\newblock In \emph{International conference on machine learning}, pages 7176--7185. PMLR, 2020.

\bibitem[Omer(2020)]{Omer_fast-pytorch-kmeans_2020}
Sehban Omer.
\newblock {fast-pytorch-kmeans}, 2020.

\bibitem[Parmar et~al.(2022)Parmar, Zhang, and Zhu]{parmar2022aliased}
Gaurav Parmar, Richard Zhang, and Jun-Yan Zhu.
\newblock On aliased resizing and surprising subtleties in gan evaluation.
\newblock In \emph{Proceedings of the IEEE/CVF conference on computer vision and pattern recognition}, pages 11410--11420, 2022.

\bibitem[Peebles et~al.(2023)]{peebles2023scalable}
William Peebles et~al.
\newblock Scalable diffusion models with transformers.
\newblock In \emph{Proceedings of the IEEE/CVF international conference on computer vision}, pages 4195--4205, 2023.

\bibitem[Rombach et~al.(2022)Rombach, Blattmann, Lorenz, Esser, and Ommer]{rombach2022high}
Robin Rombach, Andreas Blattmann, Dominik Lorenz, Patrick Esser, and Bj{\"o}rn Ommer.
\newblock High-resolution image synthesis with latent diffusion models.
\newblock In \emph{Proceedings of the IEEE/CVF conference on computer vision and pattern recognition}, pages 10684--10695, 2022.

\bibitem[Sadat et~al.(2023)Sadat, Buhmann, Bradley, Hilliges, and Weber]{sadat2023cads}
Seyedmorteza Sadat, Jakob Buhmann, Derek Bradley, Otmar Hilliges, and Romann~M Weber.
\newblock Cads: Unleashing the diversity of diffusion models through condition-annealed sampling.
\newblock \emph{arXiv preprint arXiv:2310.17347}, 2023.

\bibitem[Schuhmann et~al.(2022)Schuhmann, Beaumont, Vencu, Gordon, Wightman, Cherti, Coombes, Katta, Mullis, Wortsman, et~al.]{schuhmann2022laion}
Christoph Schuhmann, Romain Beaumont, Richard Vencu, Cade Gordon, Ross Wightman, Mehdi Cherti, Theo Coombes, Aarush Katta, Clayton Mullis, Mitchell Wortsman, et~al.
\newblock Laion-5b: An open large-scale dataset for training next generation image-text models.
\newblock \emph{Advances in neural information processing systems}, 35:\penalty0 25278--25294, 2022.

\bibitem[Shao et~al.(2024{\natexlab{a}})Shao, Yin, Zhou, Zhang, and Shen]{shao2024generalized}
Shitong Shao, Zeyuan Yin, Muxin Zhou, Xindong Zhang, and Zhiqiang Shen.
\newblock Generalized large-scale data condensation via various backbone and statistical matching.
\newblock In \emph{Proceedings of the IEEE/CVF Conference on Computer Vision and Pattern Recognition}, pages 16709--16718, 2024{\natexlab{a}}.

\bibitem[Shao et~al.(2024{\natexlab{b}})Shao, Wang, Cao, Cai, You, and Lu]{shao2024novel}
Zonghe Shao, Qichao Wang, Yuzhe Cao, Defu Cai, Yang You, and Renzhi Lu.
\newblock A novel data-driven lstm-saf model for power systems transient stability assessment.
\newblock \emph{IEEE Transactions on Industrial Informatics}, 20\penalty0 (7):\penalty0 9083--9097, 2024{\natexlab{b}}.

\bibitem[Shen et~al.(2025)Shen, Sherif, Yin, and Shao]{shen2025delt}
Zhiqiang Shen, Ammar Sherif, Zeyuan Yin, and Shitong Shao.
\newblock Delt: A simple diversity-driven earlylate training for dataset distillation.
\newblock In \emph{Proceedings of the Computer Vision and Pattern Recognition Conference}, pages 4797--4806, 2025.

\bibitem[Shi et~al.(2024)]{shi2024ot}
Liangliang Shi et~al.
\newblock Ot-clip: Understanding and generalizing clip via optimal transport.
\newblock In \emph{Forty-first International Conference on Machine Learning}, 2024.

\bibitem[Shin et~al.(2025)Shin, Bae, Sim, Kang, and Moon]{shin2025distilling}
Donghyeok Shin, HeeSun Bae, Gyuwon Sim, Wanmo Kang, and Il-Chul Moon.
\newblock Distilling dataset into neural field.
\newblock \emph{arXiv preprint arXiv:2503.04835}, 2025.

\bibitem[Shin et~al.(2023)Shin, Bae, Shin, Joo, and Moon]{shin2023loss}
Seungjae Shin, Heesun Bae, Donghyeok Shin, Weonyoung Joo, and Il-Chul Moon.
\newblock Loss-curvature matching for dataset selection and condensation.
\newblock In \emph{International Conference on Artificial Intelligence and Statistics}, pages 8606--8628. PMLR, 2023.

\bibitem[Simonyan and Zisserman(2014)]{simonyan2014very}
Karen Simonyan and Andrew Zisserman.
\newblock Very deep convolutional networks for large-scale image recognition.
\newblock \emph{arXiv preprint arXiv:1409.1556}, 2014.

\bibitem[Song et~al.(2020)Song, Meng, and Ermon]{song2020denoising}
Jiaming Song, Chenlin Meng, and Stefano Ermon.
\newblock Denoising diffusion implicit models.
\newblock \emph{arXiv preprint arXiv:2010.02502}, 2020.

\bibitem[Su et~al.(2024)Su, Hou, Gao, Tian, and Tang]{Su_2024_CVPR}
Duo Su, Junjie Hou, Weizhi Gao, Yingjie Tian, and Bowen Tang.
\newblock D{\textasciicircum}4: Dataset distillation via disentangled diffusion model.
\newblock In \emph{Proceedings of the IEEE/CVF Conference on Computer Vision and Pattern Recognition (CVPR)}, pages 5809--5818, 2024.

\bibitem[Sun et~al.(2024)Sun, Shi, Yu, and Lin]{sun2024diversity}
Peng Sun, Bei Shi, Daiwei Yu, and Tao Lin.
\newblock On the diversity and realism of distilled dataset: An efficient dataset distillation paradigm.
\newblock In \emph{Proceedings of the IEEE/CVF Conference on Computer Vision and Pattern Recognition}, pages 9390--9399, 2024.

\bibitem[Tang and Jia(2020)]{tang2020discriminative}
Hui Tang and Kui Jia.
\newblock Discriminative adversarial domain adaptation.
\newblock In \emph{Proceedings of the AAAI conference on artificial intelligence}, pages 5940--5947, 2020.

\bibitem[Um and Ye(2025)]{um2025minority}
Soobin Um and Jong~Chul Ye.
\newblock Minority-focused text-to-image generation via prompt optimization.
\newblock In \emph{Proceedings of the Computer Vision and Pattern Recognition Conference}, pages 20926--20936, 2025.

\bibitem[Villani et~al.(2008)]{villani2008optimal}
C{\'e}dric Villani et~al.
\newblock \emph{Optimal transport: old and new}.
\newblock Springer, 2008.

\bibitem[Wang et~al.(2023)Wang, Gu, Zhou, Zhu, Jiang, and You]{wang2023dimdistillingdatasetgenerative}
Kai Wang, Jianyang Gu, Daquan Zhou, Zheng Zhu, Wei Jiang, and Yang You.
\newblock Dim: Distilling dataset into generative model, 2023.

\bibitem[Wang et~al.(2025)Wang, Yang, Liu, Sun, Hu, He, and Zhang]{wang2025dataset}
Shaobo Wang, Yicun Yang, Zhiyuan Liu, Chenghao Sun, Xuming Hu, Conghui He, and Linfeng Zhang.
\newblock Dataset distillation with neural characteristic function: A minmax perspective.
\newblock In \emph{Proceedings of the Computer Vision and Pattern Recognition Conference}, pages 25570--25580, 2025.

\bibitem[Wang et~al.(2020)Wang, Zhu, Torralba, and Efros]{wang2020datasetdistillation}
Tongzhou Wang, Jun-Yan Zhu, Antonio Torralba, and Alexei~A. Efros.
\newblock Dataset distillation, 2020.

\bibitem[Wang et~al.(2019)Wang, Li, Ye, Long, and Wang]{wang2019transferable}
Ximei Wang, Liang Li, Weirui Ye, Mingsheng Long, and Jianmin Wang.
\newblock Transferable attention for domain adaptation.
\newblock In \emph{Proceedings of the AAAI conference on artificial intelligence}, pages 5345--5352, 2019.

\bibitem[Welling(2009)]{welling2009herding}
Max Welling.
\newblock Herding dynamical weights to learn.
\newblock In \emph{Proceedings of the 26th annual international conference on machine learning}, pages 1121--1128, 2009.

\bibitem[Xiao and He(2024)]{xiao2024large}
Lingao Xiao and Yang He.
\newblock Are large-scale soft labels necessary for large-scale dataset distillation?
\newblock \emph{arXiv preprint arXiv:2410.15919}, 2024.

\bibitem[Xie et~al.(2023)Xie, Yao, Shi, Liu, Zhou, Liu, Li, and Li]{xie2023difffit}
Enze Xie, Lewei Yao, Han Shi, Zhili Liu, Daquan Zhou, Zhaoqiang Liu, Jiawei Li, and Zhenguo Li.
\newblock Difffit: Unlocking transferability of large diffusion models via simple parameter-efficient fine-tuning.
\newblock In \emph{Proceedings of the IEEE/CVF International Conference on Computer Vision}, pages 4230--4239, 2023.

\bibitem[Xue et~al.(2025)Xue, Li, Liu, Wang, Shen, and Wang]{xue2025towards}
Eric Xue, Yijiang Li, Haoyang Liu, Peiran Wang, Yifan Shen, and Haohan Wang.
\newblock Towards adversarially robust dataset distillation by curvature regularization.
\newblock In \emph{Proceedings of the AAAI Conference on Artificial Intelligence}, pages 9041--9049, 2025.

\bibitem[Yang et~al.(2024)Yang, Zhu, Deng, and Russakovsky]{yang2024datasetdistillationlearning}
William Yang, Ye Zhu, Zhiwei Deng, and Olga Russakovsky.
\newblock What is dataset distillation learning?, 2024.

\bibitem[Ye et~al.(2024)Ye, Lin, Han, Xu, Liu, Liang, Ma, Zou, and Ermon]{ye2024tfg}
Haotian Ye, Haowei Lin, Jiaqi Han, Minkai Xu, Sheng Liu, Yitao Liang, Jianzhu Ma, James~Y Zou, and Stefano Ermon.
\newblock Tfg: Unified training-free guidance for diffusion models.
\newblock \emph{Advances in Neural Information Processing Systems}, 37:\penalty0 22370--22417, 2024.

\bibitem[Yin et~al.(2023)]{yin2023squeeze}
Zeyuan Yin et~al.
\newblock Squeeze, recover and relabel: Dataset condensation at imagenet scale from a new perspective.
\newblock \emph{Advances in Neural Information Processing Systems}, 36:\penalty0 73582--73603, 2023.

\bibitem[Yu et~al.(2023)Yu, Wang, Zhao, Ghanem, and Zhang]{yu2023freedom}
Jiwen Yu, Yinhuai Wang, Chen Zhao, Bernard Ghanem, and Jian Zhang.
\newblock Freedom: Training-free energy-guided conditional diffusion model.
\newblock In \emph{Proceedings of the IEEE/CVF International Conference on Computer Vision}, pages 23174--23184, 2023.

\bibitem[Zhang et~al.(2023)Zhang, Wang, Xue, Yan, Zhang, Bai, and Shou]{zhang2023datasetcondensationgenerativemodel}
David~Junhao Zhang, Heng Wang, Chuhui Xue, Rui Yan, Wenqing Zhang, Song Bai, and Mike~Zheng Shou.
\newblock Dataset condensation via generative model, 2023.

\bibitem[Zhang et~al.(2024)Zhang, Su, Zhu, Sun, and Zhang]{zhang2024gsdd}
Haiyu Zhang, Shaolin Su, Yu Zhu, Jinqiu Sun, and Yanning Zhang.
\newblock Gsdd: generative space dataset distillation for image super-resolution.
\newblock In \emph{Proceedings of the AAAI Conference on Artificial Intelligence}, pages 7069--7077, 2024.

\bibitem[Zhang et~al.(2026)Zhang, Wang, Lu, Wang, Qian, Xu, Gu, and Zhang]{zhang2026spherical}
Junyi Zhang, Yiming Wang, Yunhong Lu, Qichao Wang, Wenzhe Qian, Xiaoyin Xu, David Gu, and Min Zhang.
\newblock Spherical geometry diffusion: Generating high-quality 3d face geometry via sphere-anchored representations.
\newblock \emph{arXiv preprint arXiv:2601.13371}, 2026.

\bibitem[Zhao and Bilen(2021)]{zhao2021dataset}
Bo Zhao and Hakan Bilen.
\newblock Dataset condensation with differentiable siamese augmentation.
\newblock In \emph{International Conference on Machine Learning}, pages 12674--12685. PMLR, 2021.

\bibitem[Zhao et~al.(2020)]{zhao2020dataset}
Bo Zhao et~al.
\newblock Dataset condensation with gradient matching.
\newblock \emph{arXiv preprint arXiv:2006.05929}, 2020.

\bibitem[Zhao et~al.(2023)]{zhao2023dataset}
Bo Zhao et~al.
\newblock Dataset condensation with distribution matching.
\newblock In \emph{Proceedings of the IEEE/CVF Winter Conference on Applications of Computer Vision}, pages 6514--6523, 2023.

\bibitem[Zhong et~al.(2025{\natexlab{a}})Zhong, Tang, Zheng, Xu, Hu, and Guan]{zhong2025towards}
Wenliang Zhong, Haoyu Tang, Qinghai Zheng, Mingzhu Xu, Yupeng Hu, and Weili Guan.
\newblock Towards stable and storage-efficient dataset distillation: Matching convexified trajectory.
\newblock In \emph{Proceedings of the Computer Vision and Pattern Recognition Conference}, pages 25581--25589, 2025{\natexlab{a}}.

\bibitem[Zhong et~al.(2025{\natexlab{b}})Zhong, Fang, Chen, Gu, Qiu, Qi, and Xia]{zhong2025hierarchicalfeaturesmatterdeep}
Xinhao Zhong, Hao Fang, Bin Chen, Xulin Gu, Meikang Qiu, Shuhan Qi, and Shu-Tao Xia.
\newblock Hierarchical features matter: A deep exploration of progressive parameterization method for dataset distillation, 2025{\natexlab{b}}.

\end{thebibliography}
}


\clearpage
\setcounter{page}{1}
\maketitlesupplementary
\suppsection{Background}
\label{supsec:BG}
\setcounter{equation}{0}
\setcounter{figure}{0}
\setcounter{table}{0}

\suppsubsection{Diffusion Sample process}
Diffusion models \cite{ho2020denoising, song2020denoising,lu2025smoothed} comprise a forward process $\{q(\boldsymbol{x}_{t})\}_{t\in[0,T]}$ that gradually adds noise to data $\boldsymbol{x}_{0} \sim q(\boldsymbol{x}_{0})$, alongside a learned reverse process $\{p(\boldsymbol{x}_{t})\}_{t\in[0,T]}$ targeting to denoise the data. 

The forward process is formulated as $q(\boldsymbol{x}_{t}|\boldsymbol{x}_{0}):=\mathcal{N}(\sqrt{\alpha_{t}}\boldsymbol{x}_{0},(1-\alpha_{t})\mathbf{I})$ and $q(\boldsymbol{x}_{t}):= \int q(\boldsymbol{x}_{t}|\boldsymbol{x}_{0})q(\boldsymbol{x}_{0}) \mathrm{d}\boldsymbol{x}_{0}$, with $\alpha_{t}$ representing a noise schedule. The reverse process, initialized from $p(\boldsymbol{x}_{T}):=\mathcal{N}(\mathbf{0},\mathbf{I})$, is characterized by a parameterized denoiser $\boldsymbol{\epsilon}_{\theta}^{t}(\boldsymbol{x}_{t})$, which aims to predict the noise added to $\boldsymbol{x}_{0}$. The denoiser $\epsilon_\theta$can be optimized by minimizing:
\begin{equation}
  \mathcal{L}_{\mathrm{DM}}:=\mathbb{E}_{x_{0},t,\boldsymbol{\epsilon}}[w(t)\left\|\boldsymbol{\epsilon}_{\theta}^{t}(\sqrt{\alpha_{t}}\boldsymbol{x}_{0}+\sqrt{1-\alpha_{t}}\boldsymbol{\epsilon})-\boldsymbol{\epsilon}\right\|^{2}_{2}]
  \label{eq:diffusion_loss_app}
\end{equation}
where $\boldsymbol x_{0}\sim q(\boldsymbol x_{0}),t\sim \mathcal{U}(0,T),\boldsymbol{\epsilon} \sim \mathcal{N}(\mathbf{0},\mathbf{I})$ and $w(t)$ is a pre-specified weight function. A more widely adopted approach is the Latent Diffusion Model (LDM) \cite{rombach2022high,cao2025dimension,zhang2026spherical}, which leverages a Variational Autoencoder (VAE) \cite{kingma2013auto} to compress input $x$ into latent space samples $z$, followed by executing diffusion within this latent space. In this work, we employ LDM as a pretrained backbone model requiring no additional training, and adopt the sampling process defined by DDIM (Denoising Diffusion Implicit Models) \cite{song2020denoising, lu2025inpo}. 
DDIM first maps the noisy sample $z_t$ back to the clean data distribution, obtaining $z_{0|t}$. Then, it samples $z_{t-1}$ through the diffusion process:
\begin{equation}
    z_{0|t}=\frac{z_t - \sqrt{1 - \bar{\alpha}_t}\epsilon_{\theta}(z_t, t,c)}{\sqrt{\bar{\alpha}_t}}
    \label{eq:z0t_app}
\end{equation}
We can finally obtain the single-step denoising result via the DDIM sampling formula:
\begin{equation}
\label{ddim_app}
    z_{t - 1} =\alpha_t^1z_{0|t}(z_t) + \alpha_t^2\epsilon_{\theta}(z_t, t,c) +\alpha_t^3 \epsilon
\end{equation}
where $\alpha_t^1=\sqrt{\bar{\alpha}_{t - 1}}$, $\alpha_t^2=\sqrt{1 - \bar{\alpha}_{t - 1} - \eta^2(1 - \bar{\alpha}_t)}$, $\alpha_t^3= \eta\sqrt{1 - \bar{\alpha}_{t - 1}}$. $\eta$ is predefined noise factor. For compact representation, we define whole process as $z_{t-1}=DDIM(z_t,t,c)$. Furthermore, we can incorporate other conditional gradient guidance during sampling to achieve guided diffusion \cite{yu2023freedom}. Given a differentiable conditioning function $E(z_t,c)$, where $c$ represents a conditional input of arbitrary form, we can define a single-step guided diffusion process as:
\begin{equation}
\label{TFG_app}
    z_{t-1}=DDIM(z_t)-\rho_t\nabla E(z_t,c)
\end{equation}
However, directly evaluating $E(z_t,c)$ on noisy samples $z_t$ is challenging. Thus, we approximate it by computing it at the mapped point of $z_t$ on the clean data manifold, i.e., $E(z_t,c) \approx \hat{E}(z_{0|1}(z_t),c)$, where $z_{0|1}$ is the denoised estimate of $z_t$ via \cref{eq:z0t_app}.
\paragraph{Entropy-Regularized OT and Sinkhorn Algorithm}
To address the computational challenges of OT, \textit{entropy regularization} introduces a penalization term to the objective, smoothing the transport plan $\gamma$ and enabling efficient computation. The entropy-regularized OT problem is defined as:
\begin{equation} 
W_{\varepsilon}(\mathbf{a}, \mathbf{b}) = \min_{\gamma \in \Pi(\mathbf{a}, \mathbf{b})} \langle \gamma, \mathbf{C} \rangle - \varepsilon H(\gamma),  
\end{equation}
where $\varepsilon > 0$ controls the strength of the regularization, and $H(\gamma) = -\sum_{i,j} \gamma_{ij} \log \gamma_{ij}$ is the entropy of the transport plan. The regularization makes the problem strictly convex and allows for efficient iterative solutions, even for large $\mathbf{a}, \mathbf{b}$.
The Sinkhorn algorithm is an iterative method to solve the entropy-regularized OT problem. It leverages the fact that the optimal transport plan $\gamma^*$ under entropy regularization can be expressed in a factorized form:
\begin{equation}
P^*_{ij} = \mathbf{a}_i \mathbf{b}_j \exp\left( -\frac{C_{ij}}{\epsilon} + u_i + v_j \right)
\end{equation}
where $u\in \mathbb{R}^{n}$ and $v\in \mathbb{R}^{m}$ are dual variables ensuring the marginal constraints are satisfied. Rearranging, this simplifies to:
\begin{equation}
P^* = \text{diag}(u) \, K \, \text{diag}(v)
\end{equation}
where $K \in \mathbb{R}^{n \times m}_+$ is the kernel matrix defined as $K_{ij} = \exp\left( -\frac{C_{ij}}{\epsilon} \right)$, and $\text{diag}(u)$ (resp. $\text{diag}(v)$) is a diagonal matrix with $u$ (resp. $v$) on the diagonal.
In practice, the Sinkhorn algorithm alternates between updating $u$ and $v$ to enforce the marginal constraints. Starting with initial guesses $u_0 = \mathbf{1}$ (all ones) and $v_0 = \mathbf{1}$, the updates are:
\begin{equation}
\begin{aligned}
v_k &= \frac{\beta}{K^\top u_{k-1}}, \\
u_k &= \frac{\alpha}{K v_k}
\end{aligned}
\end{equation}

where $K^\top$ denotes the transpose of $K$, and division is element-wise. After $T$ iterations, the transport plan is approximated as $P \approx \text{diag}(u_T) K \text{diag}(v_T)$.
Besides, A simplified and numerically stable variant of the Sinkhorn algorithm is the \textit{row-column normalization method}, which directly operates on the kernel matrix $K$ without explicitly tracking $u$ and $v$. The key insight is that alternating row and column normalization of $K$ enforces the marginal constraints $\alpha$ and $\beta$ iteratively \cite{cui2025optical}. The steps are as follows:
\begin{equation}
        K_{\text{row}} = K \odot \left( \frac{\alpha}{\text{row\_sum}(K)} \right)
\end{equation}
\begin{equation}
        K = K_{\text{row}} \odot \left( \frac{\beta}{\text{col\_sum}(K_{\text{row}})} \right)
\end{equation}
where $\text{row\_sum}(K) \in \mathbb{R}^n$ is the vector of row sums of $K$, and $\odot$ denotes element-wise multiplication. $\text{col\_sum}(K_{\text{row}}) \in \mathbb{R}^m$ is the vector of column sums of $K_{\text{row}}$. After $T$ iterations, the normalized $K$ itself serves as the approximate transport plan $P \approx K$.
\suppsubsection{More Related Work} 
\paragraph{Optimization Based Methods.} Optimization based methods are classical dataset distillation algorithms. They align representations or training dynamics between synthetic datasets ($\mathcal{S}$) and real datasets ($\mathcal{T}$) via matching losses, and update the synthetic dataset through gradient optimization. Gradient Matching (GM), one of the earliest dataset distillation algorithms, updates samples by matching training gradients on $\mathcal{S}$ and $\mathcal{T}$  \cite{zhao2020dataset,zhao2021dataset,shin2023loss}. However, GM requires simultaneous gradient updates for both samples and the model, leading to a bi-level optimization dilemma. In contrast, Trajectory Matching (TM) aims to directly match training trajectories between $\mathcal{S}$ and $\mathcal{T}$ without complex gradient computations \cite{cazenavette2022dataset,guo2023towards,du2023sequential,liu2024dataset,zhong2025towards,cui2023scaling}. \citet{guo2023towards} observed that different parameter trajectories can be adopted for distillation across datasets, achieving lossless distillation on small-scale datasets for the first time. Distribution Matching (DM) seeks to ensure that $\mathcal{S}$ effectively covers $\mathcal{T}$ in the feature space, i.e., matching their feature distributions \cite{zhao2023dataset,liu2023dream,wang2025dataset,shao2024generalized,shen2025delt}. \citet{zhao2023dataset} proposed using randomly initialized feature extractors for mapping and matching the means of $\mathcal{T}$ and $\mathcal{S}$ to approximate distribution matching. \cite{liu2023dream} proposed selecting representative data via K-means clustering for matching. Optimal Transport is regarded as a key insight for enhancing distribution matching. \cite{liu2023dataset} proposed using the Wasserstein barycenter of the $\mathcal{T}$ as matching targets. OPTICAL \cite{cui2025optical} leverages mini-batch optimal transport to improve the matching relationship between samples in $\mathcal{S}$ and $\mathcal{T}$. Our method also draws on the key insight of optimal transport, designing a new OT-guided loss for the diffusion based dataset distillation framework. We further propose two key strategies: approximate distribution matching and greedy progressive matching, to ensure performance while further optimizing efficiency.
\paragraph{Disentangled Dataset Distillation}
Disentangled dataset distillation frameworks have successfully overcome the bi-level optimization dilemma, extending dataset distillation to large-scale datasets such as ImageNet\cite{yin2023squeeze,shao2024generalized,xiao2024large,xue2025towards,sun2024diversity,kim2022dataset,shin2025distilling,liu2022dataset}. SRe2L \cite{yin2023squeeze} proposed a squeeze-recover-relabel paradigm: first, it squeezes the key information of the dataset into a neural network through training; then, it optimizes samples through designed matching losses for recovery; finally, it performs relabeling based on the pretrained model. G-VBSM \cite{shao2024generalized} extended such methods via large-scale statistical matching and multi-backbone model. \citet{xiao2024large} proposed a label pruning method to optimize the label space, significantly reducing the storage space of such methods. \citet{xue2025towards} proposed a curvature regularization loss to improve the adversarial robustness of disentangled dataset distillation. Inspired by these approaches, \citet{sun2024diversity} introduced a non-optimization framework RDED, which conducts dataset distillation by directly extracting effective patches using a pre-trained model.  Inspired by this category of methods, we designed semantic matching and distribution matching objectives for diffusion based dataset distillation. Meanwhile, we further improved the matching framework specifically for diffusion models.

\paragraph{Generative Model Based Dataset Distillation}
In contrast to methods based on discriminative models, generative model based approaches can synthesize data that exhibits high consistency with the original dataset. This consistency (also termed realism) effectively enhances cross-architecture performance. Prior research \cite{cazenavette2023generalizing,zhong2025hierarchicalfeaturesmatterdeep,zhang2023datasetcondensationgenerativemodel,wang2023dimdistillingdatasetgenerative} proposed using Generative Adversarial Networks (GANs) \cite{goodfellow2014generative} as prior models for dataset distillation, synthesizing realistic data by optimizing latent space variables. \cite{zhang2024gsdd} extended GAN-based dataset distillation methods to the image super-resolution setting, further validating the immense potential of generative models in dataset distillation. Recently, researchers have increasingly focused on applying diffusion models \cite{croitoru2023diffusion,ho2020denoising,song2020denoising} to dataset distillation \cite{gu2024efficient,Su_2024_CVPR,chan2025mgd,chen2025influence}. Minimax \cite{gu2024efficient} introduced an efficient fine-tuning-based method \cite{xie2023difffit} to further align diffusion models with target datasets. D$^4$M \cite{Su_2024_CVPR} proposed a disentangled diffusion model framework: it first extracts mode means via K-means and generates representative samples through DDIM inversion; subsequently, it employs knowledge distillation for soft label annotation. MGD$^3$ \cite{chan2025mgd} devised a training-free guided diffusion model framework for dataset distillation, comprising three stages: mode discovery, mode guidance, and stop guidance. However, this method lacks attention to the distribution structure, which may lead to overemphasizing invalid mode points. IGD \cite{chen2025influence} introduce  trajectory matching into diffusion model guidance, utilizing a auxiliary trained  classifier to steer generation toward high-influence samples. However, complex trajectory optimization causes it to lose the efficient characteristics of diffusion based methods. Independently and concurrently with our work, \cite{cui2025optimizing} explored the application of optimal transport-based diffusion models in dataset distillation, with a specific focus on how optimal transport relates to soft label learning within this task. Our method rethinks the framework for applying diffusion models to dataset distillation, proposing two core objectives: semantic matching and distribution matching. For semantic matching, we demonstrate that diffusion models effectively inject semantic information and design a dynamic soft labeling approach to enhance diversity. For distribution matching, we propose an optimal transport-guided loss that effectively aligns the distribution of generated samples with the real dataset without requiring additional model training.

\suppsection{Proof}
\label{supsec:proof}
In this section, we will provide detailed proofs for the theoretical analyses presented in the paper, and in conjunction with the design space of dataset distillation, discuss how these theories guide the design of our DMGD framework.
\suppsubsection{Proof of \Cref{thm:risk_bound_app}}
\setcounter{theorem}{0}
\begin{theorem}
\label{thm:risk_bound_app}
Let $\mathcal{T}$ and $\mathcal{S}$ denote the target and surrogate datasets, respectively, with $\theta_{\mathcal{T}}^*$ and $\theta_{\mathcal{S}}^*$ being their optimally trained parameters. Define the target risk as: $R_{\mathcal{T}}(\theta) = \mathbb{E}_{(x,y) \sim \mathcal{T}} \left[ \ell(x, y, \theta) \right],$ where $\ell(\cdot)$ is an $L$-Lipschitz continuous evaluation function. Under semantic class alignment (i.e., no label mismatch), consider the marginal sample distributions $P_{\mathcal{T}}$ and $P_{\mathcal{S}}$ with optimal transport distance: $W(P_{\mathcal{T}}, P_{\mathcal{S}}) = \inf_{\gamma \in \Gamma(P_{\mathcal{T}}, P_{\mathcal{S}})} \mathbb{E}_{(x_{\mathcal{T}}, x_{\mathcal{S}}) \sim \gamma} \left[ d(x_{\mathcal{T}}, x_{\mathcal{S}}) \right],$
where $\Gamma(P_{\mathcal{T}}, P_{\mathcal{S}})$ is the set of all couplings between the distributions, and $d(\cdot,\cdot)$ is a metric on the sample space. Then the risk discrepancy satisfies:
\begin{equation}
\left| R_{\mathcal{T}}(\theta_{\mathcal{T}}^*) - R_{\mathcal{T}}(\theta_{\mathcal{S}}^*) \right| \leq 2L \cdot W(P_{\mathcal{T}}, P_{\mathcal{S}}).
\label{eq:risk_bound_app}
\end{equation}
\end{theorem}
\paragraph{Proof.} Through the optimal properties of parameters $\theta^*$, we decompose the risk discrepancy:
\begin{equation}
\begin{aligned}
    \Delta&= R_{\mathcal{T}}(\theta_{\mathcal{S}}^*) - R_{\mathcal{T}}(\theta_{\mathcal{T}}^*) \\&=R_{\mathcal{T}}(\theta_{\mathcal{S}}^*) - R_{\mathcal{S}}(\theta_{\mathcal{S}}^*)+R_{\mathcal{S}}(\theta_{\mathcal{S}}^*) - R_{\mathcal{T}}(\theta_{\mathcal{T}}^*)
    \\&\leq \underbrace{R_{\mathcal{T}}(\theta_{\mathcal{S}}^*) - R_{\mathcal{S}}(\theta_{\mathcal{S}}^*)}_{I}+\underbrace{R_{\mathcal{S}}(\theta_{\mathcal{T}}^*) - R_{\mathcal{T}}(\theta_{\mathcal{T}}^*)}_{II}
\end{aligned}
\end{equation}
For conciseness, we define $\Delta=\left| R_{\mathcal{T}}(\theta_{\mathcal{T}}^*) - R_{\mathcal{T}}(\theta_{\mathcal{S}}^*) \right|$. We review the definition of risk $R_{\mathcal{T}}(\theta) = \mathbb{E}_{(x,y) \sim \mathcal{T}} \left[ \ell(x, y, \theta) \right]$. Due to the consistency of labels and the consistency of parameters, we can express the first term as:
\begin{equation}
    I=\mathbb{E}_{x \sim P_\mathcal{T}} \left[ \ell_{\theta^*_s}(x) \right]-\mathbb{E}_{x \sim P_\mathcal{S}} \left[ \ell_{\theta^*_s}(x) \right]
\end{equation}
To explain the risk discrepancy from the perspective of optimal transport theory, we introduce the key lemma for \Cref{thm:risk_bound_app}: the Kantorovich-Rubinstein duality (\textbf{\Cref{lemma:kr_duality_app}}).
\setcounter{lemma}{1}
\begin{lemma}[Kantorovich-Rubinstein Duality \cite{villani2008optimal}]

\label{lemma:kr_duality_app}
Let $(\mathcal{X}, d)$ be a complete separable metric space (Polish space). 
For any Borel probability measures $\mu, \nu \in \mathcal{P}_1(\mathcal{X})$ 
with finite first moments, the Wasserstein distance admits the dual representation:
\begin{equation}
\begin{aligned}
W(\mu, \nu) &= \inf_{\gamma \in \Gamma(\mu, \nu)} \mathbb{E}_{(x,y) \sim \gamma} [d(x, y)]
\\&= \sup_{\substack{f \in \text{Lip}_1(\mathcal{X})}} \left( \mathbb{E}_{x \sim \mu} [f(x)] - \mathbb{E}_{y \sim \nu} [f(y)] \right)
\end{aligned}
\end{equation}
where, $\Gamma(\mu, \nu)$ denotes the set of couplings with marginals $\mu$ and $\nu$. $\|f\|_{\text{Lip}} = \sup_{x \neq y} \frac{|f(x) - f(y)|}{d(x,y)}$ is the Lipschitz semi-norm. $\mathcal{P}_1(\mathcal{X})$ is the space of probability measures with $\int d(x_0,x) d\mu(x) < \infty$ for some $x_0 \in \mathcal{X}$
\end{lemma}
 Let $\mu=P_\mathcal{T}$, $\nu=P_\mathcal{S}$. Meanwhile, since $\ell$ satisfies L-Lipschitz continuity, we can set $f(x)=\frac{\ell_{\theta^*_\mathcal{S}}(x)}{L}$. Building on the basis of \cref{lemma:kr_duality_app}, we have:
 \begin{equation}
 \begin{aligned}
I&=\mathbb{E}_{x \sim P_\mathcal{T}} \left[ \ell_{\theta^*_\mathcal{S}}(x) \right]-\mathbb{E}_{x \sim P_\mathcal{S}} \left[ \ell_{\theta^*_\mathcal{S}}(x) \right] \\&=L\cdot( \mathbb{E}_{x \sim P_\mathcal{T}} \left[ f(x) \right]-\mathbb{E}_{x \sim P_\mathcal{S}} \left[ f(x) \right])
\\&\leq L\cdot\sup_{\substack{f \in \text{Lip}_1(\mathcal{X})}} \left( \mathbb{E}_{x \sim P_\mathcal{T}} [f(x)] - \mathbb{E}_{y \sim P_\mathcal{S}} [f(y)] \right)
\\&=L \cdot W(P_\mathcal{T},P_\mathcal{S})
\end{aligned}
 \end{equation}
 Similarly, for the second term, we have:
 \begin{equation}
     \begin{aligned}
          II&=\mathbb{E}_{x \sim P_\mathcal{S}} \left[ \ell_{\theta^*_\mathcal{T}}(x) \right]-\mathbb{E}_{x \sim P_\mathcal{T}} \left[ \ell_{\theta^*_\mathcal{T}}(x) \right] \\&=L\cdot( \mathbb{E}_{x \sim P_\mathcal{S}} \left[ f(x) \right]-\mathbb{E}_{x \sim P_\mathcal{T}} \left[ f(x) \right])
\\&\leq L\cdot\sup_{\substack{f \in \text{Lip}_1(\mathcal{X})}} \left( \mathbb{E}_{x \sim P_\mathcal{S}} [f(x)] - \mathbb{E}_{y \sim P_\mathcal{T}} [f(y)] \right)
\\&=L \cdot W(P_\mathcal{S},P_\mathcal{T})
     \end{aligned}
 \end{equation}
Combining the two terms, based on the symmetric property of the optimal transport distance, i.e. $W(P_\mathcal{S},P_\mathcal{T})=W(P_\mathcal{T},P_\mathcal{S})$, we can derive \Cref{thm:risk_bound_app}:
\begin{equation}
    \left| R_{\mathcal{T}}(\theta_{\mathcal{T}}^*) - R_{\mathcal{T}}(\theta_{\mathcal{S}}^*) \right| \leq 2L \cdot W(P_{\mathcal{T}}, P_{\mathcal{S}}).
\end{equation}
\paragraph{Discussion}
The core idea of \textbf{\Cref{thm:risk_bound_app}} is to decompose the objectives of dataset distillation into two domain adaptation objectives \cite{wang2019transferable, tang2020discriminative,courty2016optimal,kocc2025domain} concerning the optimal parameters on the target dataset and the optimal parameters on the surrogate dataset, respectively. This decomposition bridges the gap between the fields of dataset distillation and domain adaptation. However, traditional domain adaptation algorithms optimize model parameters, while dataset distillation optimizes synthetic samples. This discrepancy in optimization objects makes it challenging to apply optimal transport to optimizing the joint distribution of samples and labels in the context of dataset distillation. Meanwhile, image data has the characteristic of redundancy in information dimensions, which means the semantic information only occupies a small number of dimensions in the pixel or feature space. Performing optimal transport solely on the sample distribution fails to preserve representative semantic information.\par
Therefore, we aim to handle the alignment of semantic information and distribution structures separately, which also constitutes the starting point of our \textbf{\Cref{thm:risk_bound_app}} and the DMGD framework. \textbf{\Cref{thm:risk_bound_app}} indicates that, under certain constraint guidance such that semantic alignment is satisfied, optimizing the optimal transport distance between the surrogate dataset and the target dataset is equivalent to optimizing the upper bound of the risk discrepancy.  Therefore, we only need to consider that surrogate samples must have semantic information consistent with the target class, i.e., semantic alignment. We can define semantic alignment from the perspective of conditional likelihood.
\begin{definition}[Semantic Alignment]
\label{def:semantic_alignment_app}
Let $\mathcal{X}$ be the sample space, $\mathcal{Y} = \{y_1, \dots, y_m\}$ a finite label set of semantic categories, and $\log p(\cdot|x)$ a conditional log-likelihood distribution over $\mathcal{Y}$ for a given sample $x \in \mathcal{X}$. A sample $x$ and target semantic label $y \in \mathcal{Y}$ are \emph{semantically aligned} if and only if:
\begin{equation}
    y = \arg\max_{y^* \in \mathcal{Y}}  \log p(y^*|x)
\end{equation}
\end{definition}
By \textbf{\Cref{def:semantic_alignment_app}}, we can achieve semantic alignment by optimizing the conditional log-likelihood $\log p(y|x)$. In discriminative models, the conditional log-likelihood can be estimated from the softmax output of the classifier, and synthetic samples can be optimized via backpropagation \cite{yin2023squeeze,xiao2024large,shao2024generalized}. In generative models, especially diffusion models, classifier-free guidance \cite{ho2022classifier,gu2024efficient,Su_2024_CVPR} is an effective method for estimating and optimizing conditional log-likelihood. This makes it feasible to align semantics within the diffusion model framework without the need for additional classifier training. This also forms the design basis of our semantic matching.\par
For distribution matching, we still need to first consider whether distribution alignment will lead to a mismatch of semantic information, which is also the premise for handling the two objectives separately. The traditional setup of dataset distillation provides a natural way to meet the assumptions by distilling instances for each class distribution. We perform distribution matching for each class separately to disentangle semantic information. Through optimal transport matching on class distributions, we can obtain the objective of distribution alignment that guides practice.
\begin{equation}
    \arg\min_{\mathcal{S}^c} W(P_{\mathcal{S}^c},P_{\mathcal{T}^c})
\end{equation}
$\mathcal{S}^c$ is the set of instances assigned to class $c$ in the surrogate dataset, and $\mathcal{T}^c$ is the set of samples labeled $c$ in the target dataset.
\suppsubsection{Proof of \Cref{lemma1_app}}
\setcounter{lemma}{0}
\begin{lemma}[Classifier-Free Guidance \cite{ho2022classifier}]  
\label{lemma1_app}
Consider a noise prediction network $\boldsymbol{\epsilon}_\theta(\boldsymbol{z}_t, t, y)$, where $\boldsymbol{z}_t$ denotes the representation of an original sample $\boldsymbol{x}$ at timestep $t$, and $y$ is a label. Assuming the $\boldsymbol{\epsilon}$ models both the conditional generative distribution $p(\boldsymbol{z}_t | y)$ and the unconditional distribution $p(\boldsymbol{z}_t)$, the gradient of the conditional log-likelihood $\log p(y|\boldsymbol{z}_t)$ with respect to $\boldsymbol{z}_t$ can be implicitly approximated by the difference between the network's conditional and unconditional outputs:  
\begin{equation}
    \nabla_{\boldsymbol{z}_t} \log p(y | \boldsymbol{z}_t) \approx \omega \, \Big( \boldsymbol{\epsilon}_\theta(\boldsymbol{z}_t, t, \emptyset) - \boldsymbol{\epsilon}_\theta(\boldsymbol{z}_t, t, y) \Big) 
\end{equation}
Here, $\omega$ denotes a scalar guidance scale, and $\boldsymbol{\epsilon}_\theta(\boldsymbol{z}_t, t, \emptyset)$ represents the network's unconditional output (i.e., without a specified class label).  
\end{lemma}
\paragraph{Proof.} By Bayes' theorem, the conditional likelihood decomposes as:
\begin{equation}
    p(y|z_t) = \frac{p(z_t|y) \cdot p(y)}{p(z_t)}
\end{equation}
Taking the logarithm and differentiating with respect to $z_t$:
\begin{equation}
    \nabla_{z_t} \log p(y|z_t) = \nabla_{z_t} \log p(z_t|y) - \nabla_{z_t} \log p(z_t)
\end{equation}
In diffusion models, the score functions relate to the noise prediction network via:
\begin{equation}
\begin{aligned}
\nabla_{z_t} \log p(z_t|y) &\approx -\sigma_t^{-1} \epsilon_{\theta}(z_t, t, y) \\
\nabla_{z_t} \log p(z_t) &\approx -\sigma_t^{-1} \epsilon_{\theta}(z_t, t, \emptyset)
\end{aligned}
\end{equation}
where $\sigma_t$ is the noise magnitude at timestep $t$. Substituting these identities:
\begin{equation}
\begin{aligned}
\nabla_{z_t} \log p(y|z_t) &\approx -\sigma_t^{-1} \epsilon_{\theta}(z_t, t, y) + \sigma_t^{-1} \epsilon_{\theta}(z_t, t, \emptyset) \\
&\approx \sigma_t^{-1} \left( \epsilon_{\theta}(z_t, t, \emptyset) - \epsilon_{\theta}(z_t, t, y) \right) \\
&\approx \omega \left( \epsilon_{\theta}(z_t, t, \emptyset) - \epsilon_{\theta}(z_t, t, y) \right)
\end{aligned}
\end{equation}
where the guidance scale $\omega$ absorbs the proportionality constant $\sigma_t^{-1}$ and sign convention. The final equivalence follows from reordering terms and the scalar nature of $\omega$.
\paragraph{Discussion} \textbf{\Cref{lemma1_app}} demonstrates that diffusion models can effectively estimate conditional likelihood, thereby providing a foundation for semantic alignment without the need for additional classifier training. In previous works \cite{chan2025mgd,gu2024efficient,sun2024diversity,chen2025influence}, this aspect was incorporated, but without further in-depth analysis. We are the first to elaborate on the design in this aspect and verify its significant impact on the performance of dataset distillation, as shown in \Cref{tab:aS} in the paper.

\suppsubsection{Proof of \Cref{prop:ddim_deterministic_modulation_app}}
\setcounter{proposition}{0}
\begin{proposition}
\label{prop:ddim_deterministic_modulation_app}
Given a single step sampling process (such as DDIM) based on $\epsilon_\theta$ to update $z_{t-1}^{(0)}$ using condition $y$, consider a dynamic label $\hat{y}_t = y + \delta_t$ where $\delta_t$ is a time-dependent vector. The modified sampling step admits the first-order approximation:
\begin{equation}
    z_{t-1} \approx z_{t-1}^{(0)} + \Lambda_t(\delta_t)
\end{equation}
where the condition shift operator $\Lambda_t$ is defined as:
$\Lambda_t(\delta_t) = c_t \cdot \bigl(\nabla_y \epsilon_\theta(z_t, t, y)\bigr)^\top \delta_t$ with $c_t = \sqrt{1-\alpha_{t-1}} - \sqrt{\alpha_{t-1}} \cdot \sqrt{1-\alpha_t}/\sqrt{\alpha_t}$ as the intrinsic time-scaling factor.
\end{proposition}
\paragraph{Proof.} By Taylor expansion, we can approximate the denoising model $\epsilon$ under dynamic label.
\begin{equation}
\epsilon_\theta(z_t, t, y + \delta_t) \approx \epsilon_\theta(z_t, t, y) + \nabla_y \epsilon_\theta(z_t, t, y)^\top \delta_t
\end{equation}
Neglecting the effects of higher-order terms, we substitute the approximation formula into the sampling formula of the diffusion model, taking DDIM (\Cref{ddim_app}) as an example here:
\begin{equation}
\begin{aligned}
z_{t-1} \approx &\alpha_t^1(\frac{z_t-\sqrt{1-\bar{\alpha}_t}(\epsilon_\theta(z_t, t, y) + \nabla_y \epsilon_\theta(z_t, t, y)^\top \delta_t)}{\sqrt{\bar{\alpha}}})\\&+\alpha_t^2(\epsilon_\theta(z_t, t, y) + \nabla_y \epsilon_\theta(z_t, t, y)^\top \delta_t)+\alpha_t^3\epsilon
\end{aligned}
\end{equation}
After rearrangement, we obtain:
\begin{equation}
    z_{t-1} = z_{t-1}^{(0)} + c_t\nabla_y \epsilon_\theta(z_t, t, y)^\top \delta_t
\end{equation}
where, $z_{t-1}^{(0)}$ corresponds to a standard DDIM sampling process, $c_t= \sqrt{1 - \alpha_{t-1}} - \sqrt{\alpha_{t-1}} \cdot \sqrt{1 - \alpha_t}/\sqrt{\alpha_t}$. We define the condition shift operator $\Lambda_t =c_t\nabla_y \epsilon_\theta(z_t, t, y)^\top \delta_t$, which represents the additional shift term introduced by dynamic labels in the sampling dynamics of the data distribution space.
\paragraph{Discussion}
From \Cref{prop:ddim_deterministic_modulation_app}, we can observe that the dynamic term introduces an additional shift term into the sampling dynamics of diffusion models. Researchers have demonstrated that such an offset term helps diffusion models move away from local mode points, further explore the distribution space, and thereby enhance diversity \cite{sadat2023cads}. Similarly, adding an shift term directly in the sampling process of the sample space can also achieve a similar effect. However, it should be noted that this method leads to more complex computations due to the higher dimensionality of the sample space. Meanwhile, the regulation of the shift term in the sample space is also trick, unreasonable coefficients may directly disrupt the entire sampling process. Stable regulation coefficients often need to be obtained by computing the derivative of $\epsilon$. Therefore, introducing a dynamic process in the label space is a more reasonable choice for us.\par
Furthermore, starting from our goal of generating diverse and high-information samples, we propose the design of two shift terms. The noise shift term provides an effective exploration direction, while the soft label term offers guidance toward class boundaries. Since the soft labels selected for each sample are different, the soft label term can also provide reasonable diversity guidance.

\suppsubsection{Proof of \Cref{cor:approx_bound_app}}
\setcounter{corollary}{0}
\begin{corollary}
\label{cor:approx_bound_app}
Under the conditions of \Cref{thm:risk_bound_app}, consider an approximate distribution $\widetilde{P}_{\mathcal{T}}$ satisfying $W(\widetilde{P}_{\mathcal{T}}, P_{\mathcal{T}}) \leq \epsilon$ for small $\epsilon > 0$. Assuming the distance metric satisfies the triangle inequality,  distributions lie in a Polish space. The risk discrepancy is bounded by:
\begin{equation}
\left| R_{\mathcal{T}}(\theta_{\mathcal{T}}^*) - R_{\mathcal{T}}(\theta_{\mathcal{S}}^*) \right| \leq 2L \cdot \left( W(P_{\mathcal{S}}, \widetilde{P}_{\mathcal{T}}) + W(P_{\mathcal{T}}, \widetilde{P}_{\mathcal{T}}) \right)
\label{eq:approx_bound_app}
\end{equation}
\end{corollary}
\paragraph{Proof.} Let $P_{\mathcal{S}}, P_{\mathcal{T}}, \widetilde{P}_{\mathcal{T}}$ be Borel probability measures on a Polish metric space $(X,d)$. Let $\gamma_1$ and $\gamma_2$ be the optimal couplings corresponding to $W(P_{\mathcal{S}}, \widetilde{P}_{\mathcal{T}})$ and $W(P_{\mathcal{T}}, \widetilde{P}_{\mathcal{T}})$, respectively. By the \textbf{gluing lemma}, construct a measure $\gamma$ on $X^3$ with $(x,y)$-marginal $\gamma_1$ and $(y,z)$-marginal $\gamma_2$. Project $\gamma$ to a coupling $\gamma_{13} \in \Gamma(P_S, P_T)$ via $\gamma_{13}(A \times C) = \gamma(A \times X \times C)$. Then, using the triangle inequality for $d$, we have:  
\begin{equation}
\begin{aligned}
    \int_{X^2} d(x,z) \, d\gamma_{13} &= \int_{X^3} d(x,z) \, d\gamma \\&\leq \int_{X^3} \left[ d(x,y) + d(y,z) \right] d\gamma \\& \leq\int_{X^2} d(x,y) \, d\gamma_1+\int_{X^2} d(y,z) \, d\gamma_2 \\&=W(P_{\mathcal{S}}, \widetilde{P}_{\mathcal{T}})+W(P_{\mathcal{T}}, \widetilde{P}_{\mathcal{T}})
\end{aligned}
\end{equation}
Since $W(P_{\mathcal{S}}, P_{\mathcal{T}})$ is the infimum over all couplings in $\Gamma(P_S, P_T)$:
\begin{equation}
\begin{aligned}
    W(P_{\mathcal{S}}, P_{\mathcal{T}}) &\leq\int_{X^2} d(x,z) \, d\gamma_{13} \\&\leq W(P_{\mathcal{S}}, \widetilde{P}_{\mathcal{T}})+W(P_{\mathcal{T}}, \widetilde{P}_{\mathcal{T}})
\end{aligned}
\end{equation}
Substituting the results into the \Cref{thm:risk_bound_app}, we can obtain:
\begin{equation}
    \left| R_{\mathcal{T}}(\theta_{\mathcal{T}}^*) - R_{\mathcal{T}}(\theta_{\mathcal{S}}^*) \right| \leq 2L \cdot \left( W(P_{\mathcal{S}}, \widetilde{P}_{\mathcal{T}}) + W(P_{\mathcal{T}}, \widetilde{P}_{\mathcal{T}}) \right)
\end{equation}
\paragraph{Disscusion.} \textbf{\Cref{cor:approx_bound_app}} reveals that the target risk discrepancy admits an upper bound. This decomposition provides critical guidance for practical implementation. The first term $W(P_{\mathcal{S}}, \widetilde{P}_{\mathcal{T}})$ represents the alignment error, whose optimization requires $\widetilde{P}_{\mathcal{T}}$ to be computationally tractable. The second term $W(\widetilde{P}_{\mathcal{T}}, P_{\mathcal{T}})$ quantifies the approximation error, which must be minimized to preserve distributional fidelity. To satisfy both requirements, we seek a discrete approximation $\widetilde{P}_{\mathcal{T}}$ that minimizes $W(\widetilde{P}_{\mathcal{T}}, P_{\mathcal{T}})$ while enabling efficient optimization of $P_{\mathcal{S}}$.
This leads naturally to the classical optimal quantization problem \cite{canas2012learning}. Clustering algorithms are efficient solutions with good convergence properties for this type of problem.
\suppsubsection{Proof of \Cref{prop:epsilon of W_app}}
\begin{proposition}
\label{prop:epsilon of W_app}
Let $\tilde{P}_{\mathcal{T}}^{(1)}$ denote the mean-matching approximation of $P_{\mathcal{T}}$ defined by a Dirac measure $\delta_{\mu}$ concentrated at the mean $\mu$ of $P_{\mathcal{T}}$, and $\tilde{P}_{\mathcal{T}}^{(2)}$ denote the proposed approximation constructed via our method with cluster count $K $. The Wasserstein distance satisfies:
\begin{equation}
W(P_{\mathcal{T}}, \tilde{P}_{\mathcal{T}}^{(2)}) \leq W(P_{\mathcal{T}}, \tilde{P}_{\mathcal{T}}^{(1)})
\label{eq:prop5_wasserstein_app}
\end{equation}
\end{proposition}
\paragraph{Proof.}
For $ \tilde{P}_\mathcal{T}^{(1)}$, the optimal transport cost is the integral of distances to $\mu $: 
\begin{equation}
    W(P_\mathcal{T}, \tilde{P}_\mathcal{T}^{(1)}) = \int d( x , \mu)  \, dP_\mathcal{T}(x)
\end{equation}
For $ \tilde{P}_\mathcal{T}^{(2)} $, consider transporting mass in cluster $C_i$  to its centroid  $k_i$ . The cost of this local plan is:  
\begin{equation}
    \text{Cost} = \sum_{i=1}^K \int_{C_i} d( x , k_i) \, dP_{\mathcal{T}}(x)
\end{equation}

By the key property of K-means, $k_i$ is the optimal center for $C_i$ , meaning it minimizes the local transport cost:
\begin{equation}
    \int_{C_k} d( x , k_i) \, dP_{\mathcal{T}}(x) \leq \int_{C_k}d( x , z) \, dP_{\mathcal{T}}(x), \quad \forall z \in \mathbb{R}^d
\end{equation}
Setting $z = \mu$ (the mean of $P_T$), we immediately get:
\begin{equation}
    \int_{C_i} d( x , k_i) \, dP_{\mathcal{T}}(x) \leq \int_{C_i} d( x , \mu) \, dP_\mathcal{T}(x)
\end{equation}
Summing the inequality over all clusters $i = 1, \dots, K$, we have:
\begin{equation}
    \text{Cost}\leq \sum_{i=1}^K \int_{C_i} d( x , \mu), dP_{\mathcal{T}}(x)=\int d( x , \mu) \, dP_\mathcal{T}(x)
\end{equation}
For all transport plans, the optimal transport plan achieves the minimal cost, and thus:
\begin{equation}
    W(P_\mathcal{T}, \tilde{P}_\mathcal{T}^{(2)}) \leq \text{Cost}\leq \int d( x , \mu) \, dP_\mathcal{T}(x) = W(P_\mathcal{T}, \tilde{P}_\mathcal{T}^{(1)})
\end{equation}

\paragraph{Discussion.} In a more general case, we can analyze the bounds of $W(P_\mathcal{T}, \tilde{P}_\mathcal{T}^{(2)})$. When effectively evaluating the intrinsic manifold dimension of the data, for a specific $K$, we can obtain the Wasserstein distance convergence bounds between the approximate distribution based on the K-means algorithm and the original distributions. We recommend referring to \textbf{Theorem 5.2} in \cite{canas2012learning} for more details.

\suppsection{More Implementation Details}
\label{suppsec:ID}

\suppsubsection{More Details of Method}
\begin{algorithm*}[t]
\caption{Dual Matching-Guided Diffusion Models }
\begin{algorithmic}[1]
\REQUIRE CFG factor $\omega$, semantic matching coefficient $\beta_n$ and $\beta_s$, distribution matching coefficient $\rho$, number of Support Points $K$, number of class $C$, image per class $N$, distribution matching range $[t_1,t_2]$
\REQUIRE Target dataset $\mathcal{T}$, pre-trained diffusion model $\epsilon_\theta$, VAE decoder model $V_D$. 
\ENSURE  Surrogate dataset $\mathcal{S}$

\FOR{$c=1$ \TO $C$}
\STATE Obtain the approximated distribution 
$\widetilde{P}_{\mathcal{T}}$ via \Cref{alg:km_approx}
\STATE Initialize class-aware surrogate dataset storage $\mathcal{S}^c_{[0]} \gets \{\}$
\FOR{$n=1$ \TO $N$}
\STATE Sample initial random noise $z^T_n \sim N(0, I)$;
\STATE Select $y^\star,\text{s.t.}\quad y^\star \neq c$
\FOR{$t=T$ \TO $t$}
\STATE Obtain dynamic label $\widetilde{y}_t$ via \Cref{eq:dsl}
\STATE Semantic matching guided sampling $z_{t-1}=D(z_t,t,\widetilde{y}_t,\epsilon_\theta)$.
\IF{$t \in [t_1,t_2]$}
\STATE Obtain a temporary class-aware surrogate dataset $\mathcal{S}^c_{[n]}\gets \mathcal{S}^c_{[n-1]} \cup \{z_{0|t}(z_t)\}$.
\STATE Calculate the OT loss $\mathcal{L}_{\text{OT}}(P_{\mathcal{S}^c_{\left[ i \right]}},\widetilde{P}_{\mathcal{T}^c})$ via Sinkhorn algorithm.
\STATE Distribution matching guided sampling $z_{t-1}=z_{t-1}-\rho_t\nabla_{z_t} \mathcal{L}_{\text{OT}}(P_{\mathcal{S}^c_{\left[ n \right]}},\widetilde{P}_{\mathcal{T}^c})$.
\ENDIF
\ENDFOR
\STATE Store surrogate data $\mathcal{S}^c_{[n]}\gets \mathcal{S}^c_{[n-1]} \cup \{z_0\}$
\ENDFOR
\ENDFOR
\RETURN Decoded synthetic image $\mathcal{S} = V_D(\mathcal{S}_{[N]})$
\end{algorithmic}
\label{alg:overall}
\end{algorithm*}
We provide detailed specifics of the algorithm rapid reproduction, and we will also open-source the implementation code of the paper after organizing it. \textbf{\Cref{alg:overall}} formalizes the overall framework of our approach, featuring parallel application of dual-matching guidance at targeted diffusion phases.

\begin{figure}
    \centering
    \includegraphics[width=1\linewidth]{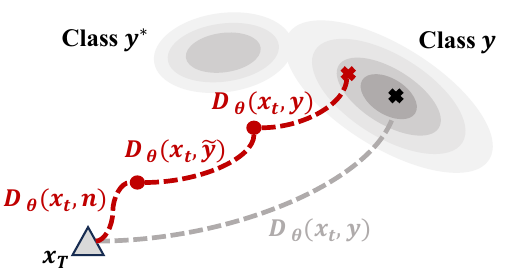}
    \caption{Intuitive demonstration of the dynamic semantic matching guided sampling process. Compared with the sampling process without dynamic guidance, our method can greatly improve diversity and avoid oversampling samples in high-density regions.}
    \label{fig:sm}
\end{figure}

\paragraph{Semantic Matching.} Building on insights from \citet{yu2023freedom}, we partition the diffusion process into three distinct phases for semantic matching:
1) Chaotic Stage: Leveraging pure noise vectors as label proxies to facilitate exhaustive stochastic exploration.
2) Semantic Stage: Employing our proposed dynamic soft labels for guided generation.
3) Refinement Stage: Conducting deterministic sampling with target vectors to ensure semantic alignment.The \cref{fig:sm} intuitively illustrates our dynamic sampling process.
The label strategy across stages is mathematically formalized as:
\begin{equation}
    \widetilde{y}_t=
    \begin{cases} 
    n &t\geq t_1\\
    \sqrt{\sigma_t}y+(1-\sqrt{\sigma_t})(\beta_sy^\star+\beta_n n) &t_2<t< t_1\\
    y&t\leq t_2
    \end{cases}
    \label{eq:dsl}
\end{equation}
where $n \sim \mathcal{N
}(0,I)$ denotes Gaussian noise, $y^\star$ is a random select label subject to $y^\star\neq y$, $\beta_n$ and $\beta_s$ are modulation coefficients, and $\sigma_t$ represents a time-dependent scheduling term defined as:
\begin{equation}
    \sigma_t=\frac{t_1-t}{t_1-t_2}
\end{equation}
Based on the observations of the stages of the diffusion model, we set $t_1=45$ and $t_2=25$. Furthermore, to maintain semantic consistency, we also perform rescaling on the label vectors. The rescaling process is determined by the mean and variance of the target label vectors.
\begin{equation}
    \widetilde y_{re}=\frac{\widetilde y-mean(\widetilde y)}{std(\widetilde y)}*std(y)+mean(y)
\end{equation}
Substituting the label vector into the denoising model and applying classifier-free guidance, we have
\begin{equation}
\begin{aligned}
    \hat\epsilon_\theta (z_t,t,\widetilde{y}_t)&=\epsilon_\theta(\mathbf{z}_t, t, \widetilde{y}_t)-\nabla_{\boldsymbol{z}_t} \log p(y | \boldsymbol{z}_t) \\&\approx (1+\omega)\epsilon_\theta(\mathbf{z}_t, t, \widetilde{y}_t)-\omega\epsilon_\theta(\mathbf{z}_t, t,\emptyset)
\end{aligned}
\end{equation}
In practice, $ \emptyset$ is also injected into the denoising model in the form of label vectors. Therefore, we suggest imposing a dynamic process on it as well to improve stability. We define the single-step dynamic sampling process as $z_{t-1}=D_{\theta}(z_t,t,\widetilde{y_t})$.

\paragraph{Distribution Matching.}

\begin{figure}
    \centering
    \includegraphics[width=1\linewidth]{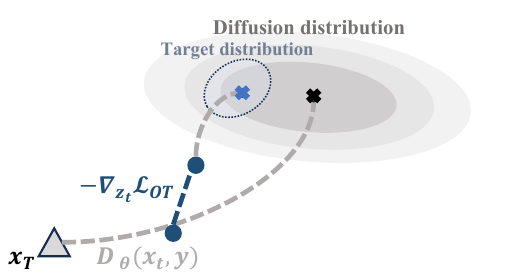}
    \caption{Intuitive demonstration of the distribution matching guided sampling process.Compared with the original sampling process , adding distribution matching guidance can enable the sampling regions to align with the distribution of the target dataset, which is particularly applicable when there is a large discrepancy between the distribution of the diffusion model and that of the target dataset.}
    \label{fig:dm}
\end{figure}
We introduce the distribution matching term in parallel to guide sampling, which show in \cref{fig:dm}. Distribution matching is performed within each class to avoid introducing additional semantic information that interferes with semantic alignment. First, we map the samples in the target dataset to the latent space of the diffusion model. Given the VAE encoder $V_E$ and the image sample $x^i \sim P_\mathcal{T^c}$ and $x^i \in \mathbb{R}^D$  , we have:
\begin{equation}
    z^i_0=V_E(x^i)
\end{equation}
Where, $z\in \mathbb{R}^N$ with $N<D$. To distinguish from noise samples, we define the samples from the data distribution in the latent space as $z_0$. We also introduce a hyperspherical projection to project latent space samples onto the hypersphere of $\mathbb{R}^{N-1}$:
\begin{equation}
    \hat{z}^i_0=\frac{z^i_0}{||z^i_0||_2}
\end{equation}
We define the Euclidean distance in the latent space as the distance metric, which is used for distribution approximation and subsequent optimal transport.
\begin{equation}
    d(z^i_0,z_0^j)=||z^i_0-z_0^j||_2
\end{equation}
Before performing distribution matching, we first need to perform distribution approximation on the distribution of the target dataset to optimize the efficiency of computing optimal transport. We adopt an implementation of the GPU-based fast K-means clustering algorithm \cite{Omer_fast-pytorch-kmeans_2020}. Taking the centroid of the cluster as support points of the approximate distribution, the normalization of the cardinality of each cluster serves as the mass coefficients. We present our distribution approximation algorithm in \textbf{\Cref{alg:km_approx}}.\par
\begin{algorithm}
\caption{K-Means based Distribution Approximation}
\begin{algorithmic}[1]
\REQUIRE Target dataset $\mathcal{T}$, cluster count $K$
\ENSURE Approximated distribution $\widetilde{P}_{\mathcal{T}}$
\STATE Initialize centroids $\{k_i^{(0)}\}_{i=1}^K$
\FOR{$iter=1$ \TO $max\_iter$}
\STATE  $C_i^{(iter)} \gets \{x \in \mathcal{T} : \arg\min_j \|x - k_j^{(iter-1)}\|\}$
\STATE  $k_i^{(iter)} \gets \frac{1}{|C_i^{(iter)}|} \sum_{x \in C_i^{(iter)}} x$
\ENDFOR
\STATE Compute masses: $m_i \gets |C_i^{(final)}|/|\mathcal{T}|$
\RETURN $\widetilde{P}_{\mathcal{T}} = \sum_{i=1}^K m_i \delta_{k_i}$
\end{algorithmic}
\label{alg:km_approx}
\end{algorithm}
Since optimizing all instances simultaneously in the diffusion model space is not feasible, we propose a greedy progressive matching strategy. We construct a memory set $\mathcal{S}^c_{[n]}=\{z^i_0,y^i\}^n_{i=1}$ to store the generated surrogate samples,where $y^i=c$. For the next surrogate sample, we first initialize it from the noise distribution $z_T^{n+1} \sim N(0,I)$ and execute the reverse process. Notably, applying distribution matching guidance in all sample stages is unnecessary (see \Cref{tab:AS ON GM} in the Appendix \ref{supsec:AR}), so we only perform it in stages where $t \in [30,45]$. For a $z_t^{n+1}$ to be optimized, we first map it to the clean data distribution through single-step diffusion.
\begin{equation}
        z^{n+1}_{0|t}=\frac{z^{n+1}_t - \sqrt{1 - \bar{\alpha}_t}\epsilon_{\theta}(z^{n+1}_t, t,c)}{\sqrt{\bar{\alpha}_t}}
\end{equation}
We construct a temporary distribution $\mathcal{S}^c_{[n+1]}$ by combining $z^{n+1}_{0|t}$ with $\mathcal{S}^c_{[n]}$, and calculate the optimal transport distance in sample distribution $P_{\mathcal{S}^c_{[n+1]}}$.
\begin{equation}
    \mathcal{L}_{\text{OT}}=W(P_{\mathcal{S}^c_{[n+1]}},P_{\mathcal{T}^c})=\langle \gamma, \mathbf{C} \rangle
\end{equation}
where $\gamma$ is the optimal coupling, $C$ is the cost matrix. We only need to focus on $z^{n+1}$, which means that $\mathcal{L}_{\text{OT}}$ can be simplified to:
\begin{equation}
\mathcal{L}_{\text{OT}}=\sum_{j=1}^K\gamma^{n+1,j}\cdot C^{n+1,j}
\end{equation}
This loss models the matching relationship between the new sample $z^{n+1}$  and the support points of the approximate distribution. Due to the presence of optimal transport, the optimization direction of $\mathcal{L}_{\text{OT}}$ will prompt $z^{n+1}$ to align with the unaligned regions in the approximate distribution, ensuring the performance of distribution alignment. We utilize training-free guidance technology \cite{yu2023freedom,ye2024tfg} to incorporate this loss into the diffusion model framework.
\begin{equation}
\begin{aligned}
    z^{n+1}_{t-1} & =D_\theta(z^{n+1}_t)-\rho_t\nabla_{z^{n+1}_t} \mathcal{L}_{\text{OT}}(P_{\mathcal{S}^c_{[n+1]}},P_\mathcal{T}) \\&=D_\theta(z^{n+1}_t)-\rho_t\nabla_{z^{n+1}_t}\sum_{j}^K \gamma{^{n+1,j}}\cdot C^{^{n+1,j}}
\end{aligned}
\end{equation}
Following the suggestions from previous work \cite{ye2024tfg}, we set $\rho_t$ as a time-dependent term and a scale term $\rho$:
\begin{equation}
    \rho_t=\rho*log\alpha_t^2
\end{equation}
$log\alpha_t^2$ is log-variance of diffusion models.
\paragraph{Diversity vs. Representativeness trade-off}
While representativeness and diversity have been demonstrated as two crucial characteristics for dataset distillation optimization \cite{sun2024diversity}, our semantic matching and distribution matching specifically enhance representativeness in the semantic space and sample space respectively. Simultaneously, our dynamic guidance mechanism provides a pathway for further diversity optimization. However, in practice, we observe that diversity enhancement does not consistently translate into performance gains. As shown in \Cref{tab:AS ON GM} and \Cref{tab:PE}, we find that for lower IPC settings, diversity enhancement may be unnecessary, whereas in higher IPC configurations, it facilitates broader exploration of the generative distribution and helps avoid local optima. This explains the more substantial performance improvements observed under high IPC conditions. We plan to formulate diversity-related parameters as functions of IPC, representing a promising direction for future research.
\suppsubsection{More Details of Experiment setting}

\paragraph{Evaluation Protocol}
In the hard label protocol, we follow \citet{gu2024efficient} original code and parameter definitions; for more details, please refer to their work. During the training of the target network, we apply the same random resize-crop and CutMix as data augmentation techniques. This protocol is used to evaluate ImageNet subsets, including ImageNet-Woof and ImageNet-Nette \cite{Howard_Imagenette_2019}. It should be noted that we also applied the hard label protocol to ImageNet-IDC \cite{kim2022dataset}; however, to align with \citet{gu2024efficient} and \citet{kim2022dataset}, our specific settings adopt \citet{kim2022dataset}'s definitions. \par
In the soft label protocol, we followed \citet{sun2024diversity} original code and parameter settings. Soft labels are generated by a pre-trained ResNet-18 \cite{he2016deep} model and applied to the training of the evaluation model. We applied this protocol to the full ImageNet-1k dataset.\par
Under the same experimental settings and repeated experiments, we directly report the experimental results of previous works \cite{gu2024efficient,sun2024diversity,chan2025mgd,kim2022dataset}.
\paragraph{Evaluation metric}
We have provided explanations for the evaluation metrics adopted in the paper, including:
\begin{itemize}
    \item \textbf{Accuracy: } The accuracy metric denotes the best TOP-1 accuracy on the test set achieved during the training process of the evaluation model. For the evaluation model, we repeated the training 5 times and report the mean and standard deviation of the best TOP-1 accuracy.
    \item \textbf{Coverage: }For the coverage metric, we first extract features of the dataset using a pre-trained VGG network \cite{simonyan2014very}. The feature space we adopt is the output of the second classification layer of the VGG network, which has a total of 4096 dimensions. We performed code in \citet{ferjad2020icml} to calculate the coverage between the surrogate dataset and the target dataset.
   \item  \textbf{Optimal Transport Dataset Distance:} We utilized the idea of \citet{alvarez2020geometric} and calculated the optimal transport between datasets to evaluate the dataset distance. We adopted the same VGGnet feature space and used t-SNE \cite{maaten2008visualizing} to map it to a two-dimensional space. Optimal transport was applied in the t-SNE space to calculate the dataset distance \cite{feydy2019interpolating}.
   \item \textbf{Diversity:} For the diversity metric, we calculate the minimum distance between each surrogate sample and all other surrogate samples in the VGG feature space, and report the average as the diversity metric.
   \item \textbf{FID:} We directly adopted the official implementation of clean-fid to calculate the FID scores between the surrogate dataset and the target dataset \cite{parmar2022aliased}.
   \item \textbf{Other generative quality metrics}: We directly applied the official implementations of \cite{ferjad2020icml} in the VGG feature space.

\end{itemize}

\suppsection{Additional Result}
\label{supsec:AR}
\suppsubsection{Experiments on Imagenet-IDC}
\begin{table}[h]
\centering
\setlength{\tabcolsep}{8pt}
\begin{tabular}{l|ccc}
\toprule
Method   & IPC-10 & IPC-20 &IPC-50 \\
\midrule
Random      &48.1$_{\pm0.8}$ &   52.5$_{\pm 0.9}$&68.1$\pm0.7$ \\
DM \cite{zhao2023dataset}      &52.8$_{\pm0.5}$ &   58.5$_{\pm 0.4}$ & 69.1$_{\pm0.8}$ \\
DiT \cite{peebles2023scalable}      &54.1$_{\pm0.4}$ &   58.9$_{\pm 0.2}$  &64.3$_{\pm0.6}$ \\
D$^4$M \cite{Su_2024_CVPR}      &51.1$_{\pm2.4}$ &   58.0$_{\pm 1.4}$ &64.1$_{\pm2.5}$  \\
Minimax \cite{gu2024efficient}      &53.1$_{\pm0.2}$ &   59.0$_{\pm 0.4}$&69.6$_{\pm0.2}$   \\
MGD$^3$ \cite{chan2025mgd}      &55.9$_{\pm0.4}$ &   61.9$_{\pm 0.9}$ &72.1$_{\pm0.8}$  \\
\textbf{Ours}      &\textbf{57.0}$_{\pm0.3}$ & \textbf{63.3}$_{\pm1.4}$  &  \textbf{73.2}$_{\pm0.7}$ \\
\bottomrule
\end{tabular}
\caption{Performance comparison between our method and state-of-the-art methods on ImageNet-IDC10, evaluated under the hard-label protocol of \citet{kim2022dataset}. Results are reported as Top-1 accuracy on ResNet-AP with average pooling. The best performance is highlighted in \textbf{bold}.}
\label{tab:imagenetIDC10_results}
\end{table}
We conducted additional experiments on ImageNet-IDC. ImageNet-IDC is a dataset consisting of 100 classes, among which Imagenet-IDC10 corresponds to the first ten classes \cite{kim2022dataset}. We adopted the hyperparameters defined on ImageNet-Woof without additional parameter tuning. The \Cref{tab:imagenetIDC10_results} presents our results on ImageNet-IDC10. Our method achieved the best performance across all IPC settings. Compared with the SOTA method MGD3, we achieved improvements of $1.1\%$, $1.4\%$, and $1.1\%$ respectively. This further validates the generalization capability of our proposed framework.\par
ImageNet-IDC100 is a more complex and larger-scale subset. The \Cref{tab:imagenetIDC100_results} presents our comparative experiments on it. Our method achieves performance comparable to state-of-the-art (SOTA) models while being more stable. This also validates the scalability of our proposed framework. Further precise parameter tuning could yield better results, but we have not conducted it due to constraints on computational resources.\par
Overall, our method achieves excellent performance on the ImageNet-IDC dataset. Compared with other dataset distillation algorithms, our method is more efficient. Under the IPC-10 setting for Imagenet-IDC100, IDC-1 \cite{kim2022dataset} requires over 100 hours of optimization time, while Minimax \cite{gu2024efficient} needs nearly 10 hours for fine-tuning. In contrast, our method introduces only a small amount of additional computation during the sampling stage and takes approximately 0.5 hours.

\begin{table}[h]
\centering
\setlength{\tabcolsep}{12pt}
\begin{tabular}{l|cc}
\toprule
Method   & IPC-10 & IPC-20  \\
\midrule
Random      &19.1$_{\pm0.4}$ &   26.7$_{\pm 0.5}$ \\
Herding \cite{welling2009herding}      &19.8$_{\pm0.3}$ &   27.6$_{\pm 0.1}$   \\
IDC-1 \cite{kim2022dataset}      &\underline{25.7}$_{\pm0.1}$ &   29.9$_{\pm 0.2}$   \\
Minimax \cite{gu2024efficient}      &24.8$_{\pm0.2}$ &   32.3$_{\pm 0.1}$   \\
MGD$^3$ \cite{chan2025mgd} &\textbf{25.8}$_{\pm0.5}$ &   \underline{33.9}$_{\pm 1.1}$   \\
\textbf{Ours}      &\underline{25.7}$_{\pm0.4}$ &   \textbf{34.0}$_{\pm0.1}$  \\
\bottomrule
\end{tabular}
\caption{Performance comparison between our method and state-of-the-art methods on ImageNet-IDC100 , evaluated under the hard-label protocol of \citet{kim2022dataset}. Results are reported as Top-1 accuracy on ResNet-AP with average pooling. The best performance is highlighted in \textbf{bold}, while the second-best is \underline{underlined}.}
\label{tab:imagenetIDC100_results}
\end{table}
\suppsubsection{Experiments on Imagenet-A,B,C,D,E}
\begin{table*}[t]
\centering
\begin{tabular}{l|c|c|c|c|c|c}
\toprule
Distil Alg. & Method & ImageNet-A & ImageNet-B & ImageNet-C & ImageNet-D & ImageNet-E \\
\midrule
\multirow{4}{*}{DC} & Pixel & 52.3\scriptsize{$\pm$0.7} & 45.1\scriptsize{$\pm$8.3} & 40.1\scriptsize{$\pm$7.6} & 36.1\scriptsize{$\pm$0.4} & 38.1\scriptsize{$\pm$0.4} \\
& GLaD\cite{cazenavette2023generalizing} & 53.1\scriptsize{$\pm$1.4} & 50.1\scriptsize{$\pm$0.6} & 48.9\scriptsize{$\pm$1.1} & 38.9\scriptsize{$\pm$1.0} & 38.4\scriptsize{$\pm$0.7} \\
& LD3M\cite{moser2025unlockingdatasetdistillationdiffusion} & 55.2\scriptsize{$\pm$1.0} & 51.8\scriptsize{$\pm$1.4} & 49.9\scriptsize{$\pm$1.3} & 39.5\scriptsize{$\pm$1.0} & 39.0\scriptsize{$\pm$1.3} \\
\midrule
\multirow{4}{*}{DM} & Pixel & 52.6\scriptsize{$\pm$0.4} & 50.6\scriptsize{$\pm$0.5} & 47.5\scriptsize{$\pm$0.7} & 35.4\scriptsize{$\pm$0.4}&36.0\scriptsize{$\pm$0.57} \\
& GLaD\cite{cazenavette2023generalizing} & 52.8\scriptsize{$\pm$1.0} & 51.3\scriptsize{$\pm$0.6} & 49.7\scriptsize{$\pm$0.4} & 36.4\scriptsize{$\pm$0.4} & 38.6\scriptsize{$\pm$0.7} \\
& LD3M\cite{moser2025unlockingdatasetdistillationdiffusion} & 57.0\scriptsize{$\pm$1.3} & 52.3\scriptsize{$\pm$1.1} & 48.2\scriptsize{$\pm$4.9} & 39.5\scriptsize{$\pm$1.5} & 39.4\scriptsize{$\pm$1.8} \\
\midrule
& MGD$^3$\cite{chan2025mgd} & 63.4\scriptsize{$\pm$0.8} & 66.3\scriptsize{$\pm$1.1} & 58.6\scriptsize{$\pm$1.2} & \textbf{46.8}\scriptsize{$\pm$0.8} & 51.1\scriptsize{$\pm$1.0} \\
\midrule
&Ours& \textbf{65.4}\scriptsize{$\pm$0.3} & \textbf{70.2}\scriptsize{$\pm$0.9} & \textbf{62.2}\scriptsize{$\pm$1.0} & \textbf{46.8}\scriptsize{$\pm$1.5} & \textbf{51.3}\scriptsize{$\pm$0.3}\\
\bottomrule
\end{tabular}
\caption{Performance comparison between our method and state-of-the-art methods on ImageNet-A, B, C, D, and E , evaluated via the benchmark of \citet{moser2025unlockingdatasetdistillationdiffusion} under IPC-10. Results are reported as mean Top-1 accuracy on AlexNet, VGG11, ResNet18,
 and ViT. The best performance is highlighted in \textbf{bold}.} 
\label{tab:lm3d} 
\end{table*}
We conducted further comparisons using the evaluation benchmark provided by LD3M\cite{moser2025unlockingdatasetdistillationdiffusion}. This benchmark comprises five ImageNet subsets, designated as A, B, C, D, and E. Evaluations were performed across four distinct network architectures (AlexNet, VGG11, ResNet18, and ViT), reporting the mean top-1 accuracy over 5 runs. We maintained the same hyperparameter configuration as ImageNet-Woof. The IPC-10 evaluation results are presented in the \Cref{tab:lm3d}, where our model achieves comprehensive superiority across all subsets. These results collectively demonstrate the strong cross-dataset and cross-architecture generalization capability of our method.
\suppsubsection{Additional Comparisons on ConvNet-6}
\begin{table}[h]
\centering
\setlength{\tabcolsep}{8pt}
\begin{tabular}{l|ccc}
\toprule
Method   & IPC-10 & IPC-20 &IPC-50 \\
\midrule
Random      &24.3$_{\pm1.1}$ &   29.1$_{\pm 0.7}$&41.3$\pm0.6$ \\
DM \cite{zhao2023dataset}      &26.9$_{\pm1.2}$ &   29.9$_{\pm 1.0}$ & 44.4$_{\pm1.0}$ \\
Glad \cite{cazenavette2023generalizing}&33.8$_{\pm0.9}$\\
IDC-1\cite{kim2022dataset}&33.3$_{\pm1.1}$&35.5$_{\pm0.8}$&43.9$_{\pm1.2}$\\
DiT \cite{peebles2023scalable}      &34.2$_{\pm0.4}$ &36.1$_{\pm0.8}$ &46.5$_{\pm0.8}$  \\
Minimax \cite{gu2024efficient}      &\textbf{37.0}$_{\pm1.0}$ &   37.6$_{\pm 0.2}$&53.9$_{\pm0.6}$   \\
MGD$^3$ \cite{chan2025mgd}      &34.7$_{\pm1.1}$ &   39.0$_{\pm 3.5}$ &54.5$_{\pm1.6}$  \\
\textbf{Ours}      &34.5$_{\pm0.1}$ & \textbf{40.1}$_{\pm0.3}$  &  \textbf{54.9}$_{\pm0.5}$ \\
\bottomrule
\end{tabular}
\caption{Performance comparison between our method and state-of-the-art methods on ImageNet-Woof. Results are reported as Top-1 accuracy on ConvNet-6. The best performance is highlighted in \textbf{bold}.}
\label{tab:covn-6}
\end{table}
We further evaluate our method using ConvNet-6 under various IPC settings. Our approach achieves superior performance at IPC-20 and IPC-50 configurations. However, at IPC-10, our results fall slightly behind Minimax, which can be attributed to our use of the same hyperparameter configuration as employed for IPC-50. Subsequent experiments revealed that allocating distinct hyperparameters for different IPC settings yields more optimal results. This is because lower IPC settings require less emphasis on diversity enhancement while prioritizing the generation of more representative samples. We provide comprehensive analysis and discussion of this aspect in \Cref{tab:AS ON GM} and \Cref{tab:PE}.
\suppsubsection{Additional Ablation Study}
In this subsection, we conduct more detailed ablation study to demonstrate the rationality of the design of our method. We focus on three key aspects of the method design: 1) the construction mechanism of dynamic labels; 2) the impact of different distribution approximation algorithms; 3) the guidance stage of matching terms.

\paragraph{Construction Mechanism of Dynamic Labels}
We compared the construction method using only random noise (Noise) with the dynamic soft label construction method we adopted (Soft label with Noise). The results are presented in \Cref{tab:AS ON DSL}. We found that constructing dynamic labels using only noise terms can also achieve effective performance gains; particularly under the IPC-50 setting, it achieves performance close to that of our full method. This further demonstrates the excellent performance of our proposed dynamic label semantic matching technique in enhancing the diversity of dataset distillation. Moreover, after adding soft label terms, the performance can be further improved, which illustrates the effectiveness of the deterministic shift term we defined for the dataset distillation task. This experiment also proves that designing effective semantic matching guidance is one of the key factors for enhancing dataset distillation performance, which has often been overlooked in previous work. \par
We further conducted an analysis by combining distribution matching (OT), and it can be observed that dataset distillation performance is further enhanced after integrating distribution matching. This also experimentally validates our proposed theoretical framework (\textbf{\Cref{thm:risk_bound_app}}).
\begin{table}[h]
\centering
\setlength{\tabcolsep}{9pt}
\begin{tabular}{l|cc}
\toprule
Dynamic label   & IPC-10 & IPC-50  \\
\midrule
DiT      & 34.7$_{\pm 0.5}$&  49.3$_{\pm0.2}$  \\
\midrule
Noise      & 36.6$_{\pm1.7}$&  59.2$_{\pm1.1}$  \\
Noise + OT       & 40.6$_{\pm1.4}$&    59.7$_{\pm0.3}$  \\
\midrule
Soft label with Noise       & 38.9$_{\pm1.2}$&   59.3$_{\pm0.4}$   \\
Soft label with Noise + OT       &\textbf{40.8}$_{\pm1.2}$ & \textbf{60.1}$_{\pm0.8}$     \\
\bottomrule
\end{tabular}
\caption{Ablation study on different dynamic label construction methods. Results are reported as Top-1 accuracy on ResNet-10 with
average pooling in ImageNet-Woof. The best performance is highlighted in \textbf{bold}.}
\label{tab:AS ON DSL}
\end{table}
\paragraph{Different Distribution Approximation}
\begin{table}[h]
\centering
\setlength{\tabcolsep}{8pt}
\begin{tabular}{c|ccc}
\toprule
Approximation   & IPC-10 & IPC-50 &IPC-100  \\
\midrule
Mean       & \underline{39.6}$_{\pm1.1}$&   58.6$_{\pm0.5}$ &62.5 $_{\pm0.4}$  \\
DBS & 39.2$_{\pm1.6}$ & \underline{59.6}$_{\pm0.5}$&\underline{64.4} $_{\pm0.5}$ \\
K-means       &\textbf{40.8}$_{\pm1.2}$ & \textbf{60.1}$_{\pm0.8}$ &\textbf{65.8} $_{\pm0.2}$  \\
\bottomrule
\end{tabular}
\caption{Ablation study on different distribution approximation methods. Results are reported as Top-1 accuracy on ResNet-10 with
average pooling in Imagenet-Woof. The best performance is highlighted in \textbf{bold}, while the second-best is \underline{underlined}.}
\label{tab:AS ON DA}
\end{table}
\begin{table}[t]
\centering
\setlength{\tabcolsep}{9pt}
\begin{tabular}{l|cc}
\toprule
 Guidance Mechanism  & IPC-10 &IPC-50   \\
 \midrule
\multicolumn{3}{l}{\textbf{Semantic matching}}\\
\midrule
  Dynamic Soft Label    &35.1$_{\pm0.8}$&55.6$_{\pm0.4}$ \\
w/ Stochastic Exploration
       &31.2$_{\pm0.8}$ & 54.3$_{\pm1.5}$     \\
w/ Semantic Refinement
       &\textbf{42.0}$_{\pm1.5}$ & \underline{59.6}$_{\pm1.6}$     \\
\midrule
\multicolumn{3}{l}{\textbf{Distribution matching}}\\
\midrule
Full-stage guidance    &40.4$_{\pm0.5}$& 57.2$_{\pm0.7}$ \\
\midrule
\textbf{Ours Full} &\underline{40.8}$_{\pm1.1}$&\textbf{60.1}$_{\pm0.8}$\\
\bottomrule
\end{tabular}
\caption{Ablation study on different guidance mechanism. Results are reported as Top-1 accuracy on ResNet-10 with
average pooling in Imagenet-Woof. The best performance is highlighted in \textbf{bold}, while the second-best is \underline{underlined}.}
\label{tab:AS ON GM}
\end{table}
Distribution approximation is a key component of our algorithm, and its performance influences the performance of distribution matching based on optimal transport. We conducted an ablation study on three different distribution approximation algorithms. The widely used classical distribution matching loss \cite{zhao2023dataset}, mean matching (mean), can be regarded as a special case of our proposed method when $K=1$. This represents allocating the mass of the entire conditional distribution to the distribution center. Density-based random sampling (DBS) is a sampling-based distribution approximation method, which selects support points by calculating the density of sample points from the original distribution to assign sampling probabilities, and normalizes the densities of different support points as mass coefficients. In the \cref{tab:AS ON DA}, we present the performance comparison of the three methods.\par
The Mean achieves effective performance gains under IPC-10. However, in high IPC settings, the Mean yields overly coarse distribution approximations and fails to fully model the distribution structure. Meanwhile, for diffusion models that can only optimize a single sample at a time, matching against the same mean point impairs diversity. This experimental result also validates our \Cref{prop:epsilon of W_app}. \par
DBS provides effective  signals for distribution matching at high IPC settings. Nevertheless, due to its randomness, DBS often fails to capture comprehensive representative points and particularly overlooks some fine-grained patterns with small $K$. This randomness impairs dataset distillation performance, especially under low IPC settings. \par
The comprehensively leading performance results of our proposed method demonstrate the effectiveness of our proposed local distribution approximation matching. Notably, our distribution approximation technique provides an efficient solution for achieving distribution alignment with limited samples under resource-constrained settings. Compared to MGD$^3$ which requires sample sizes proportional to IPC, our method maintains stable performance with fixed-size samples while achieving effective performance gains even at high IPC settings (e.g., IPC=100).

\paragraph{Guidance Mechanism}
We explored the guidance mechanism, and the results are presented in the \Cref{tab:AS ON GM}. For semantic matching, we found that the Semantic Refinement is necessary. Using only dynamic soft labels or only combining with Stochastic Exploration leads to performance degradation due to insufficient semantic alignment. This further illustrates the criticality of our proposed semantic alignment assumption. However, incorporating the Semantic Refinement can further ensure semantic alignment and improve dataset distillation performance. Especially under lower IPC settings, Dynamic Soft Label combining only Semantic Refinement achieves optimal performance.  In higher IPC settings, further introducing Stochastic Exploration can enhance performance, from $59.6\%$ to $60.1\%$. This empirically validates our earlier discussion: stochastic exploration for diversity enhancement becomes superfluous under low IPC settings, since dynamic labeling already provides sufficient diversity improvement. Therefore, we can further optimize performance by adaptively controlling the temporal window size for stochastic exploration in accordance with the IPC configuration.\par
For distribution matching, we investigated the differences between full-stage guidance and the key-stage guidance we adopted. We found that full-stage guidance fails to improve performance; on the contrary, in high IPC stages, it may significantly impair performance. This is because in the early stage of sampling, samples have not generated sufficient semantic information, and guidance through distribution matching at this point will produce erroneous guidance signals. Meanwhile, in the later stage, gradient-based guidance may also introduce artifacts into the images, so guidance should be terminated early. Our observations on loss values also illustrate this point: $\mathcal{L}_{\text{OT}}$ decreases only in the key-stage. Therefore, performing distribution guidance only in the key-stage is a reasonable and efficient choice.
\suppsubsection{Additional Hyperparameter Analysis}
\begin{table}
\centering
\setlength{\tabcolsep}{13pt}
\begin{tabular}{l|cc}
\toprule
 Hyperparameter  & IPC-10  & IPC-50  \\
\midrule
$\beta_n=0.01$   &42.7$_{\pm1.3}$&58.7$_{\pm0.7}$  \\
$\beta_n=0.04$      &  40.5$_{\pm0.2}$  &60.1$_{\pm0.7}$   \\
$\beta_n=0.1$   &36.3$_{\pm0.6}$&60.2$_{\pm0.7}$ \\
\midrule
$\beta_s=0.04$   &41.5$_{\pm0.5}$ & 59.9$_{\pm0.7}$  \\
$\beta_s=0.06$ &41.2$_{\pm0.5}$ & 59.7$_{\pm1.3}$ \\
$\beta_s=0.1$ &41.3$_{\pm1.1}$ & 59.9$_{\pm0.8}$\\
\midrule
$1+\omega=1$&19.9$_{\pm0.5}$ & 38.6$_{\pm1.2}$\\
$1+\omega=7$&38.9$_{\pm1.1}$ & 57.6$_{\pm1.6}$\\
\midrule
$tw=\left [20,45 \right ]$ &40.6$_{\pm0.8}$&60.4$_{\pm0.9}$\\
$tw=\left [30,45 \right ]$ &40.4$_{\pm1.0}$&59.6$_{\pm0.6}$\\
$tw=\left [25,40 \right ]$&39.8$_{\pm0.4}$ &59.9$_{\pm0.6}$\\
$tw=\left [25,50 \right ]$&42.0$_{\pm1.5}$&59.6$_{\pm1.6}$\\
\midrule
Ours &40.8$_{\pm1.1}$ & 60.1$_{\pm0.8}$\\
\bottomrule
\end{tabular}

\caption{Evaluation of different parameter. Results are reported as Top-1 accuracy on ResNet-10 with
average pooling in ImageNet-Woof.}
\label{tab:PE}
\end{table}
We analyzed the hyperparameters involved in semantic matching, including CFG scale $1+\omega$, temporal window $tw$, modulation coefficient $ \beta_n$ and $ \beta_s$. Results of the hyperparameter analysis are presented in the \Cref{tab:PE}.\par
$ \beta_n$ is used to regulate the intensity of the noise term. A larger $ \beta_n$ will result in stronger stochastic exploration and further enhance the diversity of generation. Therefore, a larger $\beta_n$ can enhance performance under high IPC settings but may also impair performance under low IPC settings. Specifically, we found that under low IPC settings, $ \beta_n=0.01$ achieves optimal performance. This validates our assumption that under low IPC settings, randomness should be reduced to generate representative key samples, whereas under high IPC settings, greater consideration of diversity is needed to achieve better performance.\par

    
\begin{figure*}[h]
    \centering
        \begin{subfigure}[b]{0.33\textwidth}
        \centering
        \includegraphics[width=\textwidth]{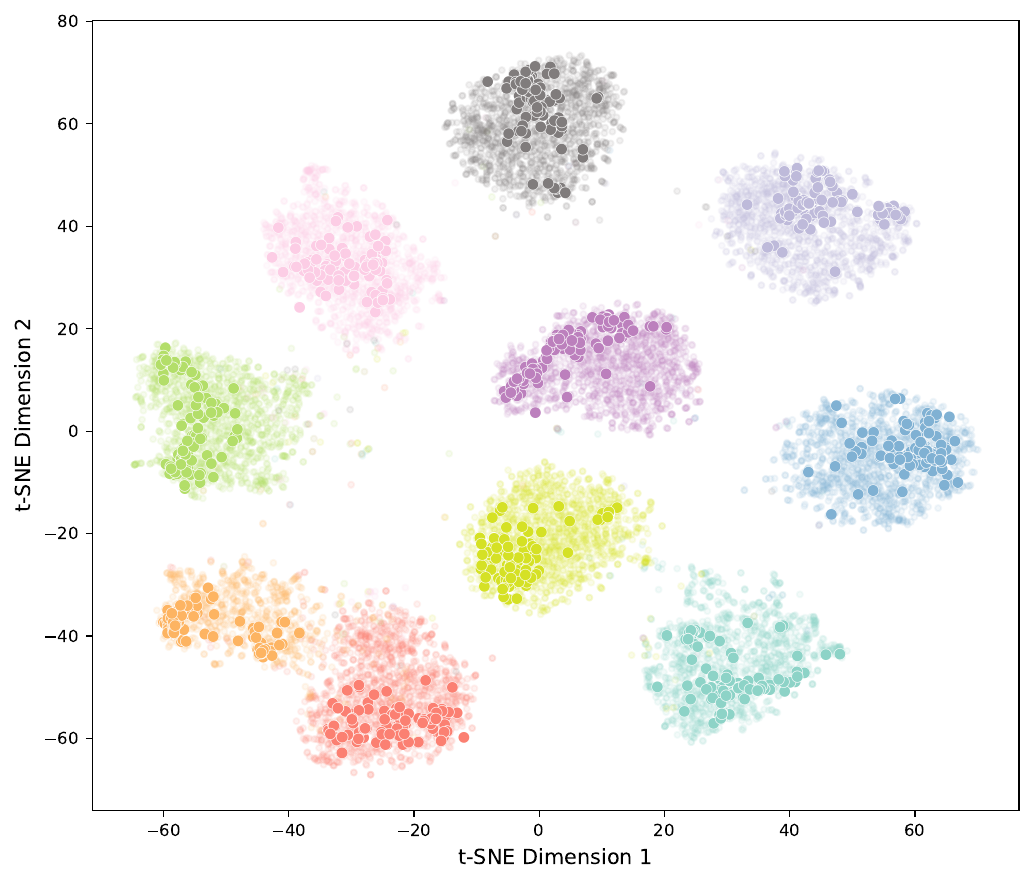} 
        \caption{DiT} 
        \label{DiT} 
    \end{subfigure}
    \begin{subfigure}[b]{0.33\textwidth}
        \centering
        \includegraphics[width=\textwidth]{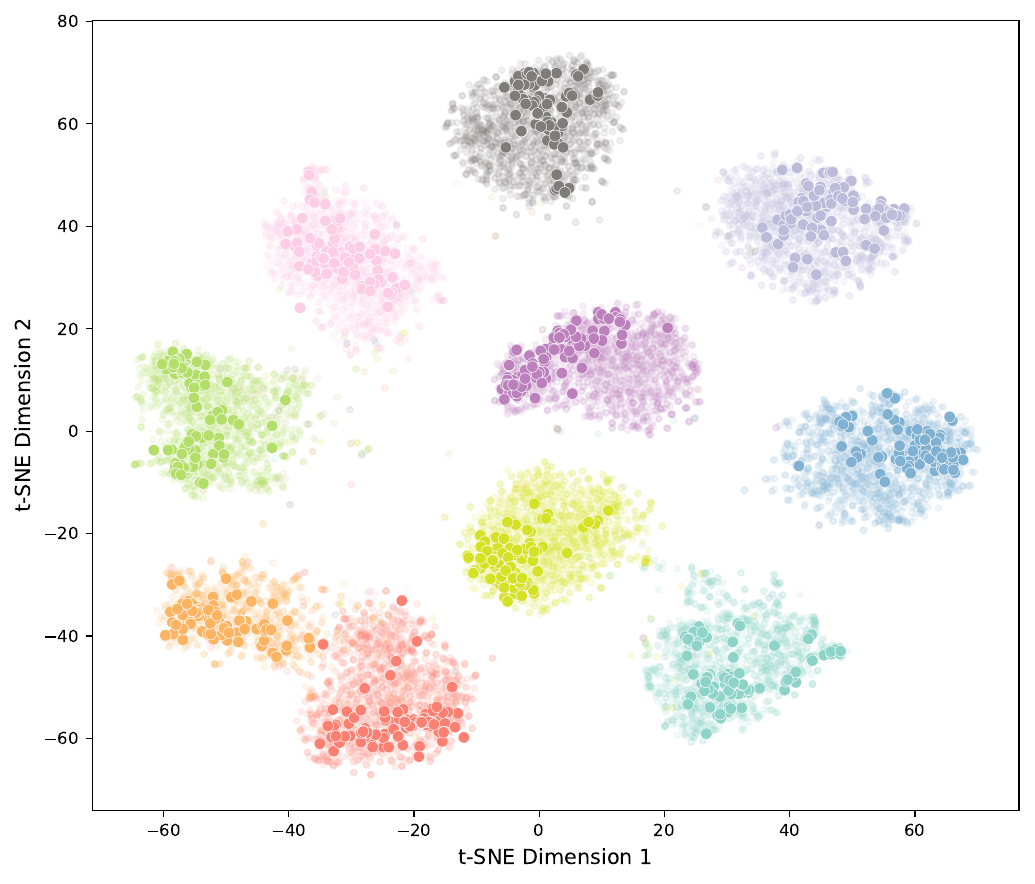} 
        \caption{Minimax}
        \label{Minimax}
    \end{subfigure}
    \begin{subfigure}[b]{0.33\textwidth}
        \centering
        \includegraphics[width=\textwidth]{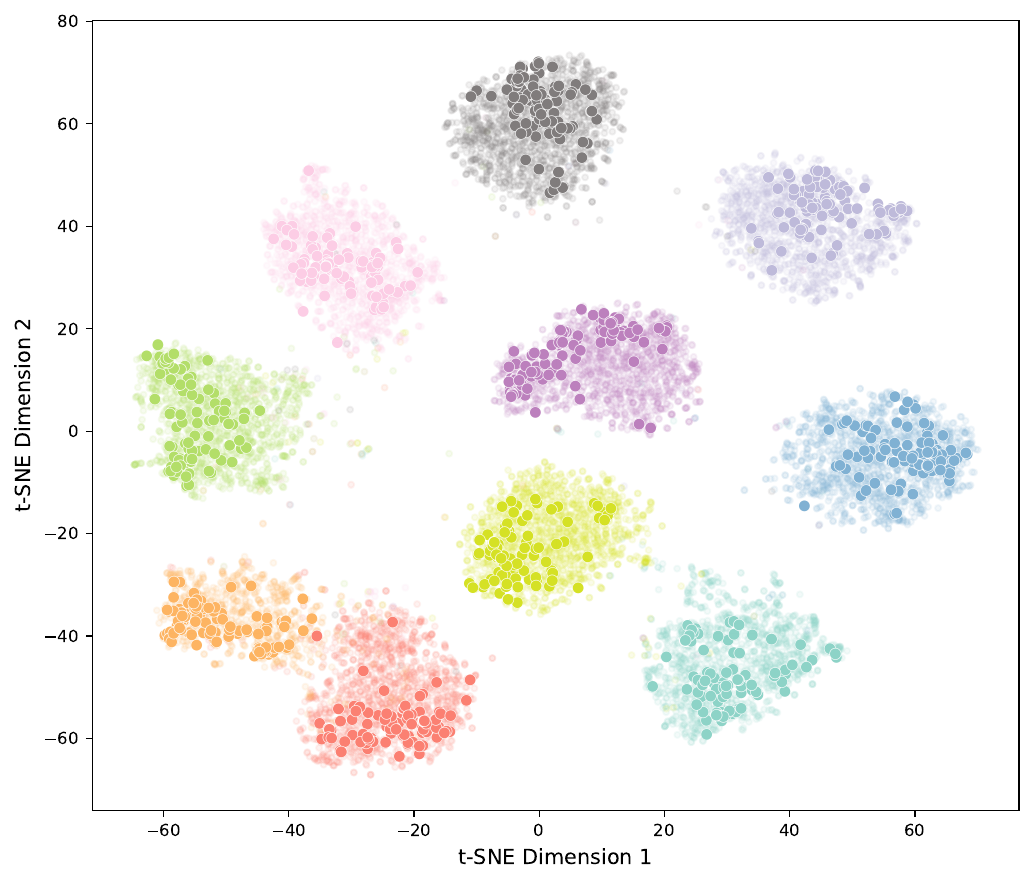} 
        \caption{Ours}
        \label{Ours}
    \end{subfigure}
    \begin{subfigure}[b]{0.33\textwidth}
        \centering
        \includegraphics[width=\textwidth]{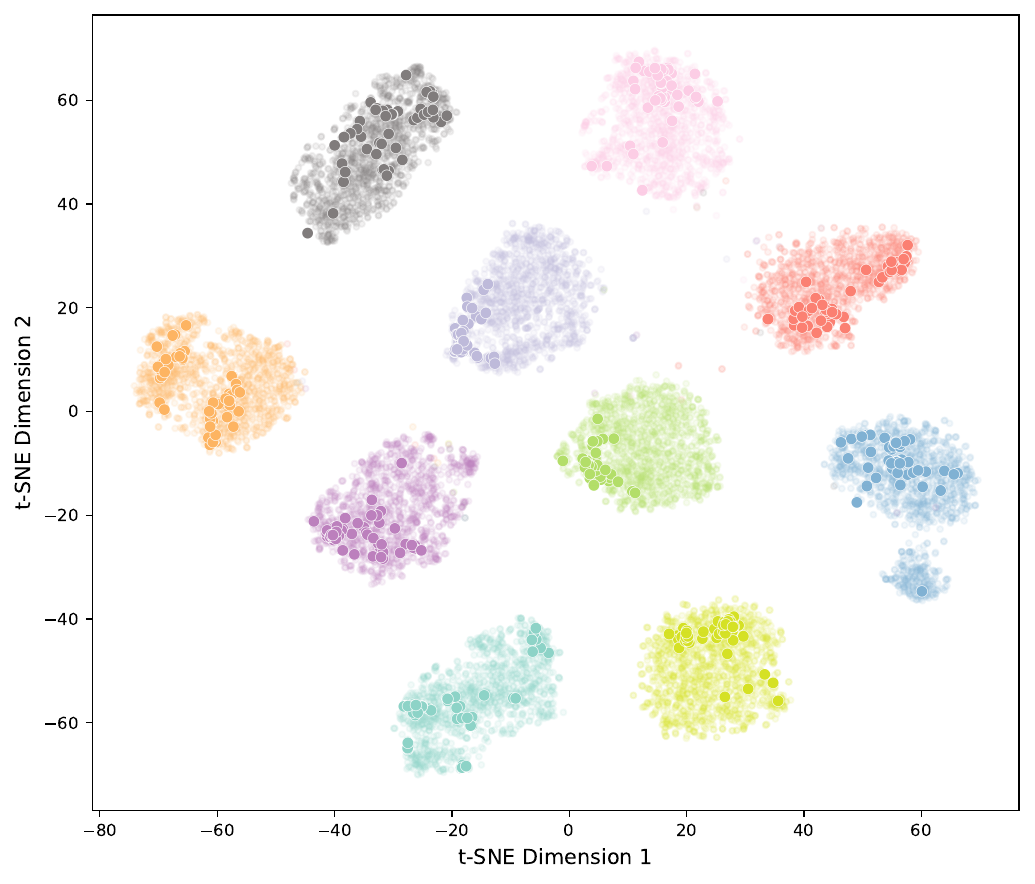} 
        \caption{DiT} 
        \label{DiT_nette} 
    \end{subfigure}
    \begin{subfigure}[b]{0.33\textwidth}
        \centering
        \includegraphics[width=\textwidth]{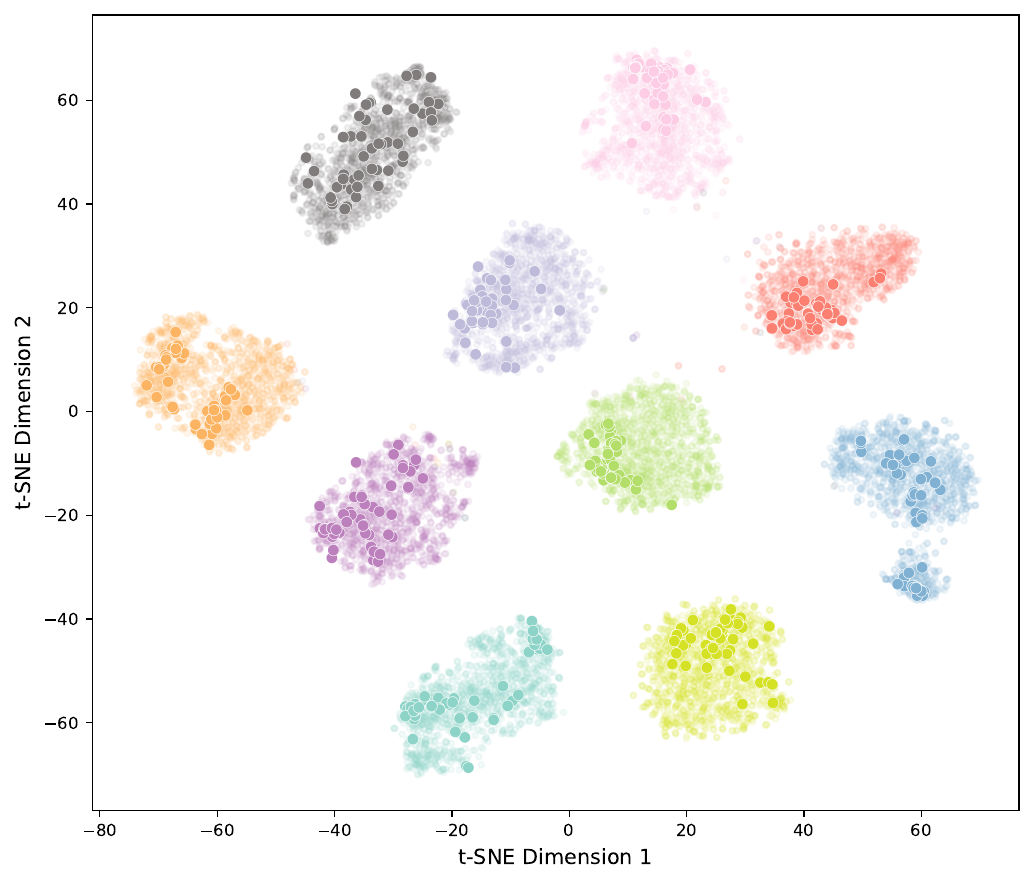} 
        \caption{Minimax}
        \label{Minimax_nette}
    \end{subfigure}
    \begin{subfigure}[b]{0.33\textwidth}
        \centering
        \includegraphics[width=\textwidth]{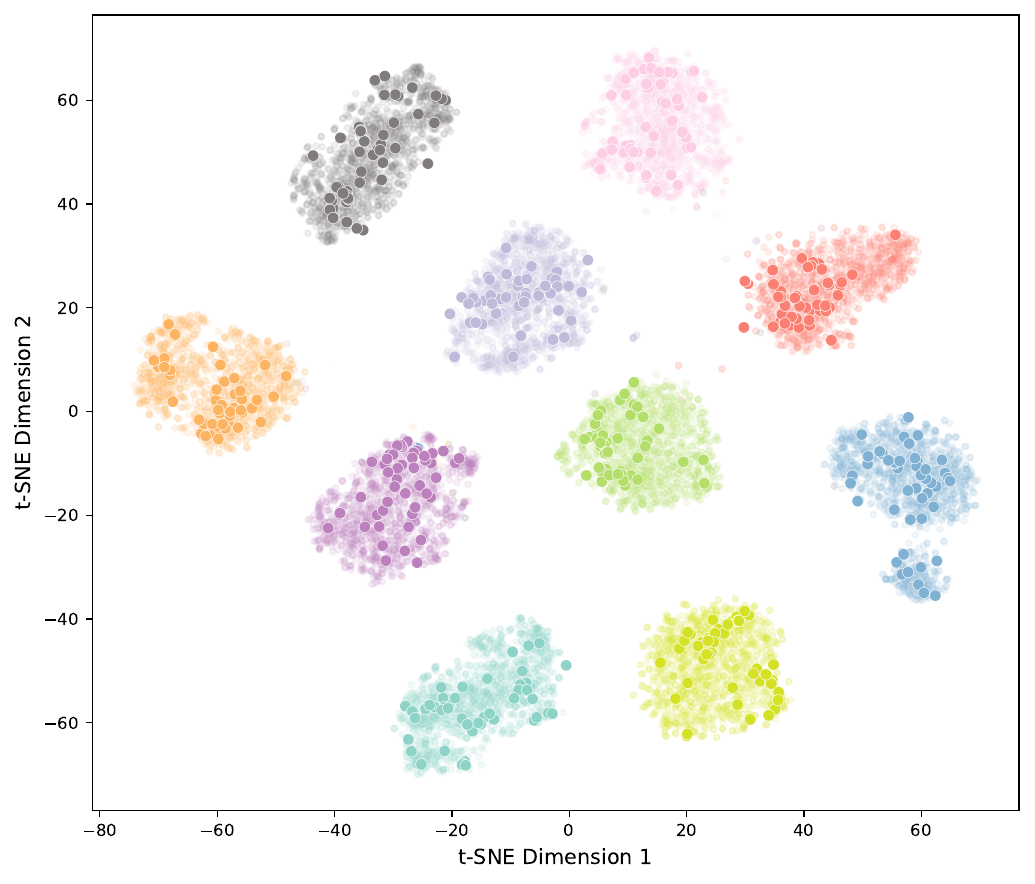} 
        \caption{Ours}
        \label{Ours_nette}
    \end{subfigure}

    \caption{\textbf{Distribution Visualization:} Visualization results of sample distributions for surrogate datasets generated by different methods and the original dataset: top row corresponds to ImageNet-Woof under IPC-100 setting, bottom row corresponds to ImageNet-Nette under IPC-50 setting.} 
    \label{fig:dsv_app} 
\end{figure*}
Our analytical experiments on $\beta_s$ further demonstrate the role of our soft label terms. We found that under low IPC settings, using stronger guidance via soft label terms can generate more informative samples, thereby further improving performance. Meanwhile, under high IPC settings, soft label terms of different intensities can all ensure stable performance. Under different IPC settings, appropriately adjusting $\beta_s$ and $\beta_m$ is undoubtedly a better choice. We recommend that for low IPC settings, $\beta_s$ can be increased while $\beta_n$ is decreased to generate representative samples with high information concentration. In contrast, under high IPC settings, we focus more on enhancing diversity, and increasing $\beta_n$ will further strengthen performance. To demonstrate the generality of our method across different IPC settings, we adopted unified parameters $\beta_s=0.01$ and $\beta_n=0.06$, which still achieve SOTA performance.\par
The parameter $1+\omega$ controls the strength of semantic matching. When $1+\omega=1$, the diffusion model fails to capture sufficient semantic information, resulting in significantly degraded performance. Conversely, at $1+\omega=7$, overly strong semantic alignment impairs generation quality, also leading to decreased performance. Therefore, we set $1+\omega=4$.\par
Additionally, we performed hyperparameter validation on the temporal window using a step size of 5. We observed that within small variation ranges, the temporal window exhibits minimal impact on the overall experimental results, which aligns with our hypothesis regarding the diversity-representativeness trade-off. For more extreme variations, we have conducted corresponding experiments as documented in \Cref{tab:AS ON GM}. The selection of this parameter is informed by empirical observations of the diffusion model sampling process in prior literature \cite{yu2023freedom}, with its effectiveness further verified through our sensitivity analysis.
\suppsubsection{Generation Quality Evaluation}
We evaluated the approach using additional generative quality metrics. Results are presented in the \Cref{tab:rd_results}. On common generative quality metrics, our method achieves performance comparable to the original DiT, demonstrating the realism of our generated samples.\par
Notably, we observe that diversity enhancement inevitably incurs a slight compromise in semantic representativeness, reflected by marginally reduced precision. However, the substantially improved recall demonstrates our method's enhanced diversity, which ultimately translates into performance gains. As previously discussed, this minor representational trade-off becomes negligible under high IPC settings. Furthermore, such representational degradation can be effectively mitigated by incorporating soft-label criteria during subsequent distillation stages.

\begin{table}[h]
\centering
\setlength{\tabcolsep}{8pt}
\begin{tabular}{l|ccc}
\toprule
Method   & DiT  & Minimax& \textbf{Ours} \\
\midrule
FID       &48.6 &   49.2 &48.8\\
Precision ($\%$)     &91.2 &   94.4&92.4\\
Recall ($\%$)      & 51.2& 49.5 &57.8 \\
Density ($\%$)      & 1.19& 1.38 &1.36 \\
\bottomrule
\end{tabular}
\caption{Evaluation of generation quality evaluation of 10 classes each with 100 images in ImageNet-Woof. }
\label{tab:rd_results}
\end{table}
\suppsubsection{Visualization}
    
\paragraph{Generated Samples Visualization:}

\begin{figure*}[htbp]
    \centering
    \includegraphics[width=0.85\linewidth]{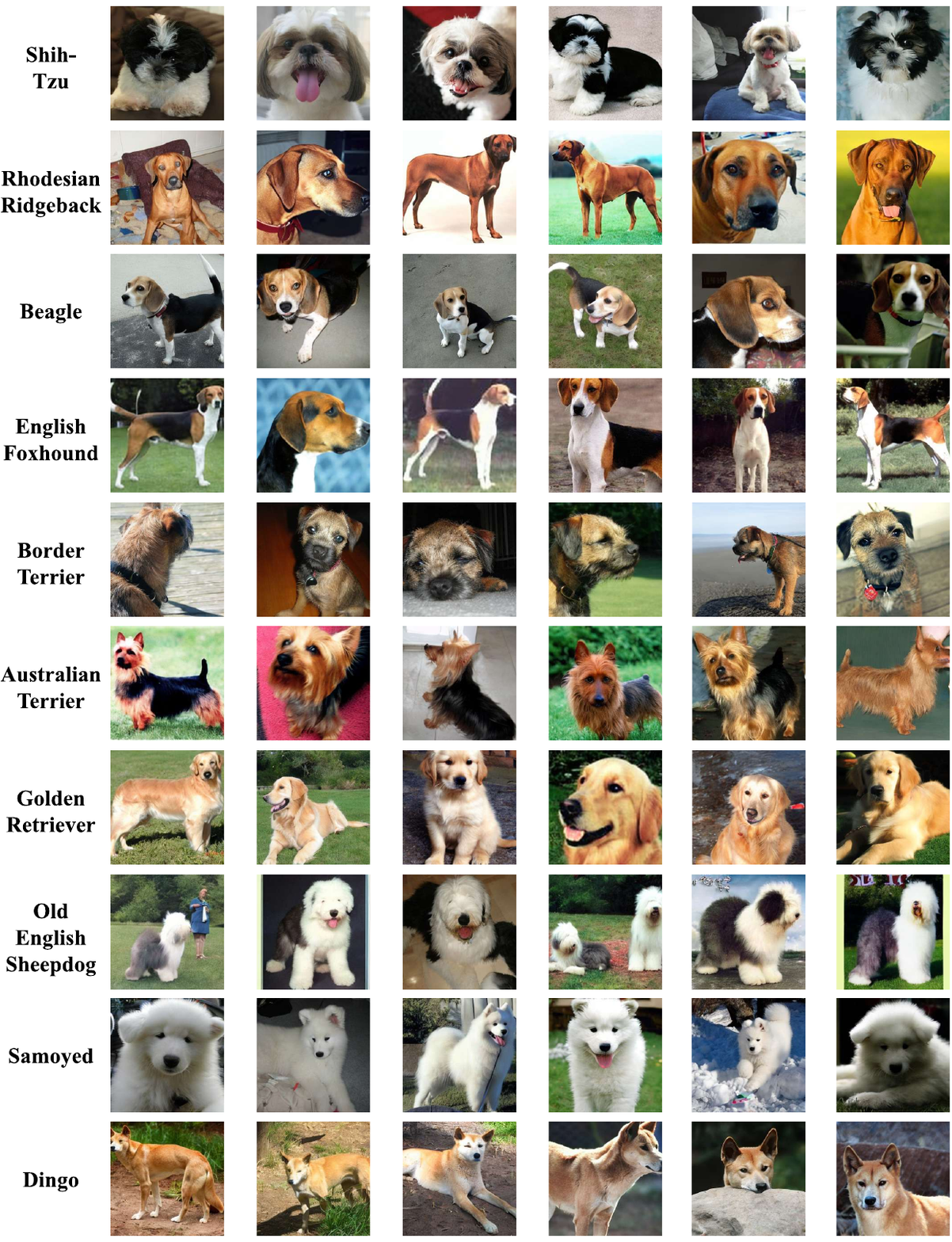}
    \caption{Generated samples are from our proposed DMGD method for the ImageNet-Woof dataset. We present the randomly selected generated samples under the IPC-50 setting. The class names are marked at the left of each row.}
    \label{fig:gsv2}
\end{figure*}

\begin{figure*}[htbp]
    \centering
    \includegraphics[width=0.85\linewidth]{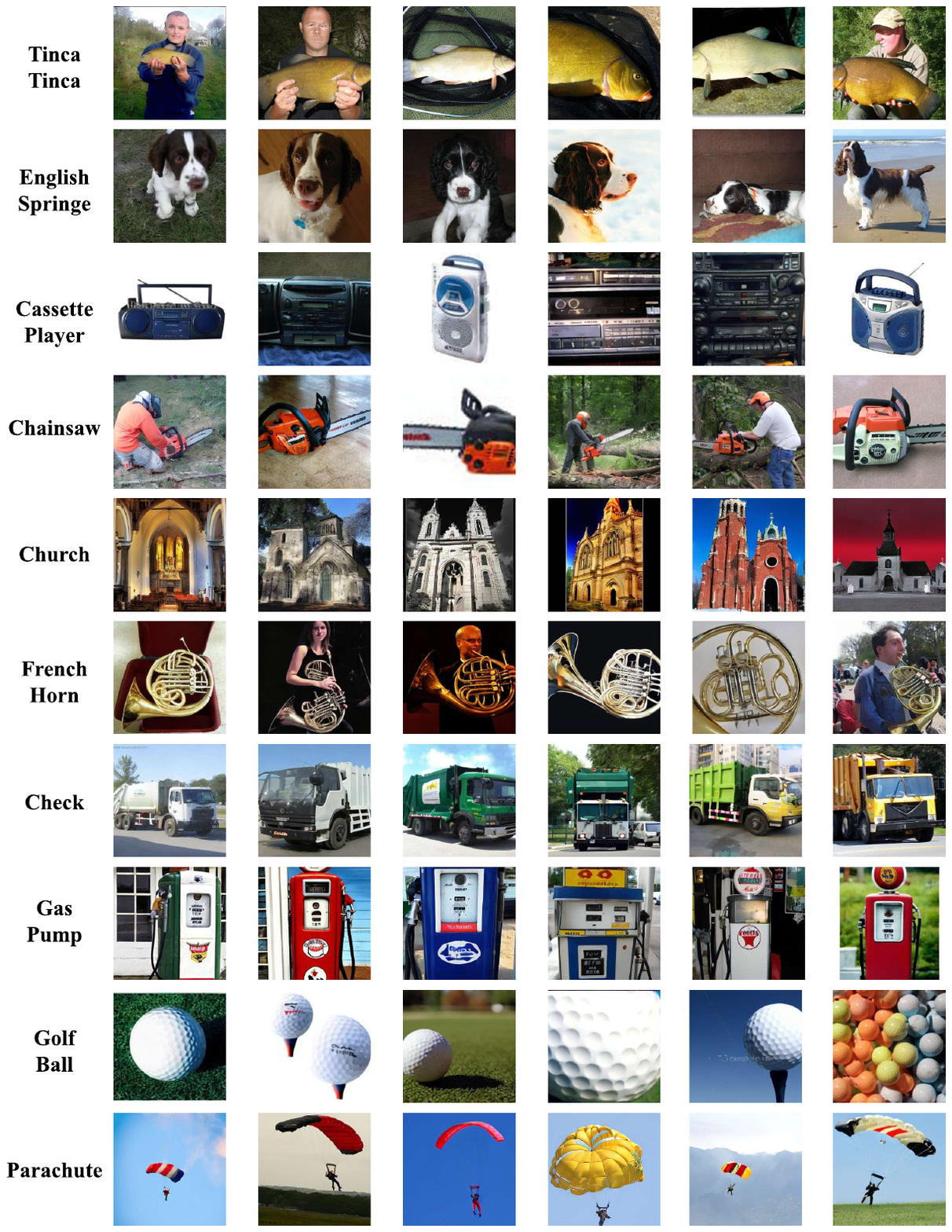}
    \caption{Generated samples are from our proposed DMGD method for the ImageNet-Nette dataset. We present the randomly selected generated samples under the IPC-50 setting.  The class names are marked at the left of each row.}
    \label{fig:gsv3}
\end{figure*}

We present generated samples for visualization. 
 \cref{fig:gsv2} and \cref{fig:gsv3} present generated examples on the ImageNet-Nette and ImageNet-Woof datasets, respectively. The generated samples were randomly selected under the IPC-50 setting and arranged from left to right in the order of generation. This visualization demonstrates our method's intra-class diversity: earlier samples exhibit stronger semantic representativeness, while later samples display greater uniqueness.
\paragraph{Distribution Visualization:}

We further visualize the sample distributions using t-SNE. \Cref{fig:dsv_app} presents the distributions of DiT, Minimax, and our method within the same feature space (Inception-v3) on Imagenet-woof and Imagenet-Nette, demonstrating our approach's diverse semantic matching capability and effective distribution alignment. Our method achieves superior coverage of the target dataset's distribution.
\begin{figure}[h]
    \centering
    \includegraphics[width=1\linewidth]{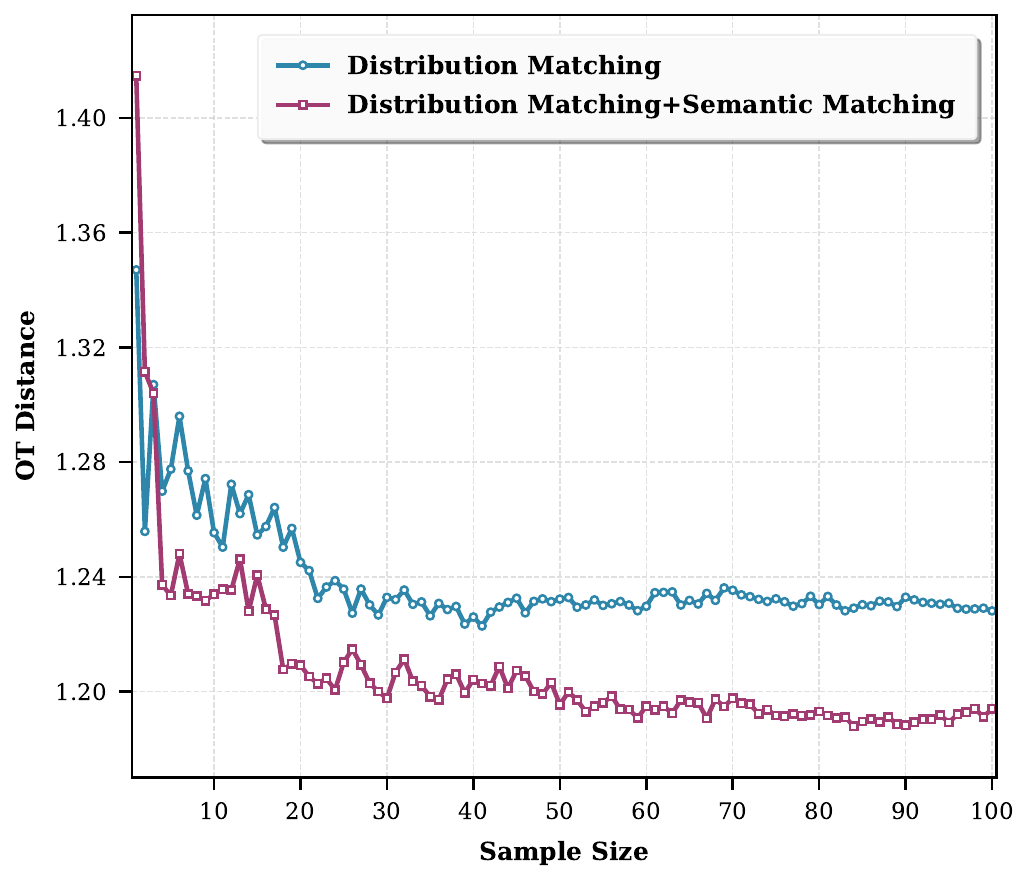}
    \caption{\textbf{OT Distance Visualization:} We systematically recorded the final optimal transport (OT) distance loss for each sample during progressive distillation. A randomly selected category from ImageNet-Woof was visualized to illustrate the results.}
    \label{fig:otv}
\end{figure}
\paragraph{OT Distance Visualization:}To better illustrate the relationship between distribution matching and semantic matching, we visualize the optimal transport (OT) distance losses under two scenarios: using distribution matching alone (Distribution Matching), and combining distribution matching with diversity-enhanced semantic matching (Distribution Matching+Semantic Matching). Each data point represents the final OT distance loss of an individual sample during progressive distillation. The visualization results for a randomly selected class from ImageNet-Woof are presented in the \Cref{fig:otv}. It can be observed that our distribution matching module effectively optimizes the OT loss during dataset distillation. However, due to the diffusion models tendency to generate homogeneous samples, this optimization is prone to converge to local optima. In contrast, the diversity-enhanced semantic matching provides superior distribution exploration capability, which not only accelerates OT loss optimization but also alleviates the local optimum problem. These findings validate that our proposed dual-matching framework exhibits no optimization conflicts, thereby further demonstrating the effectiveness of DMGD.

\subsection{Limitation}
Currently, our method is confined to dataset distillation with limited semantic scope. Exploration regarding diffusion models possessing general semantic properties and more complex datasets remains insufficient. Furthermore, due to inherent constraints of diffusion models, our approach can not generalize to other data modalities, such as audio \cite{chen2025rapverse}, video \cite{lu2025reward}, time-series \cite{shao2024novel} or embodied AI data \cite{lu2025discovery}. In future work, we aim to push the boundaries of dataset distillation towards a more universal and efficient paradigm.


\end{document}